\documentclass[10pt]{article} 
\usepackage[preprint]{tmlr}

\usepackage{natbib}
\usepackage{fancyvrb}
\usepackage{amsfonts} 
\usepackage{amssymb}
\usepackage{amsmath}
\usepackage{booktabs} 

\usepackage{xcolor}
\usepackage{graphicx}
\usepackage{comment}
\usepackage{bm}
\usepackage{caption}
\usepackage{algorithm}
\usepackage{subfigure}
\usepackage{subcaption} 
\usepackage{pifont}
\usepackage{mdframed}
\usepackage[dvipsnames]{xcolor}
\usepackage{csquotes}
\usepackage[edges]{forest}
\usepackage{algorithm}
\usepackage{algpseudocode}
\usepackage{multirow}

\usepackage{hyperref}
\usepackage{cleveref}
\usepackage{url}

\newcommand{\cmark}{{\color{ForestGreen}\ding{52}}\hspace{0.5em}}
\newcommand{\xmark}{{\color{red}\ding{56}}\hspace{0.5em}}

\newcommand{\orangepoint}{\raisebox{-0.9ex}{\textcolor{orange}{\Huge\textbullet}}}
\newcommand{\redpoint}{\raisebox{-0.9ex}{\textcolor{red}{\Huge\textbullet}}}
\newcommand{\greenpoint}{\raisebox{-0.9ex}{\textcolor{ForestGreen}{\Huge\textbullet}}}

\usepackage{amsmath,amsfonts,bm}

\def\eqref#1{equation~\ref{#1}}

\def\1{\bm{1}}

\DeclareMathAlphabet{\mathsfit}{\encodingdefault}{\sfdefault}{m}{sl}
\SetMathAlphabet{\mathsfit}{bold}{\encodingdefault}{\sfdefault}{bx}{n}

\title{On the post-hoc Evaluation of PDE Discovery:\\ A Multifaceted Challenge of Scientific Advancement}

\author{\name Baptiste Mathevon \email baptiste.mathevon@univ-st-etienne.fr \\
      \addr Université Jean Monnet Saint-Étienne, CNRS, Institut d’Optique Graduate School, Inria, Laboratoire Hubert Curien UMR 5516, F-42023, SAINT-ÉTIENNE, France
      \AND
      \name Farah Cherfaoui \email farah.cherfaoui@univ-st-etienne.fr \\
      \addr Université Jean Monnet Saint-Étienne, CNRS, Institut d’Optique Graduate School, Inria, Laboratoire Hubert Curien UMR 5516, F-42023, SAINT-ÉTIENNE, France
      \AND
      \name Amaury Habrard
      \email amaury.habrard@univ-st-etienne.fr \\
      \addr Université Jean Monnet Saint-Étienne, CNRS, Institut d’Optique Graduate School, Inria, Laboratoire Hubert Curien UMR 5516, F-42023, SAINT-ÉTIENNE, France, 
      Institut Universitaire de France (IUF)
      \AND
      \name Marc Sebban
      \email marc.sebban@univ-st-etienne.fr\\
      \addr Université Jean Monnet Saint-Étienne, CNRS, Institut d’Optique Graduate School, Inria, Laboratoire Hubert Curien UMR 5516, F-42023, SAINT-ÉTIENNE, France}

\def\month{11}  
\def\year{2026} 
\def\openreview{\url{https://openreview.net/forum?id=6RiT68mDk5}} 

\begin{document}
\maketitle

\begin{abstract}
    Partial differential equation (PDE) discovery aims to identify from data the  governing law of a physical system. Constituting  a cornerstone of scientific advancement, it has  become during the past decade a major line of research in the rapidly evolving field of Physics-informed Machine Learning (PiML). Among the  open problems to address in this domain, 
    the post-hoc evaluation of  
    discovered PDEs raises the particular difficulty of being multifaceted. Indeed, it requires  jointly considering predictive accuracy, physical consistency, interpretability, and out-of-distribution generalization capacity, some  of these properties being conflicting.  
     \textcolor{black}{Drawing on an extensive body of literature spanning machine learning, numerical analysis, information theory, and symbolic regression, we establish in this article that the overwhelming majority of papers address only certain aspects of the overall evaluation task. To better understand their limitations when used in isolation from the others, we propose, to our knowledge, the first taxonomy of PDE evaluation metrics. 
     Through experiments conducted on several PDEs, we observe that limiting the scope of the evaluation can lead to overstated conclusions as well as changes in the ranking of the identified equations (and, consequently, of the discovery algorithms). For this reason, we further provide recommendations  with the aim of promoting standardized and reliable practices. This results in the design of {\sc Pde-Check}, a simple rules-based protocol to guide practitioners  according to the scenario under consideration, including whether or not the ground truth equation is available (recovery or discovery), whether data is available beyond the training time frame and/or under varying initial conditions, or whether physical priors is available.}       We argue that this  paper is intended both for ML experts who design  new PDE discovery algorithms and for users of these methods aiming, in real applications, to discover and validate well-founded scientific  laws.
\end{abstract}


\section{Introduction}
\begin{figure}[t]
    \centering
\includegraphics[width=\linewidth]{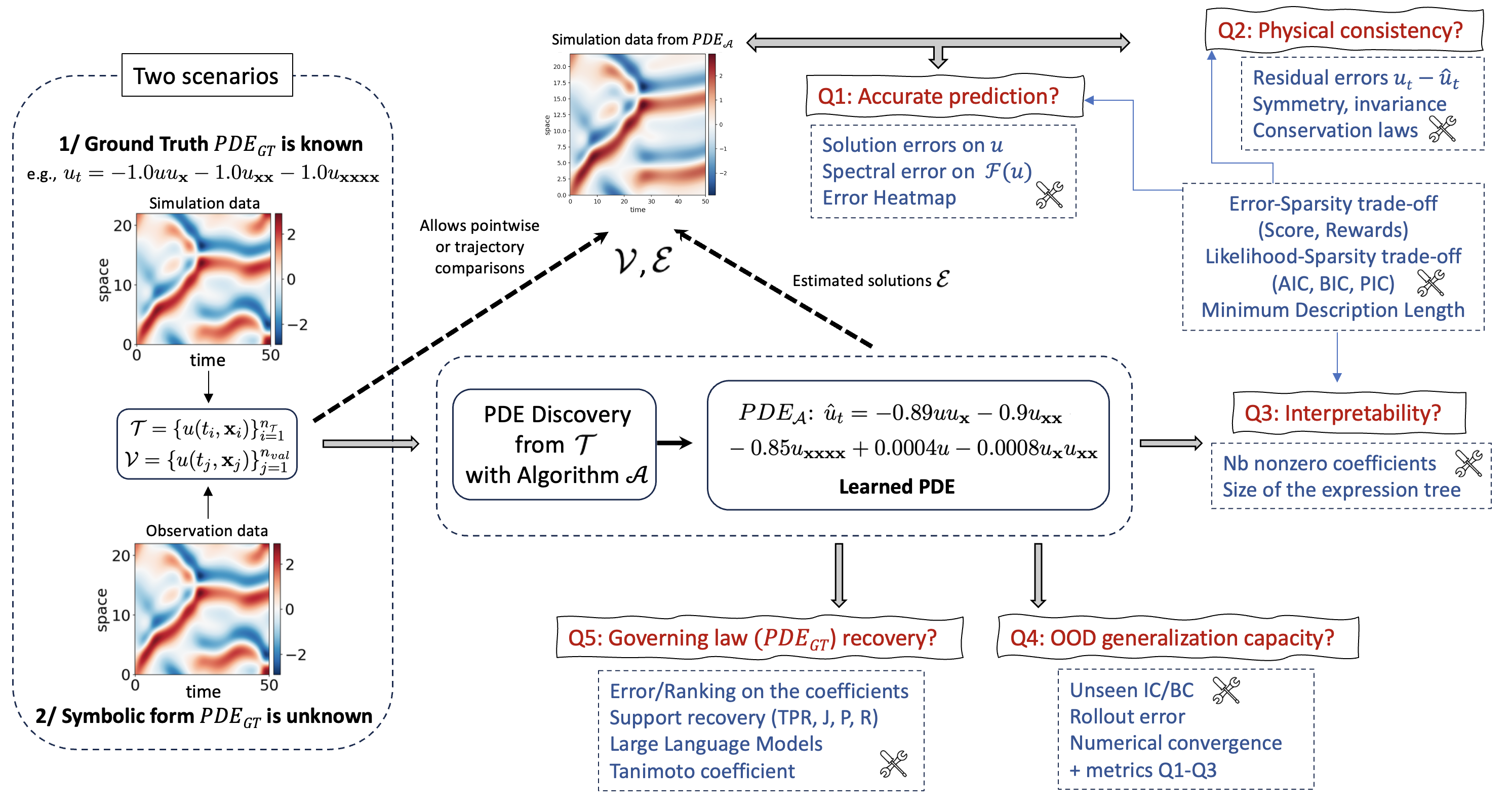}
        \caption{Overview of  PDE discovery  and metrics (in blue) devoted to address the multifaceted challenge of post-hoc PDE evaluation in accordance with the five key principles to be met (in red). The illustrations of dynamics correspond to a physical system leading to complex pattern formation and chaotic behavior. They are  simulated from either the ground truth Kuramoto-Sivashinsky equation ($PDE_{\cal GT}$) or a learned $PDE_{\cal A}$, or they correspond to observation data ({\it e.g.} obtained by laser-matter interaction for which $PDE_{\cal GT}$ is a good candidate for modeling this self-organization phenomenon). 
        Three datasets are typically used in PDE discovery: a training set ${\cal T}$ is employed for learning $PDE_{\cal A}$ with some discovery algorithm ${\cal A}$; a validation set ${\cal V}$ is compared to a set ${\cal E}$ of estimated solutions, simulated from  $PDE_{\cal A}$, for evaluating the discovered law from a pointwise or trajectory perspective. 
} \label{fig:evalPDE}
\end{figure}

Alongside {\it forward} problems such as learning fast neural surrogate solvers ({\it e.g.} PINNs \citep{raissi2019physics} or FNOs~\citep{LiKALBSA21}), partial differential equation (PDE) discovery is one of the main {\it inverse} problems studied in the  active field of Physics-informed Machine Learning (PiML) \citep{osti_1852843}. It consists in  automatically identifying from simulation or observation data the  governing equation(s) of a physical  system. This task becomes especially critical when addressing real applications  where there is no prior knowledge of the underlying physics (or only a partial understanding of it),  which may involve the complex interplay of several physical phenomena. While a broad range of PDE discovery methods (see, {\it e.g.}, \cite{Brunton2016,rudy2016datadrivendiscoverypartialdifferential,forootani2023robustsindyapproachcombining,STEPHANY2024106242,banna2025unrolledsindystableexplicitmethod}) has flourished in the past few years, often envisioning the task from the perspective of
sparse regression, it is worth noting that 
a universally accepted methodology for the systematic evaluation of identified PDEs remains unestablished~\citep{feng2025physpde}.  
 The main reason is that  post-hoc assessment, as commonly acknowledged in the literature \citep{Kaheman_2020,Xu_2023,bideh2026mdbenchbenchmarkingdatadrivenmethods}, is inherently multifaceted  and difficult to capture with a single measure, insofar as the fitness accuracy, physical consistency, parcimony and generalization capacity, have to be  considered jointly. 
More precisely, we argue that a discovered PDE  
is expected to meet several  objectives, some of which may be in conflict, 
 that  we propose to formulate in terms of the
 following key questions (see Fig.~\ref{fig:evalPDE} for an overview):   {\bf Q1:} {\it Does the identified PDE allow to lead to accurate solutions (up to numerical errors)? {\bf Q2:} Is this equation physically consistent ({\it i.e.}, is there any violation of fundamental physical principles)? {\bf Q3:} Is it simple enough so as to be interpreted as a general scientific discovery? {\bf Q4:} What is its ability to generalize out of distribution (long-horizon prediction, adaptation to different initial/boundary conditions or changes in dynamical regimes})?  {\bf Q5:} {\it To what extent does it agree with the ground-truth (GT) governing PDE, when the latter is available?}

 These questions naturally relate to the classical statistical learning setting, notably with respect to the \sloppy{bias–variance} trade-off. In particular, {\bf Q1} is  directly connected to data fidelity, 
 while {\bf Q3} refers to model complexity. As in standard machine learning, increasing the latter may improve the former and reduce bias, but comes at the cost of higher variance; conversely, reducing model complexity may increase bias while lowering variance. 
 However, beyond this similarity, the following distinction should be noted: 
the notion of complexity is stronger in PDE discovery because it demands interpretability and consistency at the level of a meaningful scientific law, thereby  requiring going beyond mere sparsity. Therefore, we argue that in PDE discovery, the "\textit{simpler is better}" paradigm is less straightforward than in conventional machine learning. On the other hand, it is worth noticing that {\bf Q4} extends the standard notion of generalization by explicitly targeting out-of-distribution extrapolation, which becomes central in dynamical systems. Indeed, 
it is often necessary in real applications to simulate the PDE under varying conditions, time horizons and regimes, allowing to assess the risk of drawing overstated conclusions. Finally, while modern neural architectures may incorporate symmetries or equivariances which partially draw inspiration from physics, PDE discovery (via {\bf  Q2}) requires adherence to actual fundamental physical principles ({\it e.g.} conservation laws), making the discovery and evaluation tasks more theoretically grounded. \textcolor{black}{It should be noted that Question \textbf{Q5} deserves special consideration. Indeed, it is what distinguishes the task of discovering new knowledge from that of recovering a known equation. As we will show in this paper, this distinction between scenarios is essential and has implications for how the evaluation should be approached, and consequently for the metrics that should be used.}

Evaluating a PDE solely through the lens of 
one of these questions can lead to drawing erroneous conclusions and formulating inconsistent hypotheses about the governing physical theory. For instance, it is well established that a small reconstruction error does not always reflect
the symbolic correctness of the discovered equation ({\it e.g.}, spurious terms that cause small  errors over a short time period can lead, over a longer horizon, to a dramatic deterioration of the solutions). Moreover, despite the realistic assumption that most physical systems are governed by PDEs composed of a few terms, it is also widely known that the simplicity of an equation  does not necessarily guarantee its physical consistency ({\it e.g.} a sparse PDE may violate symmetry properties or conservation laws). On the other hand, as theoretically studied for PINNs~\citep{girault:hal-04518335,deryck2023error,10.3150/24-BEJ1799}, observing small residuals, even on unseen validation data, does not automatically mean that the discovered equation is correct and leads to accurate solutions. 
 Considering the diversity of metrics, which are hardly comparable since they capture different properties of the PDE, one may feel at a loss when it comes to determining whether a learned equation represents a valid physical law. 
 
Although PDE evaluation is intrinsically multi-criteria, we emphasize that  this paper is not positioned within a multi-objective optimization framework \citep{Miettinen1999}, 
where one typically seeks Pareto-optimal solutions to identify trade-offs among conflicting criteria as done, {\it e.g.}, in genetic algorithms, decomposition-based methods, multi-objective gradient-based approaches, etc. Said differently, 
we do not view the metrics considered in this study as means of guiding 
PDE discovery algorithms (some of those criteria are non-separable functions and cannot even be optimized). Instead, they are treated as post-training measures to evaluate an equation, during a downstream task,   
regardless of the mechanism from which it originates. Nonetheless, certain metrics commonly used in multi-objective optimization can be leveraged in post-hoc PDE evaluation. This applies to those  employed in symbolic regression~\citep{brum2026discoveringequationsdatasymbolic}\footnote{A substantial part of the PDE discovery literature can be cast, as a special case, within this framework, albeit more challenging due to the presence of derivative terms.}, which will thus be studied below.  
 
  By decoupling the identified PDE from the discovery process,  we shift the focus of the evaluation task from the learning algorithm to the scientific discovery itself. 
 Moreover, by doing so, we argue that these metrics may thus be applied in scenarios that do not necessarily involve learning but still require a valid evaluation procedure\footnote{Note that for the sake of consistency in notation and terminology, we will remain within the scope of conventional PDE discovery throughout this article, assuming that the  equation has been learned from data, while not exploiting the algorithm itself in the downstream evaluation task.}. 
This may hold, for instance, when  \textit{human experts formulate a new hypothesis} for a physical law based on their own theoretical knowledge and  observations of a physical system. It also applies in an attempt to \textit{simplify a complex existing law}, motivated {\it e.g.} by the need (i) to  reduce the computational cost, making large-scale simulations feasible, (ii) to isolate the dominant physical mechanisms, thereby improving interpretability, or (iii) to facilitate theoretical analysis (existence/uniqueness of solutions, asymptotic guarantees, etc.) that would be impossible without sparsifying the equation. The \textit{comparison of competing conjectures} proposed in the literature to model the same phenomenon  constitutes another example of setting where post-hoc evaluation can be highly beneficial. By focusing on the discovered knowledge, note that two properties frequently studied in PDE discovery literature will not be considered in this article: the robustness (i)  to the presence of noise, and (ii) to scarcity in the training data. Both indeed characterize the performance of the discovery algorithm itself.

With this clarification in mind, we can now outline the main contributions of this paper: 
\begin{enumerate}
    \item  
Based on an {\bf extensive review} of a vast body of literature at the intersection of  machine learning, information theory, symbolic regression, but also by leveraging verification criteria from numerical analysis, \textcolor{black}{we establish that the overwhelming majority of papers on equation discovery do not address the full range of questions \textbf{Q1$\rightarrow$Q5} outlined above. More strikingly, most of them focus only on the prediction accuracy (\textbf{Q1}) and the recovery of the ground truth (\textbf{Q5}), using heterogeneous evaluation protocols and relying on a wide variety of metrics, each with its own specificities and limitations.} 
In light of this substantial diversity, we establish a {\bf taxonomy of PDE evaluation metrics}. We do not limit this classification to a mere enumeration of formulae. Instead, we provide a comprehensive analysis, examining for each of the metrics its {\bf properties, advantages}, and \textbf{limitations}. To the best of our knowledge, this is the first study which conducts such an in-depth analysis and gathers the state-of-the-art metrics previously dispersed across  different scientific domains. 
  To illustrate the practical use of these criteria,  we  rely throughout this study on a {\bf running example on the non linear fourth-order Kuramoto-Sivashinsky equation} (illustrated in Fig.~\ref{fig:evalPDE}), which presents a chaotic behavior and allows to model complex pattern formation and instabilities in various physical systems. The experiments performed with this non trivial equation (see all paragraphs entitled ``{\it Running Example}'' which constitute milestones of this paper) enable a better understanding of these metrics. 

\item \textcolor{black}{To highlight the risks of drawing overstated conclusions when considering only a subset of the properties that a discovered PDE should satisfy, we conduct a series of experiments using two PDE learning algorithms (PDE-FIND~\citep{rudy2016datadrivendiscoverypartialdifferential} and DeepMod~\citep{Both_2021}) on a set of equations (Burgers, Korteweg–de Vries, convection–diffusion and Kuramoto-Sivashinsky) covering a variety of characteristics (high-order  derivatives, nonlinearity, advection, steep gradients, etc.). As emphasized above, keeping in mind that the objective here is not to evaluate the learning algorithms themselves, the results provide evidence that a ranking obtained when relying solely on prediction-accuracy or GT-recovery metrics (which is the most common practice in the literature), can be completely reversed with metrics assessing extrapolation capabilities or compliance with physical properties.}
 
    \item  The analysis associated with our taxonomy, together with the results of the experimental study, allows us to {\bf formulate general cautions and recommendations}. Moreover, in order to address the pressing need for {\bf standardizing PDE evaluation} (as recently stated in \cite{lou2026datadrivendiscoverygoverningdifferential}) so as to avoid overstating the validity of  identified laws, we further propose an operational evaluation protocol, called {\sc Pde-Check}, consisting of simple, context-dependent rules to guide practitioners in their evaluation of discovered PDEs. This protocol depends on whether or not the GT equation is available (recovery or discovery task), whether data is available beyond the training time frame and/or under varying initial conditions, and whether physical knowledge is available. 
     Finally, we outline a {\bf set of promising research directions} to meet: (i) the need to better {\bf account for the functional properties} of the discovered equation, unlike ML-conventional pointwise metrics. We notably discuss the use of tools from functional analysis, as well as optimal transport and gradient flows; (ii) the necessity of \textbf{considering the numerical solvability} of the model as an additional key criterion to allow large and small-scale simulations when addressing real-world applications. This aspect is strikingly underexplored in the PDE evaluation literature, and (iii) the value of designing new measures  specifically tailored to the \textbf{evaluation of fractional PDEs}. Indeed, the ability of these equations to account for nonlocal effects makes them particularly attractive for modeling certain complex, multiscale, and memory-driven phenomena; however, the expressiveness and intrinsic complexity of fractional-order derivatives reduce the relevance of classical metrics. 
    \end{enumerate}
     We argue that this paper could provide valuable insights for machine learning experts developing new PDE discovery algorithms, and may also be of interest to scientists in other domains where modeling dynamics through equations plays a crucial role, {\it e.g.}, in physics, chemistry, or biology, seeking a comprehensive understanding of PDE evaluation.
      To enable them to easily compute the metrics presented throughout this survey on their own PDEs, we have developed the following user-friendly GitHub repository: \href{https://github.com/bapt-mat/PostHoc-PDE-Evaluation}{GitHub}. 

The rest of this article is organized as follows: \textcolor{black}{Section~\ref{sec:positionning} positions our work with respect to the PDE evaluation protocols introduced in the literature.} Section~\ref{sec:def} introduces the PDE discovery setting, the notations used, as well as the Kuramoto-Sivashinsky (KS) equation employed as  running example throughout our study. 
 Section~\ref{sec:GT} is devoted to the presentation of metrics that require the access to the governing equation $PDE_{\cal GT}$. Since this setting is 
 more appropriate for evaluating an algorithm’s capability to recover $PDE_{\cal GT}$ and less suited to scientific discovery, we relax this constraint in Section~\ref{sec:noGT} and present  metrics that can be applied even when the governing analytical form is not available.  
We also discuss various methods for evaluating the physical consistency of a PDE through the verification of relevant physical properties.
\textcolor{black}{We conduct a series of experiments in Section~\ref{sec:expePDEFIND-DEEpMoD} with PDE-FIND and DeepMod on Burgers, Korteweg–de Vries, convection–diffusion and Kuramoto-Sivashinsky equations, revealing   pathological behaviors when failing to evaluate PDEs across all five aspects covered by \textbf{Q1}$\rightarrow$\textbf{Q5}.}
We outline in Section~\ref{sec:guidelines} key recommendations along with the main pitfalls to be avoided when evaluating an equation. We also introduce {\sc Pde-Check}, the protocol we designed in the form of a set of rules that aims to promote standardization of PDE evaluation. 
We conclude this article in Section~\ref{sec:perspectives}  by sketching several promising directions for designing complementary metrics. 

\section{Positioning with respect to the PDE evaluation protocols in the literature} \label{sec:positionning}
To demonstrate the usefulness and scope of the taxonomy presented in this paper, followed by the recommendations derived from our experiments and the resulting evaluation protocol, we carried out an extensive review of equation discovery methods from the literature. We conducted an in-depth analysis of 36 state-of-the-art algorithms. For each method, we examined the metrics used in the evaluation protocols.
A summary of our analysis is provided in Table \ref{tab:sota2}. Each algorithm is characterized by its ability to address questions {\bf Q1}$\rightarrow${\bf Q5} (indicated with a green check mark {\color{ForestGreen}\ding{52}}), along with the equation number of the metrics used for evaluation. 

Our study led us to the following important conclusions: 
\paragraph{\textbf{- Non-holistic PDE evaluation:}} The two questions most commonly addressed in PDE discovery papers concern the ability to (i) accurately predict the solutions (\textbf{Q1}) and (ii) recover as accurately as possible the symbolic expression with the associated coefficients of the GT PDE (\textbf{Q5}), with 27 and 23 out of 36 papers, respectively, proposing metrics for this purpose. This is followed by the evaluation of sparsity (\textbf{Q3}), with more than one in three papers addressing this question. Although it may seem surprising at first glance that few papers use sparsity criteria in the post-hoc evaluation of PDEs, it is worth noting that many of them already incorporate sparsity into their optimization process (in the loss function, in the criteria for selecting the best candidates, etc.). This is particularly the case for symbolic regression (SR) methods and those based on sparse regression techniques. This explains why they do not necessarily reassess sparsity afterwards.
On the other hand, although these two questions are crucial for validating the discovery of a new physical law, \textbf{Q2} and \textbf{Q4} receive much less attention from PDE discovery algorithms. Yet, as demonstrated by the experiments conducted later in this paper with two algorithms (PDE-FIND~\citep{rudy2016datadrivendiscoverypartialdifferential} and DeepMod~\citep{Both_2021}), equations that are accurate in terms of prediction and symbolically similar to the ground truth, including in terms of their coefficients, may nevertheless correspond to physically inconsistent dynamics ({\it e.g.} violation of symmetry properties) and diverge over long time horizons or under changes in the initial conditions. As shown in Table \ref{tab:sota2}, only 7 methods over 36 address more than two questions, with a special mention for {\sc SymLang} \citep{baig2026symmetryconstrainedlanguageguidedprogramsynthesis}, which proposes metrics for each of the five questions.
\paragraph{\textbf{- High heterogeneity of evaluation metrics:}} We noted that there is substantial heterogeneity in the literature regarding the evaluation setups used for equation discovery. The most striking observation is that no fewer than 40  different metrics, each with its own specificities and limitations, as we will discuss later in our taxonomy, are employed to address the five key questions introduced above. This highlights the pressing need for a better understanding of these measures and for standardizing evaluation protocols.
\paragraph{\textbf{- Specialized evaluation protocols:}} The vast majority of papers do not actually propose a comprehensive evaluation protocol with rules,  checklist, or worked recipe to guide practitioners through the evaluation process. They rather propose a set (often a small one) of metrics without specifying an order of use depending on the scenario under consideration. Nevertheless, a few exceptions are worth mentioning here. ERBench \citep{kahlmeyer2026} starts from the postulate that the "\textit{most credible method for discovering new scientific laws is the one that most reliably recovers known GT equations}". For this reason, ERBench specializes in recovering the symbolic form and the true coefficients, justifying the use of multiple measures of similarity with the GT, including the \textit{Symbolic Recovery Rate \ref{eq:CAS}, Jaccard Index} \ref{eq:Jaccard}, and \textit{Tree Edit Distance} \ref{eq:TED}, each providing complementary information to the others. In the same vein, Weak SINDy~\citep{Messenger_2021} focuses on recovering the true PDE through several metrics that assess the algorithm’s ability to recover the symbolic form of the equation \ref{eq:TPR} and coefficient values close to those of the GT \ref{eq:nL2coef} \ref{eq:maxerror}. Nevertheless, as we will show in this paper, even a slight difference in the coefficients or functions involved in the equation can lead to substantially different long-term physical dynamics. The recently published method {\sc SymLang} \citep{baig2026symmetryconstrainedlanguageguidedprogramsynthesis} aims at evaluating identified equations on three complementary tiers: \textit{(i) Symbolic recovery, (ii) Extrapolation and physical consistency} and \textit{(iii) Partial observability}. Even though the metrics used (see \ref{eq:nMSE} \ref{eq:mdl-SymLang})   do not cover all aspects (coefficient errors are not considered, nor is spectral error, and the physical symmetry properties are imposed and assumed to be known), {\sc SymLang} is arguably the framework that most comprehensively addresses the evaluation requirements for a newly discovered knowledge. Finally, in PBS-PDE~\citep{ducos:hal-05128224}, the authors present an evaluation framework specifically dedicated to address the knowledge discovery for multiscale
modeling of biological signals. They compare  state-of-the-art PDE discovery methods according to their ability to recover a known equation. In particular, the notions of out-of-distribution extrapolation and preservation of physical properties are not addressed.

The present review of the state of the art in PDE evaluation highlights two main needs: (i) a systematic classification and analysis of the metrics proposed in the literature to understand the type of information they capture and their limitations, and (ii) a clear  protocol to guide practitioners in their post-hoc evaluation procedures. Without such a protocol, important aspects of the evaluation may be overlooked, potentially leading to overstated conclusions about the equations discovered. Addressing these needs is the objective of the following sections. We first introduce the notations used throughout this paper.

\begin{table}[t] 
\begin{small}
  \caption{\textcolor{black}{Positioning of the proposed five-question taxonomy with respect to the evaluation protocols used by the equation discovery algorithms in the literature (listed in alphabetical order). 
  A green check mark \cmark is displayed when the corresponding question is addressed, next to the equation number of the metric used in the corresponding paper and detailed in our taxonomy in Sections \ref{sec:GT} and \ref{sec:noGT}. On the other hand, \xmark means that the corresponding key question is not addressed in the referenced  paper.}}
 \label{tab:sota2}
  \centering
  \begin{tabular}{l|lllll}
    \toprule
         & \multicolumn{1}{c}{Solution} &  \multicolumn{1}{c}{Physical}     & \multicolumn{1}{c}{Interpretation} &  \multicolumn{1}{c}{Generalization}   &  \multicolumn{1}{c}{GT PDE}  \\
         & \multicolumn{1}{c}{Accuracy} & \multicolumn{1}{c}{Consistency} & \multicolumn{1}{c}{Sparsity} & \multicolumn{1}{c}{Capacity} & \multicolumn{1}{c}{Recovery}  \\
    Equation Discovery Methods & \multicolumn{1}{c}{{\bf Q1}} & \multicolumn{1}{c}{{\bf Q2}} & \multicolumn{1}{c}{{\bf Q3}} & \multicolumn{1}{c}{{\bf Q4}} & \multicolumn{1}{c}{{\bf Q5}} \\
    \midrule
    ABL-PDE \citep{ABLPDE2025Gao} & \cmark  \ref{eq:nMSE}      & \xmark & \xmark & \xmark  & \cmark \ref{eq:nL1coef} \\
    Adjoint-PDE-FIND~\citep{sadr2025datadriven}  & \xmark  &  \xmark &   \xmark & \xmark   & \cmark \ref{eq:TPR} \ref{eq:nL1coef}\\
    {\sc AI-Feynman}~\citep{udrescu2020aifeynmanphysicsinspiredmethod} & \cmark  \ref{eq:MDL-Feynman} &  \xmark &   \cmark \ref{eq:MDL-Feynman} & \xmark   & \xmark \\
    ANN-PYSR~\citep{zhang2025robustpdediscoverysparse} & \cmark \ref{eq:nMSE}        & \xmark & \xmark & \xmark  & \cmark \ref{eq:nL1coef}\\  
    AutoSINDy \citep{basiri2026discoverynonlineardynamicsautomated}  & \cmark  \ref{eq:MSE}       & \xmark & \cmark  \ref{eq:ExpTree}& \cmark   \ref{eq:n.utMSE} \ref{eq:rollout2} & \xmark \\  
DeepMod \citep{Both_2021}  & \cmark \ref{eq:MSE}       & \xmark & \xmark & \xmark  & \cmark \ref{eq:nL1coef}\\  
DISCOVER \cite{du2023discoverdeepidentificationsymbolically} & \cmark \ref{eq:reward2}    & \xmark & \cmark \ref{eq:reward2}& \xmark  & \xmark \\ 
    DHC-GEP  \citep{ma2024dimensionalhomogeneityconstrainedgene}   & \cmark  \ref{eq:MSE}      & \cmark {\tiny Dim. homog.} & \xmark & \xmark  & \cmark \ref{eq:L1coef}\\ 
    Ensemble-SINDY~\citep{Fasel_2022} & \cmark \ref{eq:MSE}   &  \xmark &   \xmark & \xmark   & \cmark \ref{eq:TPR} \ref{eq:L1coef}\\  
   ERBench  \citep{kahlmeyer2026} & \xmark        & \xmark & \xmark & \xmark  & \cmark \ref{eq:Jaccard} \ref{eq:TED} \ref{eq:CAS} \\  EqGPT~\citep{xu2025generativediscoverypartialdifferential} & \cmark \ref{eq:reward1}       & \xmark & \cmark \ref{eq:reward1} & \xmark  & \cmark \ref{eq:nepsilon1}\\
GNN~\citep{cranmer2020discoveringsymbolicmodelsdeep} & \cmark \ref{eq:Score}         & \xmark & \cmark \ref{eq:Score}& \xmark  & \xmark \\   
    iNeural-SINDy~\citep{forootani2023robustsindyapproachcombining}& \xmark        & \xmark & \xmark & \xmark  & \cmark \ref{eq:L1coef} \\
    KeplerAgent~\citep{yang2026thinklikescientistphysicsguided} & \cmark  \ref{eq:nMSE-bar}      & \xmark & \xmark & \cmark  & \cmark GPT-4o\\  
    LLM-SRBench~\citep{shojaee2025llmsrbenchnewbenchmarkscientific} & \cmark  \ref{eq:nMSE-bar}  &  \xmark &   \xmark & \xmark   & \cmark GPT-4o\\   MDBENCH~\citep{bideh2026mdbenchbenchmarkingdatadrivenmethods} & \cmark \ref{eq:nMSE}         & \xmark & \cmark \ref{eq:ExpTree}& \xmark  & \xmark \\  
    MIO-SINDY~\citep{bertsimas2022learningsparsenonlineardynamics} & \cmark \ref{eq:cAIC}       & \xmark & \xmark & \cmark \ref{eq:n.utMSE} \ref{eq:cAIC}  & \cmark \ref{eq:TPR} \ref{eq:nL2coef}\\
    NEXUS~\citep{Singh2025}  & \cmark \ref{eq:MSE}        & \cmark \ref{eq:dim-consist}& \cmark \ref{eq:ExpTreeV2} & \xmark  & \xmark \\  
      PBS-PDE~\citep{ducos:hal-05128224}  & \cmark \ref{eq:MSE}  & \xmark & \cmark \ref{eq:sparsity} & \xmark  & \cmark \ref{eq:nL2coef} \ref{eq:NDCG} \ref{eq:DCG}\\  
        PDE-BENCH \citep{PDEBench}  & \cmark \ref{eq:MSE} \ref{eq:nMSE} \ref{eq:mf_{MSE}}       & \xmark & \xmark & \xmark  & \xmark \\  
  PDE-FIND~\citep{rudy2016datadrivendiscoverypartialdifferential}& \xmark        & \xmark & \xmark & \xmark  & \cmark \ref{eq:nL1coef}\\     
      PDE-LEARN~\citep{STEPHANY2024106242}& \xmark        & \xmark & \xmark & \xmark  & \cmark \ref{eq:nL1coef}\\
    PDE-NET~\citep{long2018pdenetlearningpdesdata} & \xmark        & \xmark & \xmark & \cmark \ref{eq:nMSE-bar} & \cmark \ref{eq:L1coef}\\
     PDE-READ \citep{stephany2022pdereadhumanreadablepartialdifferential} & \xmark        & \xmark & \xmark & \xmark  & \cmark \ref{eq:nepsilon1} \\ 
     PDE-Refiner \citep{lippe2023pderefinerachievingaccuratelong} & \cmark \ref{eq:MSE}        & \xmark & \xmark & \cmark \ref{eq:rollout2} & \xmark \\  
    PITA \citep{zhu2025physicsinformed} & \cmark \ref{eq:nMSE}       & \xmark & \xmark & \xmark  & \xmark \\ 
    PhysPDE \citep{feng2025physpde} & \cmark \ref{eq:n.utMSE}        & \xmark & \xmark & \xmark  & \cmark \ref{eq:Precision} \ref{eq:Recall} \\ 
    {\sc ProbConserv}~\citep{Hansen2024}  & \cmark \ref{eq:MSE}       & \cmark {\tiny Conserv. laws} & \xmark & \xmark  & \xmark \\ 
    README \citep{lau2025readme}  & \cmark \ref{eq:nMSE-bar}       & \xmark & \cmark \ref{eq:sparsity}& \xmark  & \xmark \\ 
    SINDY~\citep{Brunton2016} & \cmark        & \xmark & \xmark & \xmark  & \xmark \\     
    SINDY-PI~\citep{Kaheman_2020} & \cmark    \ref{eq:n.utMSE}  & \xmark & \xmark & \xmark  & \cmark \ref{eq:TPR} \ref{eq:L1coef}\\  
    SRBench \citep{Imai_Aldeia_2025} & \cmark \ref{eq:nMSE-bar}        & \xmark & \cmark \ref{eq:ExpTree} & \xmark  & \xmark \\  
SRSD \citep{matsubara2024} & \cmark \ref{eq:nMAE}         & \xmark & \xmark & \xmark  & \cmark \ref{eq:TED} \\  
{\sc SymLang} \citep{baig2026symmetryconstrainedlanguageguidedprogramsynthesis}  & \cmark   \ref{eq:nMSE} \ref{eq:mdl-SymLang}    & \cmark {\tiny Sym. violation} & \cmark \ref{eq:mdl-SymLang} & \cmark \ref{eq:nMSE} & \cmark \\ 
Unrolled-SINDY \citep{banna2025unrolledsindystableexplicitmethod} & \xmark        & \xmark & \xmark & \xmark  & \cmark \ref{eq:nL1coef}\\ 
Weak SINDy~\citep{Messenger_2021} & \xmark        & \xmark & \xmark & \xmark  & \cmark  \ref{eq:TPR}\ref{eq:nL2coef} \ref{eq:maxerror}\\ 
    \bottomrule
  \end{tabular}
\end{small}
\end{table}

\section{PDE discovery setting, notations and experimental setup}
\label{sec:def}
 Let us consider equations of the following general form: 
$u_t={\cal N}[u],$ 
with $u\in C^n([0,T]\times\Omega;\mathbb{R}^{d_2})$, an $n$-times continuously differentiable function modeling  a vector field; 
$u_t$ is the time derivative, and ${\cal N}[\cdot]$ is a (possibly nonlinear) differential operator,  
with $t \in [0,T]$ the temporal variable bounded by a finite time horizon $T$ and $\mathbf{x} \in \Omega \subset \mathbb{R}^d$ the spatial coordinates belonging to a bounded input domain. The equations may also be accompanied by initial and boundary conditions (denoted by IC and BC, respectively) that must be satisfied at the initial time or on the boundaries of the domain.  In the following,  the notation $u_{\mathbf{x}}$ denotes the partial derivative with respect to the input spatial domain, $u_{\mathbf{xx}}$ corresponds to the second order derivative, etc. 

The PDE discovery paradigm is based on the realistic assumption that in most physical laws only a few important terms govern the underlying dynamics, that encourages promoting sparsity. 
For this reason, many algorithms address the problem from a sparse  regression perspective by approximating  the differential operator into the form of a parsimonious linear combination of preselected terms from a finite dictionary, typically composed of partial space derivatives ({\it i.e.} $u_{\mathbf{x}_1}$, $u_{\mathbf{x_1x_1}}$, $u_{\mathbf{x_1x_2}}$, $u_{\mathbf{x_1}}u_{\mathbf{x_2}}$, etc.), polynomials ({\it e.g.} $u^3$) and constants. 
Since the success of such strategy is closely tied to the completeness of the dictionary, the design of the latter remains a challenge. This is why free-form strategies are used in symbolic regression ({\it e.g.}, resorting to genetic programming, tree-based search algorithms, transformers, etc.) to adapt the expression being searched for to the task at hand. Nevertheless, without loss of generality, we can assume here that PDE discovery can be reduced to producing a linear combination of  terms in the form of an estimate $\hat{u}_t=\bm{\hat{\alpha}}^\top.\Theta$ where  $\Theta$ is a vector of size $|\Theta|$
whose components are evaluations of linear and non linear functions at $(t,\mathbf{x})$ 
 and $\bm{\hat{\alpha}}$ is a $|\Theta| \times d_2$ matrix containing the learned coefficients associated with each term of $\Theta$ for the $d_2$ governing equations. Let us assume that $\bm{\hat{\alpha}}$ has been learned with some PDE discovery algorithm from a training set ${\cal T}=\{u(t_i,\mathbf{x}_i)\}_{i=1}^{n_{\cal T}}$  
composed of $n_{\cal T}$ {\bf simulation data} in the scenario where the ground truth equation $PDE_{\cal GT}$ is known, or {\bf observation data} when the theory governing the dynamical system is not yet identified. Moreover, we also assume we have at our disposal a validation set ${\cal V}=\{u(t_j,\mathbf{x}_j)\}_{j=1}^{n_{\cal V}}$  that will be used, depending on the type of metrics, for evaluating the identified PDEs. Note that ${\cal V}$ might be obtained from different IC and BC and beyond the training time horizon to evaluate the capacity of the learned PDEs to adapt to out-of-distribution tasks. 
For the sake of simplifying the notations, we  assume in the following that $d_2=1$, \textit{i.e.}  $u$ is a scalar field. Thus, $\bm{\hat{\alpha}}^\top=(\bm{\hat{\alpha}}_1,\ldots, \bm{\hat{\alpha}}_i, \ldots, \bm{\hat{\alpha}}_{|\Theta|})$ becomes a  vector of size $|\Theta|$.  However, note that most of the metrics presented in this paper can be readily applied to any $d_2$-dimensional dynamics, \textit{e.g.} by  using the Frobenius norm instead of the $\ell_p$-norm.\\

\begin{figure}[t]
    \centering
    \includegraphics[width=0.29\linewidth]{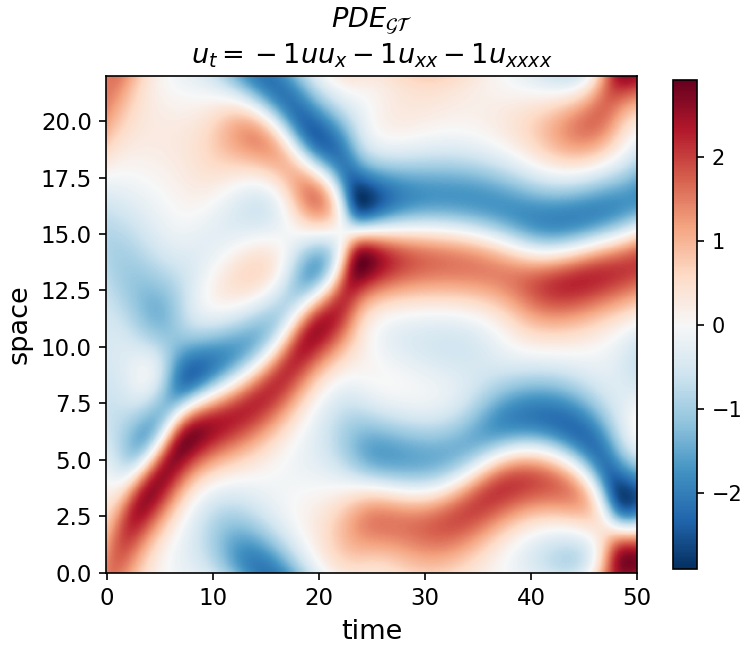}
    \includegraphics[width=0.29\linewidth]{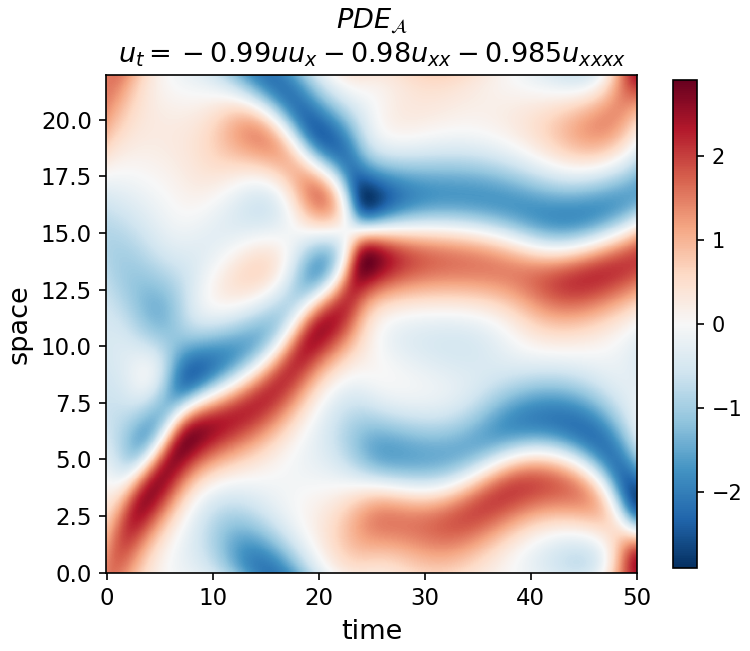}
    \includegraphics[width=0.28\linewidth]{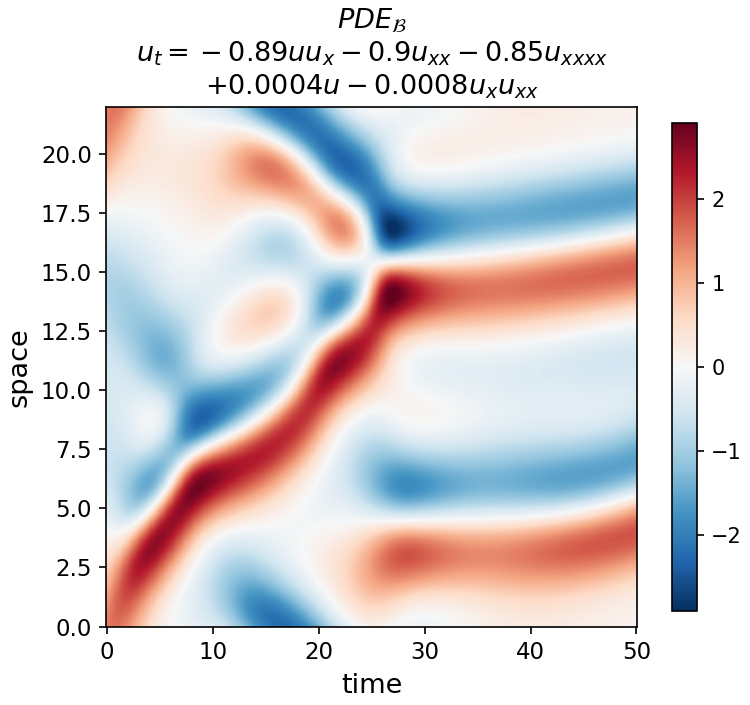}
    \caption{Simulated solutions of the Kuramoto-Sivashinsky equation $PDE_{\cal GT}$ (left) and 2 learned  equations, $PDE_{\cal A}$ (center) and $PDE_{\cal B}$ (right). For the purpose of this running example,  $PDE_{\cal A}$  is by construction much more similar to the ground truth than  $PDE_{\cal B}$. Indeed, unlike $PDE_{\cal A}$, we can observe  that $PDE_{\cal B}$ gradually diverges over time  (especially after $t>30$) from the true physical dynamics modeled by $PDE_{\cal GT}$.} \label{fig:3PDE-KS}
\end{figure}

\noindent {\bf \underline{Running Example:}} As mentioned above, we  consider the following fourth-order non linear 1D Kuramoto-Sivashinsky (KS) PDE, which will serve as running example throughout the paper: 
\begin{eqnarray}
PDE_{\cal GT}: u_t=-1.0uu_{\mathbf{x}}-1.0u_{\mathbf{xx}}-1.0u_{\mathbf{xxxx}}.
\label{eq:KS}
\end{eqnarray} 
This equation will be solved using periodic BC and the following IC: $\cos(2\pi x/L) + 0.5\cos(4\pi x/L + 0.3)$.  
Let us assume that  an algorithm ${\cal A}$  discovered from a simulation data set ${\cal T}$ the following equation:
\begin{eqnarray}
PDE_{\cal A}: {\hat u}_t=-0.99uu_{\mathbf{x}}-0.98u_{\mathbf{xx}}-0.985u_{\mathbf{xxxx}}.
\label{eq:PDEA}
\end{eqnarray} 
Let us further assume that another algorithm ${\cal B}$ identified a second equation defined as: 
\begin{eqnarray}
PDE_{\cal B}: {\hat u}_t=-0.89uu_{\mathbf{x}}-0.9u_{\mathbf{xx}}-0.85u_{\mathbf{xxxx}} + 0.0004u-0.0008u_\mathbf{x}u_\mathbf{xx}.
\label{eq:PDEB}
\end{eqnarray} 
Note that we intentionally use as $PDE_{\cal B}$ an equation that is ``farther''\footnote{Although this term remains ambiguous at this stage of the paper, it refers quite objectively to deviations in both the coefficients and the functions involved.} from  $PDE_{\cal GT}$ than $PDE_{\cal A}$ leading to larger differences between solutions, especially as time progresses  (see simulations in Fig.~\ref{fig:3PDE-KS}), thereby allowing us to assess the ability of the metrics to distinguish the quality of the two equations.
Using our notations, the true vector of coefficients $\bm{\alpha}^\top=(0,-1,-1,-1,0)$, while the  vectors learned by ${\cal A}$ and ${\cal B}$ are $\bm{\hat{\alpha}}_{\cal A}^\top=(0,-0.99,-0.98,-0.985,0)$ and $\bm{\hat{\alpha}}_{\cal B}^\top=(0.0004,-0.89,-0.9,-0.85,-0.0008)$ respectively, all of them associated with $\Theta^\top=(u,uu_{\mathbf{x}}, u_{\mathbf{xx}},u_{\mathbf{xxxx}}, u_\mathbf{x}u_\mathbf{xx})$. \\

\noindent {\bf Experimental setup:} For the experiments requiring validation data ${\cal V}$ simulated from $PDE_{\cal GT}$ and estimated solutions ${\cal E}$ obtained from $PDE_{\cal A}$ and $PDE_{\cal B}$  (this concerns all metrics discussed in Sec.~\ref{sec:noGT}), we will use the explicit numerical time-integration scheme ETD$1$ (Exponential Time Differencing of order 1) and solve the equations over a time $T=50$ with a time step $\delta_t = 0.05$ yielding a temporal discretization made of $N_t =1000$ steps.
We use a spatial mesh  $N_{\mathbf{x}}=1024$ over a periodic domain of length $L=22$, which yields a spatial spacing  $\delta_x = L/N_{\mathbf{x}} \approx 0.0215$. This way, we  collect  datasets  ${\cal V}$ and ${\cal E}$ each consisting of $n_{\cal V}=N_{\mathbf{x}} \times N_t = 1,024,000 $ data, defined as solutions evaluated at the same $(t,\mathbf{x})$ coordinates allowing pointwise comparisons. 
 Depending on the setting (in-distribution or out-of-distribution), the sets  ${\cal V}$ and ${\cal E}$ will be simulated using either the same or different IC and BC as ${\cal T}$.

\section{Metrics requiring access to the governing PDE} \label{sec:GT}

We begin this review with criteria that require access to the analytical expression of the governing law $PDE_{\cal GT}$. Although this category of PDE evaluation metrics may appear somewhat contradictory at first sight (why allow the use of $PDE_{\cal GT}$ when the goal is to discover it?), it is worth noticing that most papers presenting a new PDE discovery algorithm assume in their experimental setup that the target equation is known; and they directly use the latter in the evaluation.  This way, rather than focusing on assessing a \textbf{discovered PDE} as a new physical theory, these works primarily aim to  evaluate the algorithm’s ability to  \textbf{recover a known equation}  used for proof-of-concept purposes. Under this scenario, several metrics can be employed to check  the accuracy of either the learned symbolic expression (Sec.~\ref{sec:terms}) or the coefficients associated with its terms (Sec.~\ref{sec:coef}). 


\subsection{Recovery of the PDE terms} \label{sec:terms}
Using our notations, let us recall that PDE discovery  looks for equations of the form $\hat{u}_t=\bm{\hat{\alpha}}^\top.\Theta$.   Metrics discussed in this section are based on the functions involved in this analytical expression, {\it i.e.} those associated with coefficients $\bm{{\alpha}}_i \neq 0$. 
They primarily help answer Question {\bf Q5} (which concerns the recovery of $PDE_{\cal GT}$). 
 Additionally, they indirectly provide insights on Question {\bf Q3} (dealing with the sparsity).  \\

\noindent {\bf Support recovery}: In \cite{Lagergren_2020}, directly inspired by  the well-known $F_1$-score, the authors leverage the following {\it true positive ratio} ($TPR$):
\begin{eqnarray}
TPR(\bm{\alpha},\bm{\hat{\alpha}})=\frac{TP(\bm{\alpha},\bm{\hat{\alpha}})}{TP(\bm{\alpha},\bm{\hat{\alpha}})+FN(\bm{\alpha},\bm{\hat{\alpha}})+FP(\bm{\alpha},\bm{\hat{\alpha}})} \in [0,1], \label{eq:TPR}
\end{eqnarray}
where $TP$ (\textit{True Positives}) denotes the number of correctly-identified nonzero coefficients, and $FN$ (\textit{False Negatives}) - resp. $FP$ (\textit{False Positives}), stands for the number of coefficients that are incorrectly specified as zero (resp. nonzero). This metric has been used in several PDE discovery methods, including the Mechanistic PDE Networks~\citep{pervez2025mechanisticpdenetworksdiscovery}, Weak SINDy~\citep{Messenger_2021} or Adjoint-PDE-FIND~\citep{sadr2025datadriven}. Rather than using $TPR$ as a measure of overall structural similarity, one can exploit the so-called Precision $P$ and Recall $R$. This is the post-hoc evaluation strategy used in PhysPDE \citep{feng2025physpde}, an algorithm guiding and constraining the discovery of PDE expressions with prior physical hypotheses. $P$ and $R$ are respectively defined in the classical way as follows:
\begin{eqnarray}   P(\bm{\alpha},\bm{\hat{\alpha}})=\frac{TP(\bm{\alpha},\bm{\hat{\alpha}})}{TP(\bm{\alpha},\bm{\hat{\alpha}})+FP(\bm{\alpha},\bm{\hat{\alpha}})}  \in [0,1],   \label{eq:Precision}
\end{eqnarray}
\begin{eqnarray} 
R(\bm{\alpha},\bm{\hat{\alpha}})=\frac{TP(\bm{\alpha},\bm{\hat{\alpha}})}{TP(\bm{\alpha},\bm{\hat{\alpha}})+FN(\bm{\alpha},\bm{\hat{\alpha}})}  \in [0,1].   \label{eq:Recall}
\end{eqnarray}
In PDE evaluation, combining $P$ and $R$ instead of a single $TPR$-like measure provides a more balanced assessment by jointly capturing how well the discovered differential terms match the true underlying ones while penalizing both missing relevant differential terms and introducing spurious ones.

Note that $TPR(\bm{\alpha},\bm{\hat{\alpha}})$ (as defined in Eq.~\ref{eq:TPR}) is also referred in the literature to as Jaccard index $J(\bm{\alpha},\bm{\hat{\alpha}})$ (see  \cite{JMLR:v18:17-514} and MIO-SINDY~\citep{bertsimas2022learningsparsenonlineardynamics}) or as Tanimoto coefficient $T(\bm{\alpha},\bm{\hat{\alpha}})$ \citep{tripp2023tanimotorandomfeaturesscalable}, both defined as the following function on sets:
\begin{eqnarray}
J(\bm{\alpha},\bm{\hat{\alpha}})=T(\bm{\alpha},\bm{\hat{\alpha}})=\frac{|{\cal S}(\bm{\alpha}) \cap {\cal S}(\bm{\hat{\alpha}})|}{|{\cal S}(\bm{\alpha}) \cup {\cal S}(\bm{\hat{\alpha}})|} \in [0,1],
    \label{eq:Jaccard}
\end{eqnarray}
where ${\cal S}(\bm{{\alpha}})=\{ \Theta_i : \bm{{\alpha}}_i \neq 0 \}$ and ${\cal S}(\bm{\hat{\alpha}})=\{ \Theta_i : \bm{\hat{\alpha}}_i \neq 0 \}$. \\ 

Although $TPR$ (and hence $J$ and $T$) are not specifically designed to evaluate the sparsity of the learned model, values close to 1 indicate that the structure of $PDE_{\cal GT}$ has been successfully recovered. As it is generally accepted that most physical laws can be expressed with a small number of terms, a $TPR$ close to 1 means that the equation is therefore also likely to be parsimonious. These metric can therefore also be considered as providing a partial answer to  Question {\bf Q3} regarding sparsity.\\

In SRBench \citep{kahlmeyer2026}, the Tree Edit Distance between the expression trees of the GT and discovered differential operators  is leveraged to define the following normalized $TED$ score: 
\begin{eqnarray}
    TED(ExpTree({\cal N}[u]),ExpTree(\bm{\hat{\alpha}}^\top.\Theta)) = \frac{dist(ExpTree(\bm{\hat{\alpha}}^\top.\Theta ),ExpTree({\cal N}[u]))} {size(ExpTree(\bm{\hat{\alpha}}^\top.\Theta)+size(ExpTree({\cal N}[u]))} \in [0,1],
    \label{eq:TED}
\end{eqnarray}

where $dist(.,.)$ is the Tree Edit Distance, and $size(.)$ outputs, {\it e.g.} using the {\tt SymPy} Python library, the number of elements involved in an expression tree $ExpTree$. Note that the tree edit distance is also used as evaluation metric in SRSD \citep{matsubara2024}.

\noindent {\bf Symbolic accuracy}: Based on a paradigm that differs significantly from other classical PDE discovery  methods, Foundation Models, and in particular Large Language Models (LLMs), have recently gained interest for equation identification (see, {\it e.g.} NeuroSymBO \citep{qu2025dynamicbayesianoptimizationframework}, Data-Driven Discovery (D3) framework~\citep{NEURIPS2024_aea8bdc4} or \cite{moralesalvarado2025foundationmodelsequationdiscovery}). In LLM-SRBench~\citep{shojaee2025llmsrbenchnewbenchmarkscientific}, beyond building many challenging problems across various scientific domains, the authors employ  GPT-4o as an automated post-hoc evaluator to measure a  {\bf symbolic accuracy} between 
$PDE_{\cal GT}$ and a discovered expression, after replacing constants with placeholders. The LLM is  prompted to decide if the two equations are symbolically equivalent (with a request for algebraic justification) under a tolerance of small numerical mismatches in the coefficients. A  prompt similar to the following is used: {\it \enquote{Given the $PDE_{\cal GT}$  expression and the hypothesis $PDE_{\cal A}$, determine if there exist any constant parameter values that would
make $PDE_{\cal GT}$ equivalent to $PDE_{\cal A}$}}. Note that LLMs, as used in LLM-SRBench to prune the search space in the context of SR, are combined in \cite{saveliev2026LLM} with a utility measure for each term $\theta_i$ of the equation and its associated coefficients $\bm{{\alpha}}_i$  (while
holding the others fixed) to refine the list of candidate functions. 

A symbolic accuracy is also computed in ERBench \citep{kahlmeyer2026}. The authors use a Computer Algebra System (CAS) like Sympy that can be used to evaluate if the learned differential operator $\bm{\hat{\alpha}}^\top.\Theta$ is symbolically equivalent to that of the GT (${\cal N}[u]$). Formally, let us define $SRecov_1(\bm{\alpha},\bm{\hat{\alpha}}) =\bm{\hat{\alpha}}^\top.\Theta-{\cal N}[u]$ and $SRecov_2(\bm{\alpha},\bm{\hat{\alpha}})=\frac{\bm{\hat{\alpha}}^\top.\Theta}{{\cal N}[u]}$. One states that two differential operators are symbolically equivalent if a CAS can verify that either:
\begin{eqnarray}
    SRecov_1(\bm{\alpha},\bm{\hat{\alpha}}) =c_0 \mbox{ or } SRecov_2(\bm{\alpha},\bm{\hat{\alpha}})=c_1, \label{eq:CAS}
\end{eqnarray}
for some constants $c_0 \in \mathbb{R}$ or $c_1 \in \mathbb{R} \backslash \{0\}.$ The authors compute a Symbolic
Recovery Rate as the fraction of several test problems for which either of these conditions is met.\\

\begin{table}[t]
  \caption{Application of the term recovery  metrics (Sec.~\ref{sec:terms}) to compare $PDE_{\cal A}$ vs $PDE_{\cal B}$ using KS $PDE_{\cal GT}$ as reference. A green point \hspace{-0.2cm} \raisebox{-0.1ex}\greenpoint  \hspace{-0.2cm} (resp. \hspace{-0.4cm} \raisebox{-0.1ex}\redpoint \hspace{-0.2cm}) corresponds to a better  (resp. worse) evaluation score w.r.t. $PDE_{\cal GT}$. By construction of the equations, $PDE_{\cal A}$ is expected to achieve better  scores. An \hspace{-0.3cm} \raisebox{-0.1ex}\orangepoint \hspace{-0.2cm} point corresponds to a tie situation, where the metric is unable to distinguish between the two PDEs.}
  \label{tab:res-terms}
  \centering
  \begin{tabular}{|r|c|c|}
    \hline
        & \multicolumn{1}{|c|}{$PDE_{\cal A}$} &  \multicolumn{1}{|c|}{$PDE_{\cal B}$}  \\
     & \multicolumn{1}{|c|}{{\scriptsize ${\hat u}_t=-0.99uu_{\mathbf{x}}$}} & \multicolumn{1}{|c|}{{\scriptsize ${\hat u}_t=-0.89uu_{\mathbf{x}}-0.9u_{\mathbf{xx}}$}}\\
    &  \multicolumn{1}{|c|}{{\scriptsize $-0.98u_{\mathbf{xx}}-0.985u_{\mathbf{xxxx}}$}} & \multicolumn{1}{|c|}{{\scriptsize $-0.85u_{\mathbf{xxxx}} + 0.0004u-0.0008u_\mathbf{x}u_\mathbf{xx}$}} \\
     \hline  $TPR(\bm{\alpha},\bm{\hat{\alpha}})$ \ref{eq:TPR}  & 1 \greenpoint &   $3/5$ \redpoint \\
    \hline
    $P(\bm{\alpha},\bm{\hat{\alpha}})$ \ref{eq:Precision} & 1 \greenpoint &  $3/5$ \redpoint \\
    \hline
    $R(\bm{\alpha},\bm{\hat{\alpha}})$ \ref{eq:Recall} & 1  \orangepoint &  \hspace{0.2cm} 1  \hspace{0.04cm}\orangepoint \\
  \hline
    $TED~  \ref{eq:TED}$ & \hspace{-0.55cm} 0.14  \greenpoint &  \hspace{-0.25cm} 0.38  \hspace{0.04cm}\redpoint \\ 
 \hline
    $SRecov_1(\bm{\alpha},\bm{\hat{\alpha}})$ \ref{eq:CAS} & 0  \orangepoint &  \hspace{0.2cm} 0  \hspace{0.04cm}\orangepoint \\ 
    \hline
    $LLM(\bm{\alpha},\bm{\hat{\alpha}})$  & \parbox{4cm}{\vspace{0.1cm}{\it \small \enquote{They are  approximately equivalent, differing only by small numerical estimation errors.}\\}} \greenpoint &  \parbox{6cm}{\vspace{0.1cm}{\it \small \enquote{They are structurally similar, the two extra terms are small and likely spurious, but there is no set of constant parameter values that makes the two equivalent.}}}  \redpoint \\
    \hline
  \end{tabular}
\end{table}

\noindent {\bf \underline{Running Example:}} Let us perform experiments with the metrics presented in this section on the KS equation ($PDE_{\cal GT}$) and the learned equations $PDE_{\cal A}$ and $PDE_{\cal B}$.  
 The values of the metrics are reported in Table~\ref{tab:res-terms}.   
For $PDE_{\cal A}$, from $TP(\bm{\alpha},\bm{\hat{\alpha}_{\cal A}})=3$, $FN(\bm{\alpha},\bm{\hat{\alpha}_{\cal A}})=0$ and $FP(\bm{\alpha},\bm{\hat{\alpha}_{\cal A}})=0$, we get  $TPR(\bm{\alpha},\bm{\hat{\alpha}_{\cal A}})=1$, $P(\bm{\alpha},\bm{\hat{\alpha}_{\cal A}})=1$ and $R(\bm{\alpha},\bm{\hat{\alpha}_{\cal A}})=1$, reflecting that the structure of $PDE_{\cal GT}$ has been perfectly recovered.  On the other hand, from $TP(\bm{\alpha},\bm{\hat{\alpha}_{\cal B}})=3$, $FN(\bm{\alpha},\bm{\hat{\alpha}_{\cal B}})=0$ and $FP(\bm{\alpha},\bm{\hat{\alpha}_{\cal B}})=2$ we get for $PDE_{\cal B}$, as expected,  a lower $TPR(\bm{\alpha},\bm{\hat{\alpha}_{\cal B}})$ equal to $3/5$ with $P(\bm{\alpha},\bm{\hat{\alpha}_{\cal B}})=3/5$ and $R(\bm{\alpha},\bm{\hat{\alpha}_{\cal B}})=1$. Different from 1, $TPR$ and $P$ capture well the presence of spurious terms. 
 Using LLMs, we get a better evaluation for $PDE_{\cal A}$ than $PDE_{\cal B}$. The model further indicates that with the presence of the two additional terms $u$ and $u_\mathbf{x}u_\mathbf{xx}$, there is no set of coefficients $\bm{\hat{\alpha}_{\cal B}}$ that makes the two equations $PDE_{\cal B}$ and $PDE_{\cal GT}$ equivalent.\\

\noindent    \fbox{\textbf{PROS} \cmark vs \textbf{CONS} \xmark}
\vspace{0.3cm}

\noindent \cmark $TPR,P$ and $R$ directly measure success of structural recovery. Moreover, they are invariant to the number of terms of the equations and their interpretation is intuitive (the higher the better). Although they do not directly measure the sparsity of the discovered PDEs, a value close to 1 indicates that the parsimony of $PDE_{\cal GT}$ is recovered. LLMs have the benefit of justifying the decision and provide a binary final answer. $SRecov$ looks for symbolic equivalence between different PDEs.
\vspace{0.3cm}

\noindent \xmark These metrics require $PDE_{\cal GT}$ to be known. Moreover, $TPR,P$ and $R$  only assess the form of the PDE  without evaluating the associated coefficients which have an impact on the shape of the dynamics. Unlike $SRecov$, the symbolic accuracy provided by a LLM is not a  quantitative measure. Therefore, two close but different textual descriptions can  make it difficult to compare  competing PDEs. $TED$ is used in symbolic regression where the goal is to discover explicit mathematical expressions, typically without involving partial derivatives. Therefore, assuming a unit cost for each edit operation as done in SRBench \citep{kahlmeyer2026} is not relevant when it comes to comparing PDEs. Also note that, although $TED$ could treat coefficients as nodes of the expression tree, doing so is not advisable, since even a slight difference in coefficient values would incur an edit cost of 1.

\begin{center}
\rule{0.1\textwidth}{1pt} {\raisebox{-0.4ex}{\cmark \xmark}}
\rule{0.1\textwidth}{1pt}
\end{center}

\subsection{Recovery of the PDE coefficients} \label{sec:coef}

This section  provides an overview  of the main existing metrics built upon the magnitude of the learned coefficients $\bm{\hat{\alpha}}$. 
All of them primarily help answer Question {\bf Q5}. 
\\ 

\noindent {\bf  Unweighted error on coefficients}: Most PDE discovery papers  quantify the deviation $\epsilon^p_{coef}$ between the learned PDE and $PDE_{\cal GT}$ in terms of {\bf error on coefficients}. Often defined as an unweighted error, $\epsilon^p_{coef}$ can take different forms depending on the norm $\ell_p$ used for  the comparison between $\bm{{\alpha}}$ and $\bm{\hat{\alpha}}$. 

For instance, 
one can leverage the classic normalized squared error $\epsilon^2_{coef}(\bm{\alpha},\bm{\hat{\alpha}})$ (as used in Weak SINDy~\citep{Messenger_2021},  Ensemble-SINDY~\citep{Fasel_2022} or MIO-SINDY~\citep{bertsimas2022learningsparsenonlineardynamics})  which gives insights regarding the magnitude of spurious coefficients:
\begin{eqnarray}
    \epsilon^2_{coef}(\bm{\alpha},\bm{\hat{\alpha}}) =
    \frac{||\bm{{\alpha}}-\bm{{\hat{\alpha}}} ||_2^2}{|| \bm{{\alpha}}||^2_2} \in \mathbb{R}_{\geq 0}.
    \label{eq:nL2coef}
\end{eqnarray}
On the other hand, the  normalized absolute error $\epsilon^1_{coef}$  is  preferred over $\epsilon^2_{coef}$ (which may over-penalize an error on only one coefficient) in several articles on PDE discovery (see, {\it e.g.} SINDY~\citep{Brunton2016}, PDE-NET~\citep{long2018pdenetlearningpdesdata}, PDE-LEARN~\citep{STEPHANY2024106242}, EqGPT~\citep{xu2025generativediscoverypartialdifferential}, Adjoint-PDE-FIND~\citep{sadr2025datadriven} or ABL-PDE \citep{ABLPDE2025Gao}). 
It is defined as follows:
\begin{eqnarray}
    \epsilon^1_{coef}(\bm{\alpha},\bm{\hat{\alpha}}) = 
    \frac{||\bm{{\alpha}}-\bm{{\hat{\alpha}}} ||_1}{|| \bm{{\alpha}}||_1} \in \mathbb{R}_{\geq 0}.
    \label{eq:nL1coef}
\end{eqnarray}
Based on the physical observation that a single misidentified coefficient can completely alter the law, Weak SINDy~\citep{Messenger_2021}, and  Mechanistic PDE Networks~\citep{pervez2025mechanisticpdenetworksdiscovery} resort to the more severe $\ell_{\infty}$-based error on the true nonzero coefficients:
\begin{eqnarray}
    \epsilon_{coef}^\infty(\bm{\alpha},\bm{\hat{\alpha}}) = \max_{i:\bm{\alpha_i}\neq0}\frac{|\bm{{\alpha}}_i-\bm{{\hat{\alpha}}}_i|}{|\bm{\alpha_i}|} \in \mathbb{R}_{\geq 0}.
    \label{eq:maxerror}
\end{eqnarray}
Finally, in iNeural-SINDy~\citep{forootani2023robustsindyapproachcombining}, the absolute values are  applied on each coefficient $\bm{\alpha_i}$ separately in order to take into account the fact that their respective order might be different, describing dynamics of varying scales:
\begin{eqnarray}
    \epsilon^1_{coef(i)}(\bm{\alpha_i},\bm{\hat{\alpha}_i}) =  |\bm{{\alpha}}_i-\bm{{\hat{\alpha}}}_i| \in \mathbb{R}_{\geq 0}, \hspace{0.5cm} i=1..|\Theta|.
    \label{eq:L1coef}
\end{eqnarray}

Considering that, depending on the PDE term it is associated with, a small variation in a coefficient may have a more or less significant impact on the overall dynamics, we discuss below metrics built upon weighted errors.\\

\noindent {\bf Weighted error based on sensitivity analysis}: Metrics presented above do not distinguish between coefficients that are numerically close but dynamically different, and those that are numerically different but have negligible effect on the solution behavior, stability properties, and long-term evolution. One way to overcome this issue consists in assigning a weight $w_i$ to each coefficient $\bm{\alpha_i}$ according to the impact of the associated function $\theta_i$ on the induced dynamics. Note that this strategy requires simulation data from $PDE_{\cal GT}$ to allow the calculation of partial derivatives, as explained in the following.  

One can define $w_i$ on the basis of a local sensitivity analysis inspired by the Learning Gradient Flow (LGF) optimizer \citep{norman2026learninggradientflowusing}, as follows:
\[
w_i \propto \left\| \frac{\partial u}{\partial \bm{{\alpha}}_i} \right\|_{2}^2,
\]
leading to the following weighted error reflecting how perturbations in each coefficient $\bm{{\alpha}}_i$ propagate through the dynamics  (note that this can be also applied to the $\ell_1$-based variant):
\begin{eqnarray}
    w\epsilon^2_{coef}(\bm{\alpha},\bm{\hat{\alpha}}) = \frac{\sum_{i=1}^{|\Theta|} w_i(\bm{{\alpha}}_i-\bm{{\hat{\alpha}}}_i)^2}{\sum_{i=1}^{|\Theta|}w_i} \in \mathbb{R}_{\geq 0}.
    \label{eq:wL2coef}
\end{eqnarray}

\noindent {\bf Rank on the coefficients}: Another category of metrics  consists in evaluating  the discovered PDE using the rank of the terms, as used in \cite{ducos:hal-05128224}. The absolute values of the coefficients are sorted in decreasing order; then, the  ranking vector of the learned PDE  is compared with that of
$PDE_{\cal GT}$. This way, the metric assesses the ability of the method to
correctly prioritize from a coefficient perspective the key terms within the PDE. To compare the ranking vectors $rank(\bm{\alpha})$ and $rank(\bm{\hat \alpha})$, one can resort, {\it e.g.}, to the normalized Discounted Cumulative Gain $NDCG$:
\begin{eqnarray}
NDCG(\bm{\alpha},\bm{\hat{\alpha}})=\frac{DCG(rank(\bm{\hat \alpha}))}{DCG(rank(\bm{{ \alpha}}))} \in \mathbb{R}_{\geq 0}.
    \label{eq:NDCG}
\end{eqnarray}
where $DCG$ is  defined in~\cite{jarvelin2002cumulated} as follows:
\begin{eqnarray}
    DCG(rank(\bm{\alpha})) = \sum_{i=1}^{|\Theta|} \frac{rank(\bm{{\alpha}})_i}{log_2(i+1)} \in \mathbb{R}_{\geq 0}.
    \label{eq:DCG}
\end{eqnarray}
Even though, when equal to 1, this metric does not guarantee that $PDE_{\cal GT}$ has been recovered, it nevertheless ensures that the most important local terms at the origin of the physical dynamics are correct, and therefore $NDCG$ partially also answers Question {\bf Q2} about the physical consistency in the sense that the main coefficients governing the PDE are correctly recovered.\\

\noindent {\bf Towards jointly considering  terms and coefficients:} Although  a combination of the two may appear natural, it is worth noticing that there is no adopted  metric in the related works that explicitly merges the support recovery (Sec.~\ref{sec:terms}) {\bf and} coefficient error into a single canonical formulation.
We hypothesize that one reason could be that, since these metrics cannot be used during training (because $PDE_{\cal GT}$ is assumed to be unknown and, moreover, $TPR$ is non-separable, thus not optimizable), they are instead employed independently in a post-hoc manner, without the  need to tune an additional hyperparameter to combine them. However, note that two criteria attempt to establish a link between the two and therefore deserve to be mentioned here. 

\begin{table}[t]
  \caption{Application of the coefficient recovery  metrics (Sec.~\ref{sec:coef}) to compare $PDE_{\cal A}$ versus $PDE_{\cal B}$ using the KS ground truth equation $PDE_{\cal GT}$ as reference. } 
  \label{tab:res-coefficients}
  \centering
  \begin{tabular}{|r|c|c|}
    \hline
     & \multicolumn{1}{|c|}{$PDE_{\cal A}$} &  \multicolumn{1}{|c|}{$PDE_{\cal B}$}  \\
     & \multicolumn{1}{|c|}{{\scriptsize ${\hat u}_t=-0.99uu_{\mathbf{x}}$}} & \multicolumn{1}{|c|}{{\scriptsize ${\hat u}_t=-0.89uu_{\mathbf{x}}-0.9u_{\mathbf{xx}}$}}\\
    &  \multicolumn{1}{|c|}{{\scriptsize $-0.98u_{\mathbf{xx}}-0.985u_{\mathbf{xxxx}}$}} & \multicolumn{1}{|c|}{{\scriptsize $-0.85u_{\mathbf{xxxx}} + 0.0004u-0.0008u_\mathbf{x}u_\mathbf{xx}$}} \\
    \hline   $\epsilon^2_{coef}(\bm{\alpha},\bm{\hat{\alpha}})$ \ref{eq:nL2coef} & 0.0002 \greenpoint &  0.0149 \redpoint \\
    \hline   $\epsilon^1_{coef}(\bm{\alpha},\bm{\hat{\alpha}})$ \ref{eq:nL1coef} & 0.0150 \greenpoint &  0.1204 \redpoint \\
    \hline   $\epsilon_{coef}^\infty(\bm{\alpha},\bm{\hat{\alpha}})$ \ref{eq:maxerror} & 0.0200 \greenpoint &  0.1500 \redpoint \\
    \hline   $w\epsilon^2_{coef}(\bm{\alpha},\bm{\hat{\alpha}}) $ \ref{eq:wL2coef} & 0.0002 \greenpoint &  0.0014 \redpoint \\
    \hline    $NDCG(\bm{\alpha},\bm{\hat{\alpha}})$ \ref{eq:NDCG} &  {0.9862}  \greenpoint &  {0.8855} \redpoint \\
    \hline   $T_G(\bm{\alpha},\bm{\hat{\alpha}})$ \ref{eq:Tanimoto} & 0.9998 \greenpoint &  0.9834 \redpoint\\
    \hline
  \end{tabular}
\end{table}

The first one corresponds to a normalized version  of Eq.~\ref{eq:L1coef}, as used in \cite{thanasutives2024uncertaintypenalizedbayesianinformationcriterion}, as well as in  PDE-READ \citep{stephany2022pdereadhumanreadablepartialdifferential}, ABL-PDE \citep{ABLPDE2025Gao} and ANN-PYSR~\citep{zhang2025robustpdediscoverysparse}. This variant, which comes with the benefit  that errors in large coefficients do not dominate, is defined as follows:
\begin{eqnarray}
\mbox{if } TPR(\bm{\alpha},\bm{\hat{\alpha}})=1 \mbox{ then } N\epsilon^1_{coef(i)}(\bm{\alpha_i},\bm{\hat{\alpha}_i}) =
\begin{cases}
  \frac{|\bm{{\alpha}}_i-\bm{{\hat{\alpha}}}_i|}{|\bm{{\alpha}}_i|} \in \mathbb{R}_{\geq 0},  \forall i=1..|\Theta| \mbox{ s.t. }\bm{{\alpha}}_i\neq 0,  \\ 0 \mbox{ otherwise}.  \label{eq:nepsilon1}
\end{cases}
\end{eqnarray}
The important point in Eq.~\ref{eq:nepsilon1} is that it is conditioned on first matching the GT structural terms before comparing the coefficients. 

Second, the Tanimoto coefficient $T$, which originally deals with sets (as already defined in Eq.~\ref{eq:Jaccard}), and is often used in chemical compound discovery based on  molecular fingerprints \citep{wang2025chemrlearningreasonchemist,Lim2026},  has been extended to handle real-valued vectors. Using our notations, this generalized version $T_G$ is defined as follows:
\begin{eqnarray}
T_G(\bm{\alpha},\bm{\hat{\alpha}})=\frac{\bm{\alpha}^\top\bm{\hat{\alpha}}}{||\bm{\alpha}||^2_2+||\bm{\hat{\alpha}}||^2_2-\bm{\alpha}^\top\bm{\hat{\alpha}}} \in [0,1].
    \label{eq:Tanimoto}
\end{eqnarray}
Note that if $\bm{\alpha}$ and $\bm{\hat{\alpha}}$ are binary vectors (presence/absence of terms), the original expression $T$ (Eq.~\ref{eq:Jaccard}) and $T_G$ are equivalent. 
By comparing the coefficient vectors in this way, the extended Tanimoto metric accounts for both the PDE structure, through the presence or absence of candidate terms, and the magnitude of the associated coefficients.\\

\noindent {\bf \underline{Running Example:}} Reconsider the  KS equation $PDE_{\cal GT}$ and the learned equations  $PDE_{\cal A}$ and $PDE_{\cal B}$. The values computed from the different metrics discussed in this section are reported in Table~\ref{tab:res-coefficients}. They all indicate that $PDE_{\cal A}$ is much closer to the ground truth than $PDE_{\cal B}$. Except for the $\ell_2$-based error $\epsilon^2_{coef}(\bm{\alpha},\bm{\hat{\alpha}})$, which heavily over-penalizes local errors, all the $\epsilon^p_{coef}$ metrics behave similarly by yielding a ratio of approximately 1:7 to 1:8 in favor of $PDE_{\cal A}$. It is worth noticing that, in this example, the weighted error $w\epsilon^2_{coef}(\bm{\alpha},\bm{\hat{\alpha}})$ accounting for local sensitivity does not reveal a more pronounced difference between $PDE_{\cal A}$
  and $PDE_{\cal B}$. 
   Finally, the $NDCG$ metric appears to better capture the difference between the two learned PDEs than the Tanimoto coefficient, for which the two spurious terms have only a limited impact because of their small associated coefficients.  \\


\noindent    \fbox{\textbf{PROS} \cmark vs \textbf{CONS} \xmark}
\vspace{0.3cm}

\noindent \cmark These metrics directly measure the success of coefficient recovery. Aside from 
$w\epsilon^2_{coef}$  which requires evaluating partial derivatives w.r.t. to each $\bm{{\alpha}}_i$, they are simple to compute and easy to interpret. Tanimoto coefficient $T_G$ indirectly considers both the structure and the coefficient involved. $N\epsilon^1_{coef_i}$ is the only metric that guarantees the recovery of the structure while evaluating the coefficient errors. By leveraging a local sensitivity analysis, the weighted error conveys more information  because it takes into account the impact of each term on the induced dynamics.
\vspace{0.2cm}

\noindent\xmark They all require knowledge of $PDE_{\cal GT}$. Coefficient errors $\epsilon^p_{coef}$ are relevant when the structure of the identified  PDE is correct  otherwise there is no distinction between the coefficients of the correct terms and those of the missing or wrongly identified functions. Tanimoto coefficient $T_G$ is not able to strongly penalize PDEs containing spurious terms with small coefficients, yet likely to violate physical properties and induce large solution errors in the long term. $N\epsilon^1_{coef_i}$ can be used only if the identified structure is correct. 
The weighted error has an additional complexity burden due to the computation of derivatives w.r.t. $\bm{{\alpha}}_i$. 
Finally, a small $\epsilon^p_{coef}$ or a large $NDCG$ do not necessarily imply sparsity. 
\begin{center}
\rule{0.1\textwidth}{1pt} {\raisebox{-0.4ex}{\cmark \xmark}}
\rule{0.1\textwidth}{1pt}
\end{center}

\section{Metrics independent from the ground truth analytical form} \label{sec:noGT}

Discovering a new physical law from data, rather than recovering a ground truth equation $PDE_{\cal GT}$, corresponds to the very essence of scientific progress. This task becomes especially crucial when dealing with complex real-world phenomena which may rely on several interacting physical components making it difficult, even for experts, to derive an appropriate mathematical model. Through their expertise, scientists are  sometimes able to formulate hypotheses about the presence of certain terms in the equation, without however being able to provide its complete analytical form. This kind of scenario arises in many fields, such as complex materials, biological systems, physical flows, or socio-economic dynamics. This raises the key question of how to evaluate a discovered equation when there is no GT theory or only a partial knowledge of its expression.   This question  still holds when  $PDE_{\cal GT}$ is not intended to be used in a metric directly through its symbolic form. In both cases, the direct analytical comparison between the discovered and GT equations is not possible anymore. In this more realistic setting, 
it is generally assumed that, in addition to the training set ${\cal T}$, a \textbf{validation set} ${\cal V}$ is available, composed of $n_{\cal V}$ observation or simulation data. 
A standard way consists in numerically solving the  PDE identified from ${\cal T}$ ({\textbf{provided that initial and boundary conditions are known, as well as the time step $\delta_t$}}) and in collecting a set ${\cal E}$ of $n_{\cal V}$ \textbf{estimated solutions}  $\hat{u}(t,\mathbf{x})$. Then, it is possible to compare, most of the time from a pointwise perspective, 
the underlying dynamics of the latter with that of the validation data ${u}(t,\mathbf{x})$ comprising ${\cal V}$. 

We start this large overview with metrics based on {\it prediction errors} (Sec.~\ref{sec:errors}), followed by a few criteria focusing solely on \textit{sparsity} (Sec.~\ref{sec:sparsity}). The joint evaluation {\it prediction+sparsity} is addressed in several metrics discussed in Sec.~\ref{sec:tradeoff-sparsity-accuracy} before showing how to check the {\it physical consistency} of the learned equation in Sec~\ref{sec:consistency}, when physical properties are available. Finally, Sec.~\ref{sec:generalization} is devoted to the assessment of  the {\it out-of-distribution generalization} capability of the discovered PDE, {\it i.e.} when  ${\cal V}$ 
is generated under conditions (IC, BC, time horizon) that differ from those of ${\cal T}$.

\subsection{Prediction errors} \label{sec:errors}
Metrics presented in this section primarily address Question {\bf Q1} (which concerns the prediction accuracy), and for some of them Question {\bf Q2} (which deals with the physical consistency).\\

\noindent {\bf Solution error on $u$}: The  \textit{solution}-mean squared error $MSE({\hat u,u)}$ is arguably the most widely used metric for assessing the ability of the identified equation to accurately predict the underlying physical dynamics (see, {\it e.g.} \cite{Kaheman_2020,ducos:hal-05128224,yang2026discoveringsymbolicdifferentialequations}). $MSE({\hat u,u)}$ evaluates how well the estimated solutions in ${\cal E}$ fit \textit{in-distribution} validation data from ${\cal V}$\footnote{We assume here that the sets of validation data ${\cal V}$ and estimated solutions ${\cal E}$ are generated under the same physical conditions and parameter distributions as those used for generating the training data ${\cal T}$. Out-of-distribution dynamics will be discussed later when we will address the extrapolation ability of the discovered PDE.} when both ${\cal V}$ and ${\cal E}$ have been collected at the same $(t,\mathbf{x})$ coordinates allowing pointwise comparisons:
\begin{eqnarray}
MSE({\hat u,u})= \frac{1}{n_{\cal V}}\sum_{i=1}^{n_{\cal V}}\big( \hat{u}(t_i,\mathbf{x}_i)-u(t_i,\mathbf{x}_i) \big)^2 \in \mathbb{R}_{\geq 0}.
    \label{eq:MSE}
\end{eqnarray}
As suggested in PDE-BENCH \citep{PDEBench} where the authors compare several PiML models (PINN~\citep{raissi2019physics}, FNO~\citep{LiKALBSA21}, U-Net, etc.), it may be relevant to consider the scale-independent normalized version of $MSE({\hat u,u)}$ defined as follows (also used, {\it e.g.}, in ANN-PYSR~\citep{zhang2025robustpdediscoverysparse},  PITA \citep{zhu2025physicsinformed} and in \cite{Chen_2021}):
\begin{eqnarray}
nMSE({\hat u,u)}= \frac{\sum_{i=1}^{n_{\cal V}}\big(\hat{u}(t_i,\mathbf{x}_i)-u(t_i,\mathbf{x}_i)\big)^2}{\sum_{i=1}^{n_{\cal V}}u(t_i,\mathbf{x}_i)^2} \in \mathbb{R}_{\geq 0},
    \label{eq:nMSE}
\end{eqnarray}
or the normalized mean absolute error ($nMAE$):
\begin{eqnarray}
nMAE({\hat u,u)}= \frac{\sum_{i=1}^{n_{\cal V}}|\hat{u}(t_i,\mathbf{x}_i)-u(t_i,\mathbf{x}_i)|}{\sum_{i=1}^{n_{\cal V}}|u(t_i,\mathbf{x}_i)|} \in \mathbb{R}_{\geq 0}.
    \label{eq:nMAE}
\end{eqnarray}
Note that in a couple of PDE discovery methods, {\it e.g.}, PDE-NET~\citep{long2018pdenetlearningpdesdata}, LLM-SRBench~\citep{shojaee2025llmsrbenchnewbenchmarkscientific} or KeplerAgent~\citep{yang2026thinklikescientistphysicsguided},  a slight modification of Eq.~\ref{eq:nMSE} is used by subtracting from each term of the denominator  the spatial mean $\Bar{u}$ of $u$. 
\begin{eqnarray}
nMSE_{\Bar{u}}({\hat u,u)}= \frac{\sum_{i=1}^{n_{\cal V}}\big(\hat{u}(t_i,\mathbf{x}_i)-u(t_i,\mathbf{x}_i)\big)^2}{\sum_{i=1}^{n_{\cal V}}(u(t_i,\mathbf{x}_i)-\Bar{u}_i)^2} \in \mathbb{R}_{\geq 0}.
    \label{eq:nMSE-bar}
\end{eqnarray}

This allows to remove the constant spatial mode and ensure that the metric primarily reflects the reconstruction accuracy of the non-trivial spatial structures. Note that the $R^2$ score, defined as the complement of the $nMSE_{\Bar{u}}({\hat u,u)}$, is often used as an accuracy metric.\\

\noindent {\bf Residual error from $u_t$}: Instead of checking the prediction accuracy of  the vector field solution $\hat{u}$, 
the time derivatives ${\hat u}_t$ can be considered and compared with ${u}_t$. This amounts to evaluating the capacity of the PDE to accurately capture the local dynamics of the system, as done, {\it e.g.}, in PhysPDE~\citep{feng2025physpde}, SINDY-PI~\citep{Kaheman_2020}, MDBENCH~\citep{bideh2026mdbenchbenchmarkingdatadrivenmethods},  or MIO-SINDY~\citep{bertsimas2022learningsparsenonlineardynamics}. The \textit{residual}-mean squared error is defined as follows:
\begin{eqnarray}
nMSE_t({\hat u_t},u_t)= \frac{\sum_{i=1}^{n_{\cal V}}\big(\hat{u}_t(\mathbf{x}_i)-u_t(\mathbf{x}_i)\big)^2}{\sum_{i=1}^{n_{\cal V}}u_t(\mathbf{x}_i)^2} \in \mathbb{R}_{\geq 0}.
    \label{eq:n.utMSE}
\end{eqnarray}

\noindent {\bf Spectral error on ${\cal F}(u)$}: In order to evaluate the learned PDE from a signal processing perspective,  the prediction error in the Fourier space $f_{MSE}$  is computed in PDE-BENCH \citep{PDEBench}, constrained to low, middle, and high-frequency regions. 
This allows to quantify how well the equation reproduces the solution across different spatial scales, providing insight into whether both large-scale structures and fine-scale dynamics are accurately captured. 
$f_{MSE}$ is defined from the estimated solutions $\hat{u}(t,\mathbf{x})$ and validation data $u(t,\mathbf{x})$ as follows:
\begin{eqnarray}
f_{MSE}\big ({\cal F}(\hat{u}(t,.)),{\cal F}({u}(t,.)) \big )= \frac{\sum_{k_{min}}^{k_{max}}|{\cal F}(\hat{u}(t,.))-{\cal F}(u(t,.))|^2}{k_{max}-k_{min}+1} \in \mathbb{R}_{\geq 0},
\label{eq:f_{MSE}}
\end{eqnarray}
where ${\cal F}$ is a discrete Fourier transformation, and $k_{min}$, $k_{max}$ are the minimum and maximum indices in Fourier coordinates. From Eq.~\ref{eq:f_{MSE}}, one can compute the overall error along the trajectory as follows:\\
\begin{eqnarray}
\overline{f}_{MSE}\big({\cal F}(\hat{u}),{\cal F}({u})\big)
=
\frac{1}{N_T}
\sum_{i=1}^{N_T}
f_{MSE}\big({\cal F}(\hat{u}(i*\delta_t,.)),{\cal F}({u}(i*\delta_t,.))\big) \in \mathbb{R}_{\geq 0},
\label{eq:mf_{MSE}}
\end{eqnarray}
where, as a reminder, $N_T$ corresponds to the number of discrete temporal evaluations in $[0,T]$ and $\delta_t$ is the time step. Note that at time $t=0$, $u(0,\mathbf{x})=\hat{u}(0,\mathbf{x})$ by construction.\\
\begin{figure}[t]
    \centering
    \includegraphics[width=0.25\linewidth]{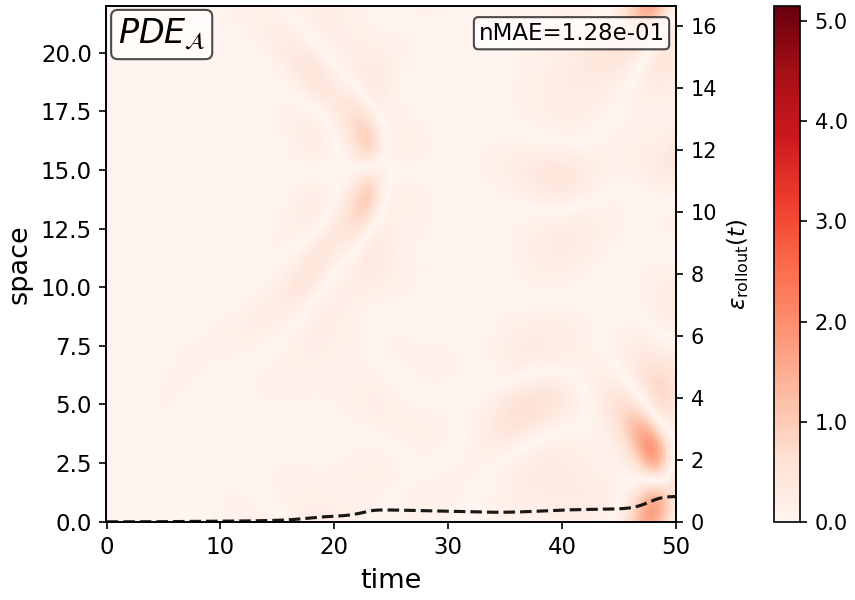}
    \includegraphics[width=0.25\linewidth]{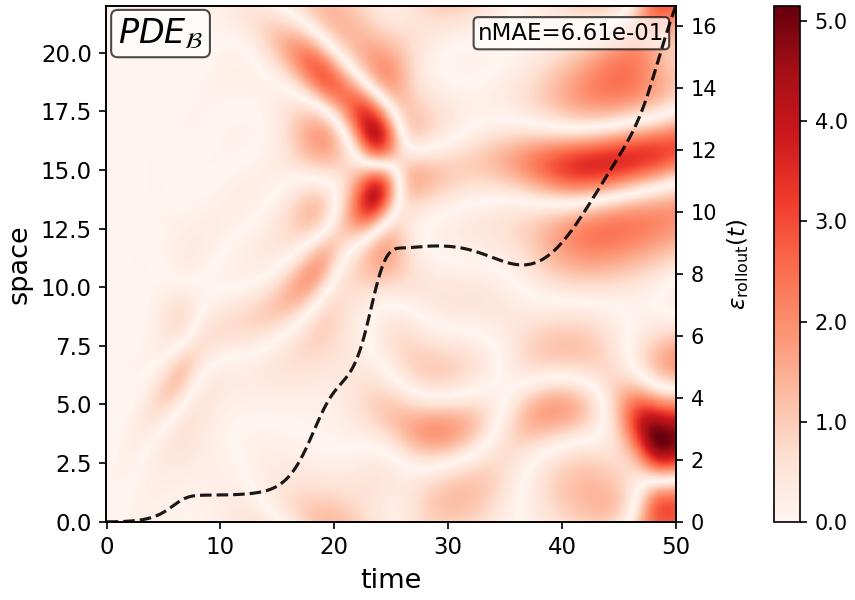}
    \includegraphics[width=0.22\linewidth]{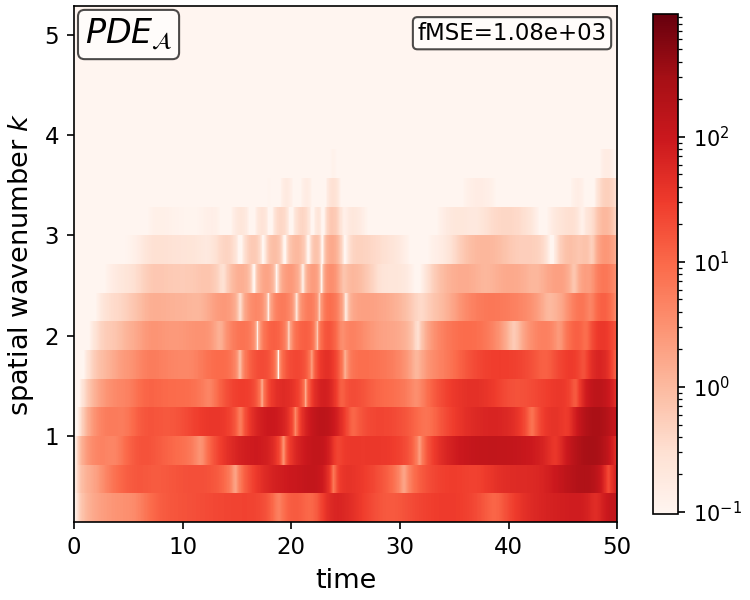}
    \includegraphics[width=0.22\linewidth]{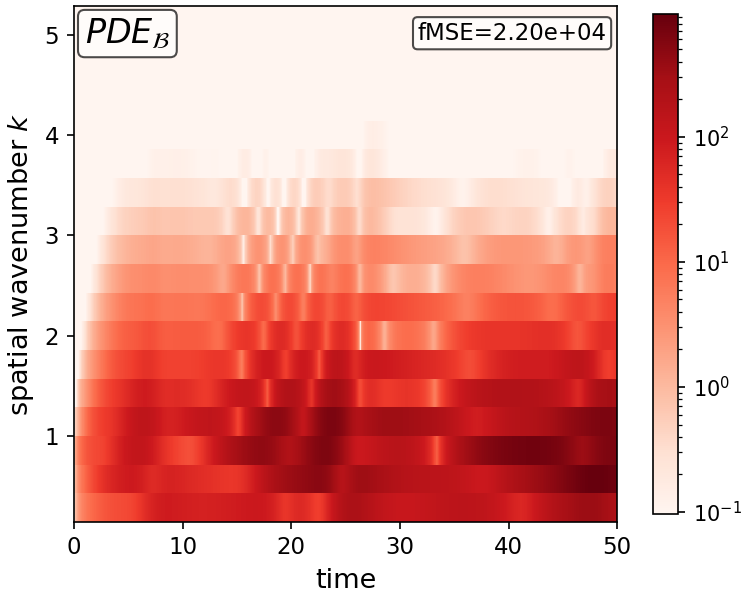}    
\caption{Pointwise prediction absolute error between ${\hat u}(t,\mathbf{x})$ and $u(t,\mathbf{x})$ for $PDE_{\cal A}$ (1st subfigure) and $PDE_{\cal B}$ (2nd), and the normalized mean absolute error $nMAE$, as defined in  Eq.~\ref{eq:nMAE}.  On both figures is also reported the rollout error $\epsilon^t_{rollout}({\hat u},u)$ (Eq.~\ref{eq:rollout}) over time (dashed-lines). The 3d and 4th panels represent the pointwise $MSE$ for $PDE_{\cal A}$ and $PDE_{\cal B}$ respectively, between the spatial Fourier transforms of the predicted and GT solutions as a function of time and spatial wavenumber (up to 5). The global $f_{MSE}$ is also reported on the two figures.} \label{fig:heatmap-KS}
\end{figure}

\noindent {\bf Pointwise error heatmaps:} 
Beyond the previous metrics which synthesize with one scalar the prediction error of the discovered equation, most PDE discovery papers also assess visually the quality of the equation by reporting in a heatmap,  at each spatio-temporal coordinate $(t,\mathbf{x})$, the corresponding pointwise  solution reconstruction absolute error  $|{\hat u}(t,\mathbf{x}) - u(t,\mathbf{x})| \in \mathbb{R}_{\geq 0}$, or $|{\hat u}_t-u_t| \in \mathbb{R}_{\geq 0}$ if  the residual error is considered. This visualization is  particularly well adapted for $(1d+t)$ PDEs (see Figure~\ref{fig:heatmap-KS}, first two panels, for the KS PDE). 

Another heatmap can provide valuable physical insights by reporting the pointwise MSE or MAE between the spatial Fourier transforms of the predicted and GT solutions as a function of time and spatial wavenumber. This kind of representation (as illustrated in Fig.~\ref{fig:heatmap-KS}, last two panels) reveals how the prediction error is distributed across spatial scales over time, indicating whether the discovered PDE accurately captures both large-scale (low-wavenumber) and fine-scale (high-wavenumber) dynamics.\\

\noindent {\bf Rollout error over time $t$}:
Even if, at first glance, two equations may appear similar (as observed with $PDE_{\cal A}$ and $PDE_{\cal B}$ in Fig.~\ref{fig:3PDE-KS}) yielding low $MSE$ or $MAE$ values over a limited temporal domain, it should be kept in mind that the models recursively propagate their own predictions. Initially, if the predicted trajectory may remain close to the GT, over longer rollout horizons, small initial inaccuracies accumulate and eventually may lead to significant deviations. This error accumulation cannot be  well captured by pointwise metrics. To assess the stability and dynamical fidelity of the learned PDE, the rollout error $\epsilon_{rollout}$ is often used. This trajectory discrepancy involves a spatial integral 
 at each time step  leading in the discrete case to the following expression:
\begin{eqnarray}
    \epsilon_{rollout}({\hat u},u)=\frac 1{N_T} \sum_{i=1}^{N_T} \epsilon^i_{rollout}({\hat u},u)= \frac 1{N_T} \sum_{i=1}^{N_T} ||{\hat u}(i*\delta_t,.)-u(i*\delta_t,.) ||_{L^2(\Omega)}^2 \in \mathbb{R}_{\geq 0}.
    \label{eq:rollout}
\end{eqnarray} 
As we will discuss later when addressing the question of generalization (see Section~\ref{sec:generalization}), $\epsilon_{rollout}$ is particularly well suited for evaluating the long-term extrapolation ability of the learned equation. Instead of summarizing this error with one scalar as defined in Eq.~\ref{eq:rollout}, it may be worthwhile to plot the curve representing the rollout error $\epsilon^t_{rollout}({\hat u},u)$ over time (see Figure~\ref{fig:heatmap-KS}). Note that this curve is not monotonically increasing, as it  represents  the average error over all time steps between 0 and $t$.

This  metric has been widely studied in  recent PiML articles addressing the accumulation of prediction errors when neural solvers are used as surrogate models for numerical schemes (see, {\it e.g.}, PDE-Refiner \citep{lippe2023pderefinerachievingaccuratelong}, DiffusionRollout \citep{yoo2026diffusionrolloutuncertaintyawarerolloutplanning} or \cite{lu2026stablelonghorizonpdeforecasting}, to cite some of them).\\

\noindent {\bf \underline{Running Example:}}  Back to the Kuramoto-Sivashinsky equation.  Table \ref{tab:res-prediction} summarizes the evaluation of $PDE_{\cal A}$ and $PDE_{\cal B}$ according to the prediction metrics discussed in this section.
For both the solutions $u$ and the temporal derivatives $u_t$, all evaluation metrics consistently demonstrate that $PDE_{\cal A}$ provides substantially more accurate estimated predictions $\hat{u}$ and $\hat{u}_t$ than $PDE_{\cal B}$.
This is visually confirmed by  the first two heatmaps presenting the pointwise  errors on Fig.~\ref{fig:heatmap-KS} which further allow to identify which parts of the dynamics lead to the the largest errors. For $PDE_{\cal B}$, interestingly, we can notice that the error pattern is strongly correlated with the shock regions of the ground truth (Fig.~\ref{fig:3PDE-KS}, left): while $PDE_{\cal B}$ captures the overall dynamics, it exhibits larger deviations near propagating discontinuities and steep fronts, where the solution is more challenging to approximate with this erroneous equation. 
The heatmaps on the Fourier transforms (two last panels of Fig.~\ref{fig:heatmap-KS}) show that for both $PDE_{\cal A}$ and $PDE_{\cal B}$, the prediction error is mainly concentrated in the lowest wavenumber modes (up to 4 for the latter), indicating that the largest discrepancies between the predicted and ground-truth spectral components occur at large spatial scales. However, the $\overline{f}_{MSE}({\cal F}(\hat{u}),{\cal F}({u}))$ error is approximately one order of magnitude smaller for $PDE_{\cal A}$ than for $PDE_{\cal B}$, confirming the lower quality of the latter. 
Finally, the curves of rollout error $\epsilon^t_{rollout}({\hat u},u)$, reported on Fig.~\ref{fig:heatmap-KS} (dashed-lines), show that the small initial error incurred by $PDE_{\cal B}$ diverges dramatically as time passes while the error associated with $PDE_{\cal A}$ remains small. This is also reflected in the overall rollout error computed at $t=50$, which is twenty times larger for $PDE_{\cal B}$ ($0.08170$ versus $16.6252$).

This running example clearly demonstrates that the graphical representations (heatmaps and rollout error curves) provide a valuable complement to quantitative metrics such as $MSE$, $MAE$, $\epsilon_{rollout}({\hat u},u)$ and $\overline{f}_{MSE}({\cal F}(\hat{u}),{\cal F}({u}))$. In particular, we must acknowledge that it is rather difficult to interpret the value of the latter on its own and to determine whether 1081.7 can be considered small without considering it in light of the Fourier heatmap.
\\

\begin{table}[t]
  \caption{Application of the prediction error metrics (Sec.~\ref{sec:errors}) to compare $PDE_{\cal A}$ versus $PDE_{\cal B}$ using estimated solutions $\hat{u}$ from both of them and validation data $u$ simulated from the KS ground truth equation $PDE_{\cal GT}$ as reference.}   
  \label{tab:res-prediction}
  \centering
  \begin{tabular}{|r|r|r|}
    \hline
     & \multicolumn{1}{|c|}{$PDE_{\cal A}$}  &  \multicolumn{1}{|c|}{$PDE_{\cal B}$} \\
    & \multicolumn{1}{|c|}{{\scriptsize ${\hat u}_t=-0.99uu_{\mathbf{x}}$}} & \multicolumn{1}{|c|}{{\scriptsize ${\hat u}_t=-0.89uu_{\mathbf{x}}-0.9u_{\mathbf{xx}}$}}\\
    &  \multicolumn{1}{|c|}{{\scriptsize $-0.98u_{\mathbf{xx}}-0.985u_{\mathbf{xxxx}}$}} & \multicolumn{1}{|c|}{{\scriptsize $-0.85u_{\mathbf{xxxx}} + 0.0004u-0.0008u_\mathbf{x}u_\mathbf{xx}$}} \\
    \hline   $MSE({\hat u,u)}$ \ref{eq:MSE} & 0.0371 \greenpoint &  0.7557 \redpoint \\
    \hline   $nMSE({\hat u,u)}$ \ref{eq:nMSE} & 0.0308 \greenpoint &  0.6277 \redpoint \\
    \hline   $nMAE({\hat u,u)}$ \ref{eq:nMAE} & 0.1281 \greenpoint &  0.6605 \redpoint \\
    \hline   $nMSE_t({\hat u_t},u_t)$ \ref{eq:n.utMSE} & 0.0003 \greenpoint &  0.0507 \redpoint \\
    \hline  $\overline{f}_{MSE}({\cal F}(\hat{u}),{\cal F}({u}))$ \ref{eq:mf_{MSE}} & 1081.7 \greenpoint &  22011.1 \redpoint \\
    \hline   $\epsilon_{rollout}({\hat u},u)$ \ref{eq:rollout} & 0.8170 \greenpoint &  16.6252 \redpoint \\
    \hline
  \end{tabular}
\end{table}

\begin{samepage}
\noindent    \fbox{\textbf{PROS} \cmark vs \textbf{CONS} \xmark}
\vspace{0.1cm}

\noindent \cmark $nMSE$ and $nMAE$-based metrics are scale invariant; $f_{MSE}$ provides insight into the capacity of the PDE to reproduce the solution across different spatial scales. Unlike the other metrics, $\epsilon_{rollout}({\hat u},u)$ allows to evaluate the stability of the PDE over time. The pointwise error heatmap enables the identification of  regions of the domain with the largest prediction errors (often in shock regions). The heatmap on the Fourier transform reveals how  errors are distributed across spatial scales over time.
\vspace{0.1cm}

\noindent \xmark 
Small prediction errors do not ensure sparse PDEs and physical consistency.
Therefore, the sole use of these metrics does not guarantee the correct PDE has been discovered.  These pointwise-metrics evaluate locally the PDE and therefore do not capture higher-level functional properties beyond local discrepancies. The heatmaps cannot be used directly for multi-dimensional PDEs.
\begin{center}
\rule{0.1\textwidth}{1pt} {\raisebox{-0.4ex}{\cmark \xmark}}
\rule{0.1\textwidth}{1pt}
\end{center}
\end{samepage}

\subsection{Sparsity} \label{sec:sparsity}

Based on the realistic assumption that most physical dynamics are governed by PDEs  involving only a few terms \citep{Feynman1965}, the metrics presented in this section 
focus exclusively on Question {\bf Q3}, which concerns the simplicity and, hopefully, interpretability of the discovered law. 
 The numerous metrics which aim at jointly considering the simplicity and the precision will be discussed in Section~\ref{sec:tradeoff-sparsity-accuracy}.

\begin{table}[t]
  \caption{Application of the sparsity metrics (Sec.~\ref{sec:sparsity}) to compare $PDE_{\cal A}$ versus $PDE_{\cal B}$.} 
  \label{tab:res-sparsity2}
  \centering
  \begin{tabular}{|r|r|r|}
    \hline 
     & \multicolumn{1}{|c|}{$PDE_{\cal A}$}  &  \multicolumn{1}{|c|}{$PDE_{\cal B}$} \\
    & \multicolumn{1}{|c|}{{\scriptsize ${\hat u}_t=-0.99uu_{\mathbf{x}}$}} & \multicolumn{1}{|c|}{{\scriptsize ${\hat u}_t=-0.89uu_{\mathbf{x}}-0.9u_{\mathbf{xx}}$}}\\
    &  \multicolumn{1}{|c|}{{\scriptsize $-0.98u_{\mathbf{xx}}-0.985u_{\mathbf{xxxx}}$}} & \multicolumn{1}{|c|}{{\scriptsize $-0.85u_{\mathbf{xxxx}} + 0.0004u-0.0008u_\mathbf{x}u_\mathbf{xx}$}} \\    \hline  $S_{terms}(\bm{\hat{\alpha}})$ \ref{eq:sparsity} & $3/5$ \greenpoint &  1 \redpoint\\
    \hline   ${\cal C}(\bm{\hat{\alpha}}^\top.\Theta)$ \ref{eq:ExpTree} & 11 \greenpoint &  18 \redpoint \\
    \hline
${\cal C}_2(\bm{\hat{\alpha}}^\top.\Theta)$ \ref{eq:ExpTreeV2} & 25 \greenpoint &  37 \redpoint \\
    \hline    
  \end{tabular}
\end{table}

\noindent {\bf Nonzero coefficients in $\bm{\hat{\alpha}}$:} A natural metric for evaluating the PDE sparsity simply consists in counting the number of nonzero coefficients in the learned equation $\hat{u}_t=\bm{\hat{\alpha}}^\top.\Theta$, while assigning an infinite score to an expression with no terms. This  leads to the following score $S_{terms}(\bm{\hat{\alpha}})$, used, {\it e.g.}, in \cite{ducos:hal-05128224}:
\begin{eqnarray}
S_{terms}(\bm{\hat{\alpha}}) = \left\{
    \begin{array}{ll}
    \infty & \mbox{if $\bm{{\hat{\alpha}}}$ is a zero vector,}\\
        1-\frac{|\{ \bm{{\hat{\alpha}}}_i | \bm{{\hat{\alpha}}}_i=0\}|}{|\Theta|} \in [0,1]  & \mbox{otherwise.}
    \end{array}
    \label{eq:sparsity}
\right.
\end{eqnarray}
\noindent The closer $S_{terms}$ is to 0, the sparser the PDE. It is worth noting that this score strongly depends on the size of the dictionary considered. Consequently, the same PDE may yield different scores depending on the number of functions available in $\Theta$. For this reason, $S_{terms}$ is  mainly useful when comparing competing PDEs learned from the same set of terms. \\

\noindent {\bf Complexity of the  expression tree of $\bm{\hat{\alpha}}^\top.\Theta$:}  
Since the notion of model complexity is often directly associated with the level of sparsity, the number of nodes, constant terms, and operations in the expression tree representing the equation is frequently  employed in symbolic regression during the exploration of the search space. This measure of complexity can be defined as follows: 
\begin{eqnarray}
    {\cal C}(\bm{\hat{\alpha}}^\top.\Theta)=size \left (ExpTree(\bm{\hat{\alpha}}^\top.\Theta )\right ) \in \mathbb{R}_{> 0}, 
    \label{eq:ExpTree}
\end{eqnarray}
where $size(.)$ outputs the number of elements involved in the expression tree $ExpTree$ of the learned equation. 
As done in AutoSINDy \citep{basiri2026discoverynonlineardynamicsautomated} and \cite{bideh2026mdbenchbenchmarkingdatadrivenmethods}, this metric can be easily computed by using the {\tt SymPy} Python library. 
In  NEXUS~\citep{Singh2025}, the complexity of the expression tree is defined slightly differently from the expression tree and for some constants $w_1, w_2, w_3 \in \mathbb{R}_+$:
\begin{eqnarray}
    {\cal C}_2(\bm{\hat{\alpha}}^\top.\Theta) & = & w_1.NbNodes \left (ExpTree(\bm{\hat{\alpha}}^\top.\Theta )\right ) + w_2.Depth\left (ExpTree(\bm{\hat{\alpha}}^\top.\Theta )\right ) \nonumber\\
    & + &w_3.OperatorWeight\left (ExpTree(\bm{\hat{\alpha}}^\top.\Theta )\right )\in \mathbb{R}_{> 0}. 
    \label{eq:ExpTreeV2}
\end{eqnarray}
 The number of
nodes in the expression tree is also used as a complexity measure for evaluating equations in README \citep{lau2025readme}, a framework for rapid equation
discovery. Although ${\cal C}(\bm{\hat{\alpha}}^\top.\Theta)$ allows for a valid assessment of the complexity of certain functions (as is the case, for example, for polynomials), note that it may be misleading if differential terms are treated as atomic symbols, as this fails to account for their differential order. Therefore, particular attention should be paid to properly representing the expressiveness associated with differential terms.

\noindent {\bf \underline{Running Example:}} 
Table \ref{tab:res-sparsity2} shows the values obtained from $PDE_{\cal A}$ and $PDE_{\cal B}$ for  $S_{terms}(\bm{\hat{\alpha}})$, ${\cal C}(\bm{\hat{\alpha}}^\top.\Theta)$ and ${\cal C}_2(\bm{\hat{\alpha}}^\top.\Theta)$ (where, for the latter,  $w_1=w_2=w_3=1$ and the \textit{OperatorWeight} is the sum of the arities of the expression tree). The three of them capture the fact that $PDE_{\cal A}$ is simpler than $PDE_{\cal B}$. On the other hand, since we have restricted the dictionary $\Theta$, for the sake of simplicity,  to a set containing only five functions, $PDE_{\cal B}$ is considered, according to  $S_{terms}(\bm{\hat{\alpha}})$, as not being sparse at all. Even $PDE_{\cal A}$, which consists of only three terms, is not considered to be that sparse. \\

\begin{samepage}
\noindent    \fbox{\textbf{PROS} \cmark vs \textbf{CONS} \xmark}
\vspace{0.3cm}

\noindent \cmark $S_{terms}(\bm{\hat{\alpha}})$, ${\cal C}(\bm{\hat{\alpha}}^\top.\Theta)$ and ${\cal C}_2(\bm{\hat{\alpha}}^\top.\Theta)$  directly measure the sparsity, and thus, evaluate the potential interpretability of the learned PDE. They are simple to compute and easy to interpret. 
${\cal C}(\bm{\hat{\alpha}}^\top.\Theta)$ and  ${\cal C}_2(\bm{\hat{\alpha}}^\top.\Theta)$ allow us, unlike $S_{terms}(\bm{\hat{\alpha}})$, to distinguish a simple term $u$ from a $k^{th}$-degree monomial. 
\vspace{0.3cm}

\noindent \xmark A high sparsity ({\it i.e.} small $S_{terms}(\bm{\hat{\alpha}})$, ${\cal C}(\bm{\hat{\alpha}}^\top.\Theta)$ or  ${\cal C}_2(\bm{\hat{\alpha}}^\top.\Theta)$) does not guarantee a valid PDE has been discovered. 
Two PDEs that are equivalent up to an analytical reformulation can lead to different sparsity scores or complexities, depending for the latter on how they are represented by the expression tree. $S_{terms}(\bm{\hat{\alpha}})$ depends on the size of the dictionary $\Theta$ and thus can be overly optimistic when a lot of functions are available, or overly pessimistic otherwise, as is the case in our KS PDE illustration. If partial derivatives are considered atomic terms, neither criterion accurately reflects the actual complexity of the expression.  ${\cal C}_2(\bm{\hat{\alpha}}^\top.\Theta)$ requires tuning three hyperparameters ($w_1, w_2, w_3$).
\begin{center}
\rule{0.1\textwidth}{1pt} {\raisebox{-0.4ex}{\cmark \xmark}}
\rule{0.1\textwidth}{1pt}
\end{center}
\end{samepage}

\vspace{0.2cm}
The limitations and risks associated with evaluating the quality of a PDE solely based on the number of  its constituent parts have led researchers to consider the ability of the equation to accurately predict the observed dynamics alongside the sparsity. This strategy has resulted in the proposal of numerous  metrics based on a trade-off between the two characteristics. They  are presented in the following section.

\subsection{Trade-off between prediction and sparsity} \label{sec:tradeoff-sparsity-accuracy}

The metrics examined in Sec.~\ref{sec:errors} and \ref{sec:sparsity} were intended to capture only one aspect of the PDE evaluation. In this section, we review measures that deal with the task from a joint perspective, addressing both the \textit{prediction accuracy} (Question {\bf Q1}) and \textit{sparsity} issues (Question {\bf Q3}).  
These different metrics differ from one another in the way they define and combine the two concepts.\\

\noindent {\bf Solution error–sparsity trade-off:} In \cite{cranmer2020discoveringsymbolicmodelsdeep}, considering the close relationship between the expressiveness of a PDE containing many terms and the risk of overfitting, 
the authors present a method for symbolic representation distillation. Recast in our context, given training data $u$ and a current reference equation $PDE_{\cal A}$. Between $k$ competing $PDE_{{\cal B}_i}$, $i=1\ldots k$, with more terms than $PDE_{\cal A}$, we select the one which maximizes the ratio between the (log) gain in prediction accuracy with respect to $u$ and the increase in complexity (due to the use of additional terms) compared to $PDE_{\cal A}$. 
The latter can be evaluated as the size of its expression tree as in Eq. \ref{eq:ExpTree}. 
To this end, the following $Score(\bm{\hat{\alpha}_{{\cal B}_i}}| \bm{\hat{\alpha}_{\cal A}},u)$ is computed (the larger the value, the better the evaluation in favor of ${{\cal B}_i}$):
\begin{eqnarray}
    Score(\bm{\hat{\alpha}_{{\cal B}_i}}|\bm{\hat{\alpha}_{\cal A}}, u) = \frac{ \log(nMAE({\hat u_{\cal A}, u)})-\log(nMAE({\hat u_{{\cal B}_i},u)})}{C(\bm{\hat{\alpha}_{{\cal B}_i}})-C(\bm{\hat{\alpha}_{\cal A}})} \in \mathbb{R}.
    \label{eq:Score}
\end{eqnarray}
In \cite{zhang2023discoveringreactiondiffusionmodelalzheimers}, the authors also leverage this metric in the context of PDE discovery for modeling Alzheimer’s disease. They use $Score(\bm{\hat{\alpha}_{{\cal B}_i}}| \bm{\hat{\alpha}_{\cal A}},u)$ for selecting the equation governing the best the underlying reaction-diffusion system.\\

\noindent {\bf Residual error–sparsity trade-off:} 
In \cite{xu2025generativediscoverypartialdifferential}, the following reward function is defined  to assess the identified PDE, as a multiplicative trade-off between a sparsity term and the coefficient of determination $R^2$ on the residuals of the equation:
\begin{eqnarray}
    Reward_1(\bm{\hat{\alpha}})=\underbrace{(1-c_0 \log_{10} |{\cal S}(\bm{\hat{\alpha}})|)}_{Sparsity} \times \underbrace{\left ( 1-\frac{\sum_{i=1}^{n_{\cal V}} \left ( u_t(\mathbf{x}_i)-\Theta_i^\top.\bm{\hat{\alpha}} \right )^2}{\sum_{i=1}^{n_{\cal V}}\left ( u_t(\mathbf{x}_i)- {\bar u}_t \right ) ^2} \right )}_{1-Residual \hspace{0.1cm} error \hspace{0.1cm}\approx\hspace{0.1cm} R^2} \in \mathbb{R}_{> 0},
    \label{eq:reward1}
\end{eqnarray}

where $|{\cal S}(\bm{\hat{\alpha}})|=|\{ \Theta_i : \bm{\hat{\alpha}}_i \neq 0 \}|$ corresponds to the number of nonzero coefficients involved in the PDE, and $c_0$ is a constant that governs the trade-off between simplicity and accuracy; the second term considers the residuals of the PDE that can be computed from the validation set ${\cal V}$. The higher $Reward_1(\bm{\hat{\alpha}})$ is, the better is the evaluation of the identified equation ${\hat u}_t=\bm{\hat{\alpha}}^\top.\Theta$. 

In the same vein, a reward function $Reward_2(\bm{\hat{\alpha}})$ is  presented in DISCOVER \cite{du2023discoverdeepidentificationsymbolically} to evaluate the PDE from a sparsity and residuals perspective. This metric is defined as follows:
\begin{eqnarray}
Reward_2(\bm{\hat{\alpha}})=\frac{1- \xi_1\times |{\cal S}(\bm{\hat{\alpha}})|-\xi_2 \times depth\left (ExpTree(\bm{\hat{\alpha}}^\top.\Theta )\right )}{1+Err(\bm{\hat{\alpha}}^\top.\Theta)} \in \mathbb{R},
    \label{eq:reward2}
\end{eqnarray}
where $Err(\bm{\hat{\alpha}}^\top.\Theta)$ is defined as the error on residuals $||u_t-\bm{\hat{\alpha}}^\top.\Theta||^2_2$, $depth\left (ExpTree(\bm{\hat{\alpha}}^\top.\Theta )\right )$ is the depth of the PDE expression tree, and $\xi_1$ and $\xi_2$ are two constants that control the trade-off. In NeuroSymBO \citep{qu2025dynamicbayesianoptimizationframework}, the authors leverage a criterion similar to Eq.~\ref{eq:reward2}, with $\xi_2=0$ and using $Err(\bm{\hat{\alpha}}^\top.\Theta)=nMSE({\hat u,u)}$ as defined in Eq.~\ref{eq:nMSE}.\\

\noindent {\bf Likelihood–sparsity trade-off:} Information-theoretic criteria such as Akaike Information Criterion ($AIC$) \citep{Akaike1973},  Bayesian Information Criterion ($BIC$)~\citep{Schwarz_1978}, and their variants, can be used to evaluate PDEs according to an accuracy/complexity trade-off. Their difference with the previously discussed metrics lies in the use of the likelihood of the dataset as a  probabilistic measure of fitness that can be further employed for uncertainty estimation. For all of them, a smaller value indicates a better evaluation.\\

Akaike Information Criterion ($AIC$) \citep{Akaike1973} is a statistical metric defined as follows:
\begin{eqnarray}
    AIC(\bm{\hat{\alpha}})=2|{\cal S}(\bm{\hat{\alpha}})|-2\log(L({\cal T},\bm{\hat{\alpha}})) \in \mathbb{R},
    \label{eq:AIC}
\end{eqnarray}
where, as a reminder,  $|{\cal S}(\bm{\hat{\alpha}})|$ corresponds to the number of nonzero coefficients involved in the expression, and $L({\cal T},\bm{\hat{\alpha}})$ is the 
likelihood of the data,  given the considered PDE. A common likelihood function in PDE discovery (assuming that the errors are normally distributed) is based on the residual sum of squares which amounts to calculating $||u_t-\Theta^\top . \bm{\hat{\alpha}}||_2^2$. In this case, $AIC(\bm{\hat{\alpha}})$, for the training set ${\cal T}$ of size $n_{\cal T}$, can be expressed as:
\begin{eqnarray}
    AIC(\bm{\hat{\alpha}})=n_{\cal T} \log \frac{||u_t-\Theta^\top . \bm{\hat{\alpha}}||_2^2}{n_{\cal T}} + 2|{\cal S}(\bm{\hat{\alpha}})| \in \mathbb{R}.
    \label{eq:cAIC}
\end{eqnarray}
In \cite{Mangan_2017}, the authors employ Eq.~\ref{eq:cAIC} as model selection criterion for SINDY, especially to rank the candidate models with respect to the regularization parameter associated with the sparsity-inducing norm. 
Note that Eq.~\ref{eq:cAIC} can be employed in the same way by using the field values $u$ instead of the time derivatives. In this case, it is necessary to consider the squared differences between  $u(t,\mathbf{x})$ and estimated solutions ${\hat u}(t,\mathbf{x})$ from the identified PDE.

Also an established metric, the $BIC$ (Bayesian Information Criterion)~\citep{Schwarz_1978}, known as the Schwarz information criterion, has experienced renewed interest in model selection, particularly through recent symbolic regression models for comparing learned analytical expressions. It differs from $AIC$ in the way it defines model complexity. While $AIC$ solely considers the number of functions in Eq.~\ref{eq:AIC}, the complexity is heavier in $BIC$ in the sense that it takes into account the size $n_{\cal T}$ of the dataset ${\cal T}$, so penalizing more and more complex models as the dataset grows. More formally, $BIC$ is defined in our PDE evaluation setting as follows:
\begin{eqnarray}
BIC(\bm{\hat{\alpha}})=log(n_{\cal T})|{\cal S}(\bm{\hat{\alpha}})|-2log(L({\cal T},\bm{\hat{\alpha}})) \in \mathbb{R},
\label{eq:BIC}
\end{eqnarray}
where $L({\cal T},\bm{\hat{\alpha}})$  can be defined in the same way as for $AIC$.

While it is argued in \cite{Mangan_2017} that  $BIC$ can  be efficiently used to select the best equation inferred by SINDY, it is demonstrated in \cite{thanasutives2024uncertaintypenalizedbayesianinformationcriterion}  that the minimization of $BIC$ could lead to
overfitted equations with unnecessary hypotheses, especially when the model is learned from  an overcomplete candidate library $\Theta$. To overcome this issue, the authors present an uncertainty-penalized Bayesian Information Criterion ($UBIC$) which penalizes the identified PDE both in terms of complexity and quantified uncertainty. The latter is estimated by using Bayesian regression to compute the coefficient of variation 
of the posterior PDE coefficients.

Finally, a Physics-informed Information Criterion ($PIC$) is used in \cite{Xu_2023} to evaluate the quality of a learned PDE. $PIC$ takes the following form:
\begin{eqnarray}
PIC(\bm{\hat{\alpha}})=L_r(\bm{\hat{\alpha}})\times L_p(\bm{\hat{\alpha}}) \in \mathbb{R},
    \label{eq:PIC}
\end{eqnarray}
where $L_r$ is a redundancy loss  measuring the parsimony of the PDE and $L_p$ is a physical loss evaluating its precision.  Using a moving horizon technique, $L_r$ represents the coefficient of variation $CV$ of the learned parameters over different moving horizons, under the realistic assumption that correct dominant terms are stable with small $CV$. $L_p$ measures the errors of a PINN optimized using the learned PDE as a ground truth. The authors show evidence that $PIC$ performs better than $BIC$ and $AIC$ in the presence of high noise. However, this improvement should be considered in light of the additional computational burden  associated with training (i) a neural surrogate to generate smoothed meta-data and calculate
derivatives and (ii) a PINN to evaluate the physical loss.\\

\noindent {\bf Kolmogorov complexity of $\Theta^{\top}\bm{\hat{\alpha}}$:}
Minimum Description  Length principle~\citep{KOLMOGOROV1998387} ($MDL$) is closely related to $AIC$ and $BIC$, but it is based on a distinct information-theoretic principle. $MDL$ posits that the best representation of a dataset ${\cal T}$ is the one that compresses it the most (leading to the so-called Kolmogorov complexity), using symbols from a finite alphabet. In this context, the presence of any regularity in ${\cal T}$, as is often observed in physical dynamical systems, can be leveraged to compress the data through  derivative terms. A model is preferred over another if it leads to the shortest description length, achieving the best trade-off between simplicity and predictive accuracy.

As far as we know, \textit{MDL}, used as a criterion for discovering mathematical expressions, has been applied in symbolic regression primarily on equations 
without spatial derivative terms.  The main reason is that its direct application to PDEs is non-trivial due to the additional complexity induced by differential operators, which cannot be properly assessed simply by counting the number of terms in the equation or the number of nodes in its expression tree.  Nevertheless, allowing accuracy and complexity to be directly traded-off, \textit{MDL} principle has been leveraged in~\cite{10136815}  for selecting ODE candidate representations lying on the Pareto front. 
Since the Kolmogorov complexity associated to \textit{MDL} has been shown to be incomputable~\citep{Vit_nyi_2020}, 
in {\sc MDLformer} \citep{yu2025symbolicregressionmdlformerguidedsearch}, the authors suggest to learn a neural network  which enables a robust estimation. In  {\sc AI-Feynman}~\citep{udrescu2020aifeynmanphysicsinspiredmethod}, \textit{MDL} is also employed for selecting the winning function among symbolic regression candidates. Recast in our setting, this criterion is defined as follows:
\begin{eqnarray}
MDL_{Fey}(\bm{\hat{\alpha}})=\log_2 N(\bm{\hat{\alpha}})+\lambda \log_2 \left  [max \left (1,\frac{\epsilon rr(\bm{\hat{\alpha}})}{\epsilon_d} \right ) \right ] \in \mathbb{R},
    \label{eq:MDL-Feynman}
\end{eqnarray}
where $\epsilon rr(\bm{\hat{\alpha}})$ is a fitting error of $\Theta^{\top}\bm{\hat{\alpha}}$ ({\it e.g.} $MSE$, $MAE$), $\epsilon_d$ is a normalization constant, and $N(\bm{\hat{\alpha}})$ is the rank of the equation size on the list of all candidates compared.

In {\sc SymLang} \citep{baig2026symmetryconstrainedlanguageguidedprogramsynthesis}, a \textit{MDL}-regularized Bayesian
model selection is presented based on the following  score:
\begin{eqnarray}
{MDL_{Sym}}(\bm{\hat{\alpha}})= -\log p\left ({\cal T}|e(\Theta^{\top}\bm{\hat{\alpha})} \right)+\lambda len \left (ExpT ree(\Theta^{\top}\bm{\hat{\alpha})}\right ) \in \mathbb{R},
    \label{eq:mdl-SymLang}
\end{eqnarray}
where $\lambda$ is a trade-off hyperparameter, and the description length $len \left (ExpT ree(\Theta^{\top}\bm{\hat{\alpha})}\right )$ counts the cost of encoding
the expression tree $ExpT ree(\Theta^{\top}\bm{\hat{\alpha}})$ of the equation and its constants. This function is defined as follows:
$len \left (ExpT ree(\Theta^{\top}\bm{\hat{\alpha})}\right )= \sum_{v \in nodes(e)}\log_2|{\cal O}_v|+|\Theta|.\log_2(c_{max}/\epsilon),$
where $|{\cal O}_v|$ is the local branching factor at node $v$, and $c_{max}$ and $\epsilon$ are constant range and precision. 
The authors point out that this criterion is
related to $BIC$~\ref{eq:BIC} and $AIC$~\ref{eq:AIC} but, unlike them, they stress the fact that the structural complexity is better measured in ${MDL_{Sym}}(\bm{\hat{\alpha}})$ by tree description length than parameter count alone.\\

\begin{table}[t]
  \caption{Application of the accuracy-sparsity metrics (Sec.~\ref{sec:tradeoff-sparsity-accuracy}) to compare $PDE_{\cal A}$ versus $PDE_{\cal B}$ using validation data ${\cal V}$ simulated  from the KS ground truth  $PDE_{\cal GT}$ as reference. Note that N/A means \enquote{\textit{Not Applicable}} because $Score$ is conditional on the increase in complexity (in our case from $PDE_{\cal A}$ to $PDE_{\cal B}$) and not the other way around.  } 
  \label{tab:res-trade-off}
  \centering
  \begin{tabular}{|r|r|r|}
   \hline 
     & \multicolumn{1}{|c|}{$PDE_{\cal A}$}  &  \multicolumn{1}{|c|}{$PDE_{\cal B}$} \\
    & \multicolumn{1}{|c|}{{\scriptsize ${\hat u}_t=-0.99uu_{\mathbf{x}}$}} & \multicolumn{1}{|c|}{{\scriptsize ${\hat u}_t=-0.89uu_{\mathbf{x}}-0.9u_{\mathbf{xx}}$}}\\
    &  \multicolumn{1}{|c|}{{\scriptsize $-0.98u_{\mathbf{xx}}-0.985u_{\mathbf{xxxx}}$}} & \multicolumn{1}{|c|}{{\scriptsize $-0.85u_{\mathbf{xxxx}} + 0.0004u-0.0008u_\mathbf{x}u_\mathbf{xx}$}} \\      
    \hline  $Score(\bm{\hat{\alpha}_{\cal B}}| \bm{\hat{\alpha}_{\cal A}},u)$ \ref{eq:Score} &  N/A  \hspace{0.6cm} &  -0.2343 \hspace{0.6cm} \\
    \hline   $Reward_1(\bm{\hat{\alpha}})$ \ref{eq:reward1} & 0.9043 \greenpoint &  0.8166 \redpoint \\
    \hline   $Reward_2(\bm{\hat{\alpha}})$ \ref{eq:reward2} & 0.0306 \greenpoint &  0.0002 \redpoint \\
    \hline   $AIC(\bm{\hat{\alpha}})$ \ref{eq:cAIC} & $-1.1$E+07 \greenpoint &  $-5.4$E+06 \redpoint \\
    \hline   $BIC(\bm{\hat{\alpha}})$ \ref{eq:BIC} & $-1.1$E+07 \greenpoint &  $-5.4$E+06 \redpoint \\
    \hline   $MDL_{Fey}(\bm{\hat{\alpha}})$ \ref{eq:MDL-Feynman} & 35234.7 \greenpoint &  42738.1 \redpoint \\
    \hline   ${MDL_{Sym}}(\bm{\hat{\alpha}})$ \ref{eq:mdl-SymLang} & $-5.3$E-06 \greenpoint &  $-2.7$E-06 \redpoint \\
    \hline
  \end{tabular}
\end{table}


\noindent {\bf \underline{Running Example:}}  
Table \ref{tab:res-trade-off} presents the results obtained from $PDE_{\cal A}$ and $PDE_{\cal B}$  on our running example with the metrics that consider the trade-off between accuracy and sparsity. Since several of them require specifying hyperparameter values, we used the ones reported in the original papers introducing these metrics. Following their experimental setups, we used: $c_0 = 0.2$ in $Reward_1(\bm{\hat{\alpha}})$ \citep{xu2025generativediscoverypartialdifferential}; 
$\xi_1 = 0.01$ and $\xi_2 = 0.0001$ in $Reward_2(\bm{\hat{\alpha}})$ \citep{du2023discoverdeepidentificationsymbolically}; 
$\epsilon rr$ is defined as the $MSE$, $N = 1$, $\epsilon_d = 10^{-15}$,  $\lambda = n_{\mathcal{V}}^{1/2}$ in $MDL_{Fey}(\bm{\hat{\alpha}})$~\citep{udrescu2020aifeynmanphysicsinspiredmethod}; 
$c_{max} = 10$, $\epsilon = 10^{-6}$, $\lambda = 0.1$ in ${MDL_{Sym}}(\bm{\hat{\alpha}})$ \citep{baig2026symmetryconstrainedlanguageguidedprogramsynthesis}.
As visually highlighted by the green dots in the table, all metrics (but $Score(\bm{\hat{\alpha}_{\cal B}}| \bm{\hat{\alpha}_{\cal A}},u)$) are able to identify $PDE_{\cal A}$ as the best equation. Nevertheless, some differences can be observed in the gap between the two PDEs. 
In particular, $Reward_2(\bm{\hat{\alpha}})$ and ${MDL_{Sym}}(\bm{\hat{\alpha}})$ seem to better capture the expected large difference in quality between $PDE_{\cal A}$ and $PDE_{\cal B}$, thanks to the presence of the size of the expression tree in the complexity formulae. The other metrics tend to smooth out this difference due to the use of the number of functions involved in the equations. Regarding $AIC$ and $BIC$, the experiments on the Kuramoto-Sivashinsky data reveal a particular behavior of these two information criteria. By construction, the likelihood term in both metrics rapidly outweighs the complexity penalty as the number of data points increases. Because this experiment comprises 1,024,000 data points, the $AIC$ and $BIC$ values for $PDE_{\cal A}$ and $PDE_{\cal B}$ mainly reflect the predictive performance of the equations rather than their sparsity. The latter is hidden behind the massive numbers obtained from the likelihood.\\

\noindent    \fbox{\textbf{PROS} \cmark vs \textbf{CONS} \xmark}
\vspace{0.3cm}

\noindent \cmark All these metrics have the advantage of combining two desirable properties of a good PDE: accuracy and simplicity. They reduce equation evaluation to a single number, making it easy to rank competing equations. The size of the expression tree seems to be more relevant for measuring the complexity than simply counting the number of terms in  equations that may comprise functions with different expressiveness. $PIC$ further takes into account information about the residuals of the learned equation.
\vspace{0.3cm}

\noindent \xmark Tuning hyperparameters is required for the evaluation of  $Reward_1, Reward_2, MDL_{Fey}$ and $MDL_{Sym}$. 
$Reward_1$, \textit{AIC}, \textit{BIC}, and \textit{PIC} use the number of terms in the equation as a measure of simplicity. This is valid for PDEs composed of functions with similar structural complexity. By construction, the likelihood term in both $AIC$ and $BIC$ 
quickly overwhelms the effect of the complexity penalty as the sample size increases. Finally, $PIC$ comes with an additional computational burden (learning of two neural networks).  
\begin{center}
\rule{0.1\textwidth}{1pt} {\raisebox{-0.4ex}{\cmark \xmark}}
\rule{0.1\textwidth}{1pt}
\end{center}

\subsection{Physical consistency} \label{sec:consistency}
Although difficult to implement due to its reliance on prior knowledge that is not always available, verifying that the identified equation does not violate important physical properties is crucial for scientific advancement. 
In particular, checking physical consistency makes it possible to avoid overstating the validity of a supposedly new law. Indeed, a low reconstruction error associated with a sparse expression often prompts us to report a new discovery, even though it does not guarantee that the PDE represents a plausible physical theory. 
In this section, we discuss various ways of verifying that certain physical constraints are satisfied, 
thereby addressing Question {\bf Q2}.\\

\begin{figure}[t]
    \centering
    \includegraphics[width=0.8\linewidth]{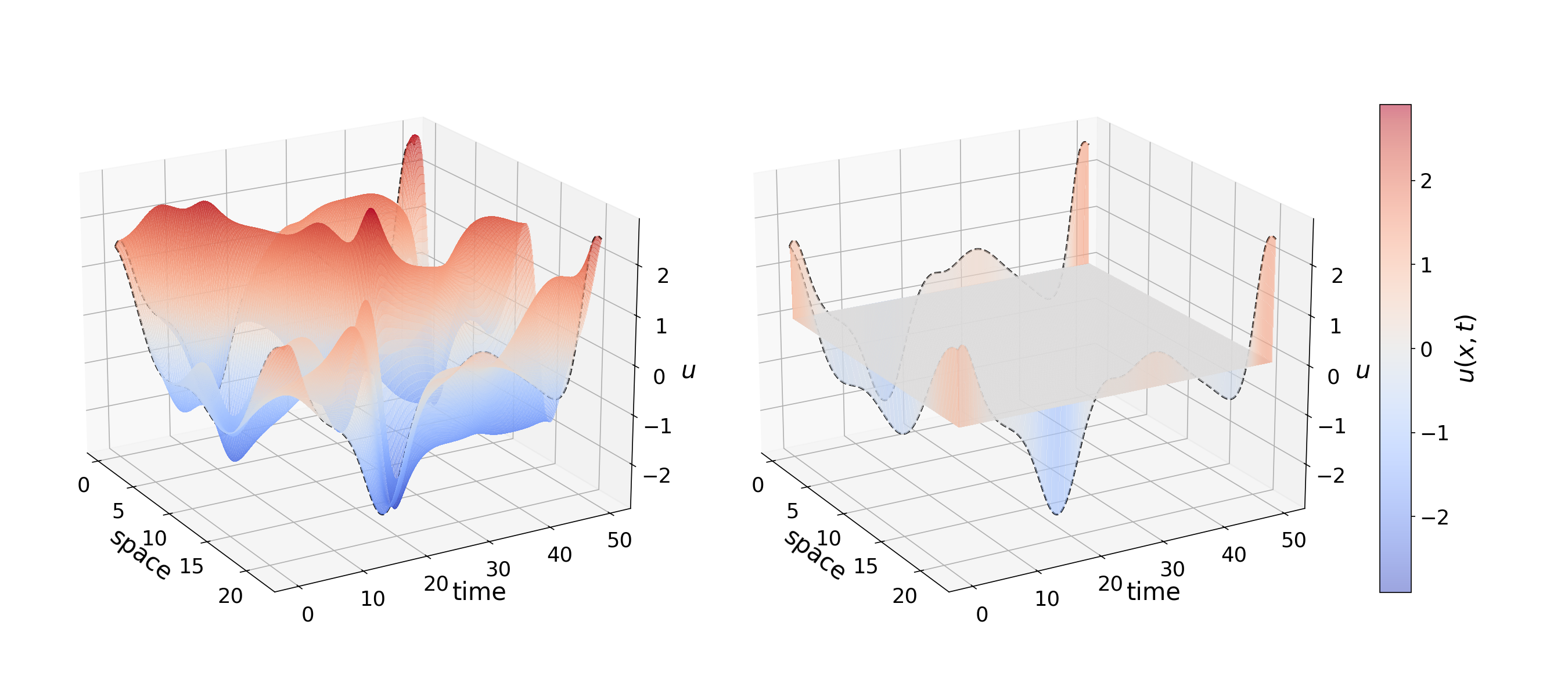}
    \caption{Possible pitfall of over-interpreting small residuals: 3D solutions of the Kuramoto-Sivashinsky equation $PDE_{\cal GT}$ (left) and  solutions simulated from a learned equation that would satisfy the initial/boundary conditions and return 0 everywhere else (right), leading thus to zero residuals.}
    \label{fig:residuals_KS}
\end{figure}

\noindent {\bf Consistency checking through small residuals:} A widely used strategy in PiML to assess the physical consistency of a model consists in verifying the PDE residuals are low. For instance, in PINNs \citep{raissi2019physics}, this means that the underlying dynamics optimized during training is {\bf consistent with the physics modeled by the known equation $PDE_{\cal GT}$}. The situation is substantially different in post-hoc evaluation. In particular, observing small residuals offers \textbf{weaker physical guarantees when the analytical expression $PDE_{\cal GT}$ is either unavailable or intentionally not used}.  In this case, a small residual error (as defined in Eq.~\ref{eq:n.utMSE})  on validation data ${\cal V}$ only indicates the absence of overfitting through a compatibility with the learned differential operator $\hat{\cal N}[u]$. But it does not ensure that the discovered PDE is the true governing law of the system. 
To illustrate the risk of over-interpreting low residuals, 
Fig.~\ref{fig:residuals_KS}
reports in 3D the solutions of both the Kuramoto-Sivashinski equation ($PDE_{\cal GT}$ on the left) and of a PDE that would satisfy the initial/boundary conditions and
return 0 everywhere else (right)\footnote{Although such a hypothetical equation does not satisfy the $n$-times differentiability requirement of a valid $n$-order PDE,  it can still be learned by a PiML algorithm, if no precautions are taken.}. We can see that this PDE leads to zero residuals while completely missing the true physical dynamics. Note that this problem has been addressed from both optimization and statistical generalization perspectives in \cite{10.3150/24-BEJ1799} when dealing with PINNs. 

\noindent {\bf Consistency checking through explicit physical properties:} Beyond analyzing the residuals, one can also leverage different types of physical properties for evaluating the validity of a PDE. As discussed later in Sec.~\ref{sec:guidelines}, depending on the domain of study and the level of expertise available, these properties can be either {\bf (i) a priori known, (ii) assumed to hold or (iii) learned from data}. They can be expressed through various types of constraints to be satisfied, such as {\it symmetries, conservation laws, dimensional consistency}, etc. 

Among them, \textit{symmetries} play a key role as they correspond to a fundamental principle in physical systems, characterizing their invariance under  translations, rotations, and scaling. 
For this reason, 
symmetries, which are present in many datasets, have become an important algorithmic and theoretical object of study in the machine learning community (see, {\it e.g.} \cite{perin2025abilitydeepnetworkslearn,otto2025unifiedframeworkenforcediscover}). Their incorporation into PiML processes has  concerned the optimization of neural PDE surrogate solvers \citep{akhoundsadegh2023liepointsymmetryphysics,huang2025geometricphysicalconstraintssynergistically,brandstetter2022liepointsymmetrydata}, as well as the discovery of equations from data \citep{yang2026discoveringsymbolicdifferentialequations,udrescu2020aifeynmanphysicsinspiredmethod,baig2026symmetryconstrainedlanguageguidedprogramsynthesis,pmlr-v235-dalton24a}. Regarding this last research direction, the use of symmetries 
in symbolic regression allowed to drastically reduce the number of candidate expression trees by pruning the hypothesis space before search. This can therefore prevent the discovery of inconsistent PDEs that may  appear relevant at first glance if one considers only prediction errors and  sparsity criteria. 

Respecting \textit{conservation laws }is also crucial in PDE discovery, as it improves the physical plausibility of the identified equation. In {\sc ProbConserv}~\citep{Hansen2024}, a general framework is presented for satisfying  conservation laws in equation discovery from the finite volume perspective. The authors address this task by combining the integral form (rather than its differential form) of the conservation laws with a Bayesian update. In \cite{baez2024guaranteeing}, the authors present {\sc PINN-Proj}, a PINN-based model which leverages a new projection method to satisfy conservation laws. Experimental results are reported showing that   {\sc PINN-Proj} outperforms a classic PINN thanks to this  physical consistency requirement. 

Additionally, checking the \textit{dimensional consistency} can be also of great help to filter out invalid models. This strategy has been efficiently used in the DHC-GEP method \citep{ma2024dimensionalhomogeneityconstrainedgene} for discovering governing equations in gene expression programming. In \cite{BRENCE2023742}, the authors use attribute grammars to ensure the dimensional consistency of the discovered equations and demonstrate their effectiveness on  Feynman benchmarks.
Note that, although dimensional consistency is generally considered as a hard (binary) constraint, some equation discovery algorithms employ it through a regularization term,   penalizing candidates that do not satisfy it. This is the case for NEXUS~\citep{Singh2025}, built on graph-based equation representation, evolutionary optimization, and adaptive reinforcement learning and where the dimensional consistency  
for a candidate equation $\hat{u}_t=\Theta^{\top}\bm{\hat{\alpha}}$ 
is defined, using our notations, as follows: 
\begin{eqnarray}
D\!\left(\Theta^{\top}\bm{\hat{\alpha}}\right)
=
\sum_{t\in T\!\left(\Theta^{\top}\bm{\hat{\alpha}}\right)}
\log\!\left(
\frac{[t]}{[u_t]}
\right),
\label{eq:dim-consist}
\end{eqnarray}
where $[t]$ denotes the dimension of sub-expression $t$, $[u_t]$ is the target dimension and $T\!\left(\Theta^{\top}\bm{\hat{\alpha}}\right)$ is the set of all sub-expressions in $\Theta^{\top}\bm{\hat{\alpha}}$.

Beyond symmetries, conservation laws, and dimensional consistency, note that other physical properties can be also subject to verification, such as \textit{positivity}, \textit{stability, causality, time reversibility, etc.}. Some of them will be discussed further down in this section.

It is important to emphasize that the  references cited   above  primarily fall within the PDE discovery setting rather than the post-hoc evaluation framework. They therefore focus on enforcing physical constraints embedded in the optimization process or used during model training, as in symbolic regression, to select the most consistent equation along the Pareto front. Nevertheless, we argue that these very same physical properties can also be exploited to evaluate a posteriori learned PDEs, although they are surprisingly often overlooked in the post-hoc evaluation literature. In this setting, they will complement certain physical properties that may be assumed to hold for the physical system under study ({\it e.g.} convection, diffusion, advection, etc.) or other consistency-based metrics that are derived from numerical analysis, such as \textit{forward–backward verification, mesh refinement study, temporal convergence}, as presented in the following. Below we attempt to categorize the main physical properties that can be checked  post-hoc according to the difficulty of their assessment. 
\begin{enumerate}
    \item {\bf Inspection of the structural form of the PDE}: Some structural properties can be evaluated by  examining the symbolic expression of the equation. Here are a few simple examples intended to facilitate non-specialist reader's understanding:  (i) if the underlying physical process is known to be \textit{dissipative} (as in heat conduction, viscous fluid flows, damped mechanical systems, etc.), one would therefore expect the term  $-\nabla \cdot \left( c \nabla \mathbf{u} \right)$ to appear in the PDE; (ii) if we are studying systems related to atmospheric circulation, ocean currents, or heat transport in fluids, the \textit{convection} term  $\nabla \cdot (c\mathbf{u})$ should be present; (iii) as a third illustration, \textit{self-advection}, which occurs when a field transports itself, as in fluid dynamics, implies that 
    the term $(\mathbf{u}\cdot\nabla)\mathbf{u}$ should appear in the symbolic expression. Even though  identifying diffusion, convection, or advection mechanisms solely from the structure of a PDE can provide useful physical insights, note that it does not guarantee that the corresponding physical properties are actually satisfied by the model. The verification generally requires more advanced analytical manipulations, such as deriving conservation laws or energy estimates (see next point \textbf{2.}). 
    Other properties can be assessed by inspecting the form of the PDE. This is the case of  {\it spatial and temporal symmetries} that can be verified if there is no term of the form $\mathbf{x}\mathbf{F}(\mathbf{u})$ or $t\mathbf{F}(\mathbf{u})$. On the other hand, while often difficult to enforce during the identification process, \textit{dimensional consistency} is much easier to check a posteriori by simply inspecting each term of the equation. 
    
    \item {\bf Analytical derivation from the PDE:}  Many properties, including {\it conservation laws, positivity preservation, energy dissipation}, and certain {\it invariances} require symbolic manipulation of the equation. For instance, checking for \textit{conservation laws} consists in determining whether the PDE can be rewritten in conservative form, $u_t +\nabla\cdot\mathbf{F}=S$, by identifying a flux function $\mathbf{F}$, and a source term $S$.  As an example, for reaction–diffusion PDEs, the mass conservation law is not generally expected to hold, while this is required for Navier–Stokes PDEs which describe the motion of viscous fluids. On the other hand, \textit{positivity preservation} is generally harder to assess than conservation laws, as it is not solely determined by the structure of the PDE but depends on the global properties of the associated evolution operator. One possibility is to resort to the \textit{maximum principle} \citep{HoldenH:02a} according to which if the solution of the PDE were to become negative, it would attain a first negative minimum, contradicting the maximum principle. 
     Note that this analytical verification is  limited to elliptic and parabolic PDEs.   Verifying \textit{energy dissipation} is also a non-trivial task in general: it requires deriving an appropriate energy law and proving analytically that the associated energy decreases over time, which may be difficult for complex nonlinear models.
    \item  {\bf Simulation-based verification:} Because a PDE is meant to model dynamics leading to a solution $u(t,\mathbf{x})$, {\it well-posedness} and {\it uniqueness} are also desirable properties when assessing the quality of a learned PDE.    
    However, these structural properties, which generally require advanced mathematical analysis, are often difficult to establish rigorously. A practical solution is to leverage simulations (provided the initial/boundary conditions are known) and manipulating data instead of analytical equations to evaluate a property.   For instance, by performing a \textit{convergence study}, {\it i.e.} by refining the spatial mesh or time step, and checking if the solution $u(\mathbf{x},t)$ converges, we can evaluate the risk of instability or  ill-posedness of the learned PDE (see Sec.~\ref{sec:generalization} for more details). When priors are available about the \textit{reversibility}~\citep{hairer2006geometric} of the physical phenomenon, an inverse reconstruction procedure  can be performed consisting of simulating the learned PDE forward in time and then integrating backward from the predicted final state $u(T,\mathbf{x})$ to assess whether the initial conditions are recovered or not. 
    For example, if the model is assumed to describe a dissipative system ({\it e.g.} reaction-diffusion, Navier–Stokes PDEs), the reconstruction of the initial conditions is not expected to hold. On the other hand, to verify some physical \textit{invariance}, it suffices to apply the considered transformation to the PDE, then solve the equation  (provided that it is numerically possible) and compare the resulting solutions with the original ones. The validation of the invariance  can then be done (up to numerical errors and the choice of a decision threshold) using standard metrics such as those discussed so far ({\it e.g.}, $MSE$, $MAE$, etc.).
Note that other properties such as the preservation of {\it pattern formation}, can  be assessed visually from simulations (this is the case of the Kuramoto-Sivashinsky equation), particularly in the case of $(1d + t)$ PDEs.  
\end{enumerate}


Let us conclude this section with a real-world illustration of how physical properties can be used for evaluating a PDE. The considered applied setting is surface engineering, and more specifically self-organization of matter induced by laser–matter interaction, for which the Kuramoto–Sivashinsky PDE is a candidate to (partially) model this complex phenomenon. Therefore, through this example, we remain within the scope of our running example.\\

\noindent {\bf \underline{Running Example:}}  
In laser–matter interaction, a femtosecond laser is typically employed to texture the surface of a matter and endow it with specific physical features ({\it e.g.}, wettability, hydrophobicity, antibacterial properties, etc.). The observed dynamics (see, \textit{e.g.}, SEM images on Fig.~\ref{fig:allsem} used in \cite{alibanna:hal-05012225}), together with the prior knowledge that this type of physical system leads to pattern formation, 
makes it possible to establish a number of properties that a learned PDE should satisfy. \textbf{Without assuming that the KS equation is a possible candidate} for modeling this phenomenon, one may reasonably expect the discovered PDE to be spatially symmetric under translations, to possess both linear (for diffusion or anti-diffusion) and nonlinear terms (saturation of instabilities, generation of chaos), and to be dissipative. These properties can thus be checked post-hoc to assess the physical consistency of the learned PDE. 

 As a toy example, let us consider the following discovered equation: ${\hat u}_t=-1.0uu_{\mathbf{x}}-1.0u_{\mathbf{xx}}-1.0u_{\mathbf{xxxx}}+1.0\mathbf{x}u$ which is very close to $PDE_{\cal GT}$. 
 We can observe that: (i) $uu_{\mathbf{x}}$ seems relevant because it represents the self-advection of the field and is responsible for nonlinear distortion; (ii) similarly, $u_{\mathbf{xx}}$ plays the role of a destabilizing term that allows the growth of structures; (iii) ${u_\mathbf{xxxx}}$ makes sense because it confirms the dissipative nature of the system and prevents blow-up by stabilizing small scales which is key in pattern formation; (iv) Finally, $\mathbf{x}u$ assumes that pattern formation is spatially dependent which is not consistent with the observed dynamics. In conclusion, without additional calculations or simulations, provided some background expertise, we can state by simply inspecting its structural form that this learned equation is not a physically plausible law  because of the spurious last term. \\ \\

\begin{samepage}
\begin{figure}[t]
\centering
\includegraphics[width=1\textwidth]{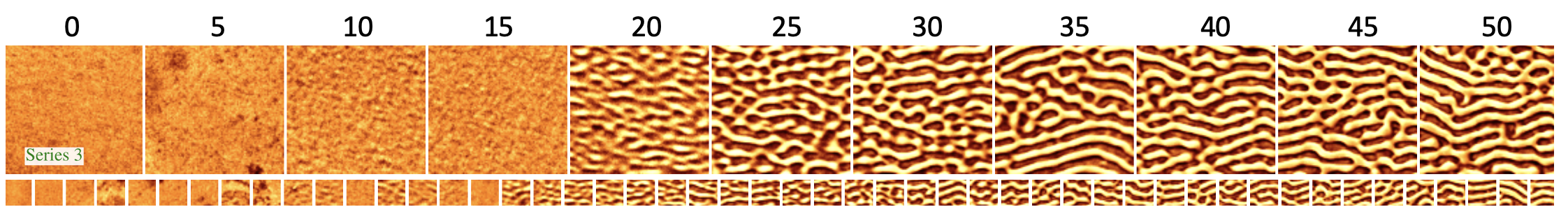}
\caption{Subset of Scanning Electron Microscopy (SEM) images of surfaces irradiated by a femto-second laser with cross-polarized double pulses, varying the number of double pulses between 0 and 50, when the laser fluence $= 0.25$ J/$cm^2$ and the delay is 36ps (between the two laser pulses) \citep{alibanna:hal-05012225}. Even without expertise, the visual inspection reveals that spatial invariance and the growth of structures should be reflected in the discovered PDE.}\label{fig:allsem}
\end{figure}

\noindent    \fbox{\textbf{PROS} \cmark vs \textbf{CONS} \xmark}
\vspace{0.3cm}

\noindent \cmark Well-founded physics-informed approach to PDE evaluation. Help detect physical inconsistency  even when prediction errors are small and the PDE is sparse. As shown in our protocol {\sc Pde-Check}~(see Algo. \ref{alg:pde-check}), checking the satisfiability of physical properties makes it possible to stop the evaluation process immediately if any of them is violated, thereby avoiding the computation of the other metrics. 
\vspace{0.3cm}

\noindent \xmark  Require prior knowledge of the physical properties to be verified. If not available, they can be learned from data  without guaranteeing that they are valid though. Some properties may be difficult or impossible to formally verify. 
Although being relevant during a PDE discovery process (used as a regularization), $D\!\left(\Theta^{\top}\bm{\hat{\alpha}}\right)$ is of limited interest during post-hoc evaluation, where strong constraints about the dimensionality conservation must be enforced.

\begin{center}
\rule{0.1\textwidth}{1pt} {\raisebox{-0.4ex}{\cmark \xmark}}
\rule{0.1\textwidth}{1pt}
\end{center}
\end{samepage}

\subsection{Out-of-distribution generalization} \label{sec:generalization}

As in classic statistical learning, the ability of a PDE, 
optimized from a set of data ${\cal T}$, to accurately predict unseen states 
within the training conditions does not imply that this equation is the governing one. 
Indeed, even an incorrect PDE may accurately predict the  system's evolution over short time horizons, 
while failing to capture the true governing dynamics.
 As a result, reconstruction errors remain small, and the evaluation metrics discussed so far and computed on an {\it in-distribution} validation set ${\cal V}$ may lead to an overly optimistic assessment of the model's quality. 
This is why Question {\bf Q4}, which concerns the capacity of  a discovered PDE to generalize {\it out-of-distribution} (OOD), is key in PDE evaluation.

OOD generalization can be assessed through several experimental settings that probe the robustness and stability of an equation beyond the conditions used for its identification. Common scenarios encountered in PDE discovery include the generalization to {\it unseen initial conditions (IC)}, {\it unseen boundary conditions (BC)} or \textit{longer prediction horizons}. Other settings, such that \textit{parameter values outside the training range} or {\it varying dynamical regimes} can be considered but  they are less applicable when dealing with post-hoc evaluation.  Indeed, regarding the former, it is less meaningful to modify a posteriori a learned coefficient associated with a derivative term. \\

\noindent \textbf{Varying initial and boundary conditions:} Performing experiments under changing IC and BC
provides insight into whether the discovered PDE captures the underlying physical law rather than merely fitting the observed data distribution.
 In real world situations, where the governing $PDE_{\cal GT}$ is unknown or only partially characterized, assessing the OOD generalization capacity is generally more difficult because it requires either additional observation data obtained from varying conditions (which is not always possible) or a learned (neural) operator allowing the generation of OOD trajectories from different IC and BC (see Sec.~\ref{sec:guidelines} for more details). When $PDE_{\cal GT}$ is known, evaluating OOD generalization capabilities is relatively straightforward, since the  governing law can be used at will to generate simulation data under changing IC and BC. The following evaluation protocol can then be applied.

 \noindent 
 Let us consider an equation $PDE_{\cal T}$ learned from the training set ${\cal T}$  generated with the following IC and BC respectively:
    $$u(0,\mathbf{x}) \sim {\cal D}^{ic}_{\cal T}, \forall \mathbf{x} \in \Omega$$
    $$u(t,0), u(t,L) \sim {\cal D}^{bc}_{\cal T}, \forall t \in [0,T]$$ where the spatial domain is supposed to be $\Omega=[0,L].$
\begin{enumerate}    
    \item Generation of two sets of $n_{\cal V}$ OOD \textbf{validation data} from $PDE_{\cal GT}$:
    \begin{itemize}
        \item Simulate  a set ${\cal V}_{ic}$ with $u(0,\mathbf{x}) \sim {\cal D}^{ic}_{\cal OOD}$, where ${\cal D}^{ic}_{\cal OOD} \neq {\cal D}^{ic}_{\cal T}$   ({\it e.g.} Gaussian, Sinusoidal, random Fourier modes, parametric, etc.).
         \item Simulate  a set ${\cal V}_{bc}$ with $u(t,0)$ and $u(t,L) \sim {\cal D}^{bc}_{\cal OOD}$ by selecting one of the following boundary condition distributions (different from ${\cal D}^{bc}_{\cal T}$): 
    \begin{itemize}
        \item {\it Dirichlet}: solutions are directly imposed at the boundary;
        \item {\it Neumann}: values of the derivative are imposed;
        \item {\it Robin}: combination of both;
        \item {\it Periodic}: derivatives must match at opposite boundaries; 
        \item others.
    \end{itemize}
    \end{itemize}
    \item Generation of two sets of $n_{\cal V}$ OOD \textbf{estimated solutions} from  $PDE_{\cal T}$: :
    \begin{itemize}
    \item  Simulate  a set ${\cal E}_{ic}$ with $u(0,\mathbf{x}) \sim {\cal D}^{ic}_{\cal OOD}$.
    \item  Simulate  a set ${\cal E}_{bc}$ with $u(t,0)$ and $u(t,L) \sim {\cal D}^{bc}_{\cal OOD}$.
    \end{itemize}
    It is important to note that to allow pointwise or trajectory comparisons, ${\cal V}_{ic}$ (resp. ${\cal V}_{bc}$) and ${\cal E}_{ic}$ (resp. ${\cal E}_{bc}$) are simulated at the same $(t,\mathbf{x})$ coordinates.
    \item Compare ${\cal V}_{ic}$ (resp. ${\cal V}_{bc}$) with ${\cal E}_{ic}$ (resp. ${\cal E}_{bc}$) and evaluate whether $PDE_{\cal T}$ still accurately captures the underlying governing law $PDE_{GT}$ under varying conditions. This evaluation can be performed using primarily metrics presented in Sec.~\ref{sec:errors}. 
\end{enumerate}

Note that the previously described protocol is similar to the one  used in PhysPDE~\citep{feng2025physpde} where the authors change the boundary conditions and evaluate their model's OOD generalization capacity through the mean squared residuals.\\

\begin{figure}[t]
    \centering
    \includegraphics[width=0.49\linewidth,height=2.7cm]{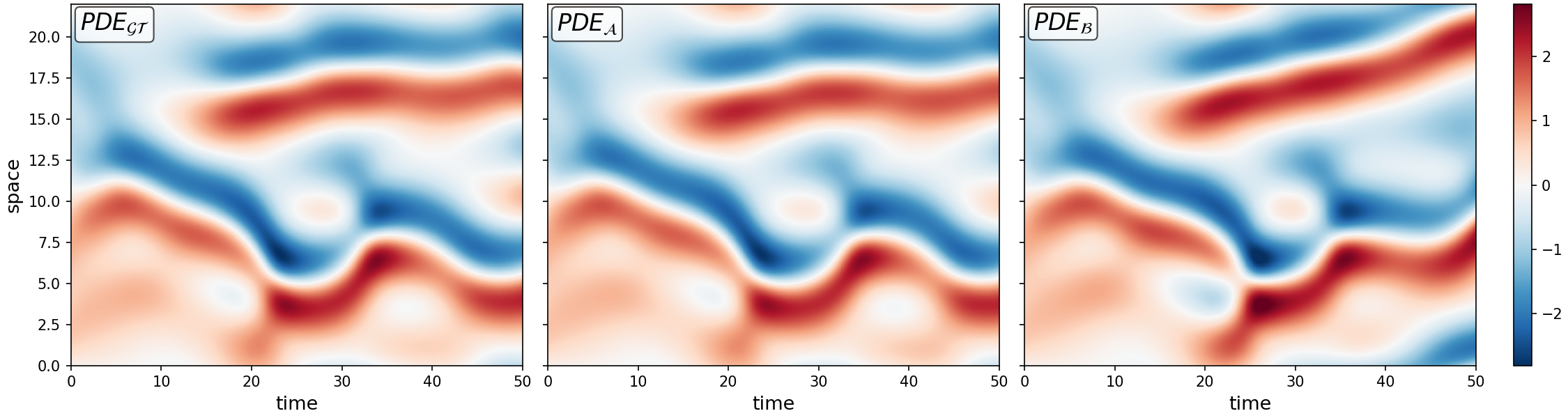}
\includegraphics[width=0.49\linewidth,height=2.7cm]{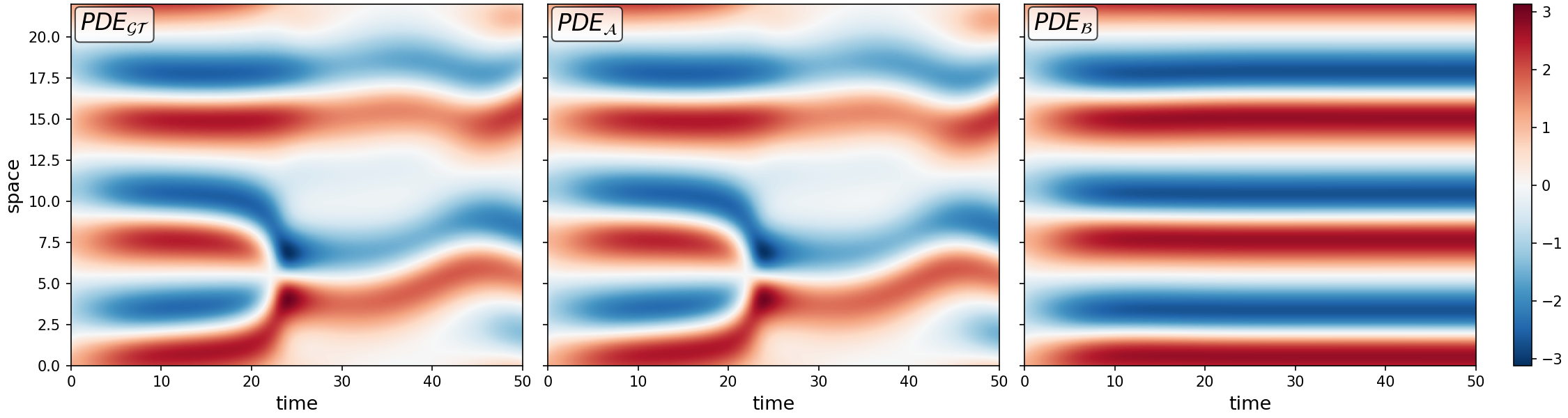}
    \caption{Simulation data obtained from the GT KS equation $PDE_{\cal GT}$, $PDE_{\cal A}$ and $PDE_{\cal B}$ with  two initial conditions different from those used in the running example: (left) $IC_2=\sin(2\pi x/L) + 0.3\sin(6\pi x/L + 0.5)$; (right): $IC_3=\cos(6\pi x/L) + 0.2\cos(10\pi x/L + 0.7)$. We can note that changing the IC while keeping the same analytical expression leads to very different pattern formations.}
    \label{fig:different_ic}
\end{figure}

\noindent \textbf{Long-horizon prediction:} The {\it temporal generalization} is  also widely employed as an OOD scenario. Instead of evaluating the learned PDE on unseen IC and BC, $PDE_{\cal T}$ is simply assessed over longer horizons than those used during the discovery process. In this case, one can resort to the rollout error $\epsilon_{rollout}({\hat u},u)$  (already defined in Eq.~\ref{eq:rollout}) which is particularly well suited for evaluating the long-term stability of $PDE_{\cal T}$ (as used, \textit{e.g.}, in AutoSINDy \citep{basiri2026discoverynonlineardynamicsautomated}). We remind below its definition:
\begin{eqnarray}
    \epsilon^{T'}_{rollout}({\hat u},u)=\frac 1{N_T'} \sum_{i=1}^{N_T'} \epsilon^i_{rollout}({\hat u},u)=\frac 1{N_T'} \sum_{i=1}^{N_T'} ||{\hat u}(i*\delta_t,.)-u(i*\delta_t,.) ||_{L^2(\Omega)}^2 \in \mathbb{R}_{\geq 0},
    \label{eq:rollout2}
\end{eqnarray}
where $N_T'$ is supposed here much higher than the number of temporal steps $N_T$ used for learning $PDE_{\cal T}$.\\

\noindent \textbf{Numerical OOD generalization:} To conclude this section, we mention in the following an approach borrowed from numerical analysis \citep{citeulike:8936200} to investigate the generalization ability of the discovered PDE through the lens of a convergence study, which can be viewed as a form of \textit{numerical distribution shift }evaluation.  
The principle  consists in assessing the mesh convergence by numerically solving the equation at different spatial resolutions. More formally, the following property is expected to hold:
\begin{eqnarray}
Conv_{\mathbf{x}}(\bm{\hat{\alpha}})=0,
    \label{eq:mesh-refinement}
\end{eqnarray}
where $Conv_{\mathbf{x}}(\bm{\hat{\alpha}})=\lim_{{\delta_x} \rightarrow 0}||\hat{u}-\hat{u}_{\delta_x}||$. 
 Eq.~\ref{eq:mesh-refinement} means that as the spatial discretization is progressively refined, the numerical solution converges to a stable limit. This approach can therefore reveal pathological behaviors and help identify PDEs exhibiting low numerical robustness and potential ill-posedness. Note that the convergence analysis can be equally applied to the temporal discretization, checking that:
\begin{eqnarray}
Conv_{t}(\bm{\hat{\alpha}})=0,
\label{eq:time-refinement}
\end{eqnarray}
with $Conv_{t}(\bm{\hat{\alpha}})=\lim_{\delta_t \rightarrow 0}||\hat{u}-\hat{u}_{\delta_t}||.$

\begin{table}[t]
  \caption{Capacity of out-of-distribution generalization of $PDE_{\cal A}$ and $PDE_{\cal B}$ in three scenarios: (i) varying initial conditions $IC_1$, $IC_2$, $IC_3$. The $MAE$ is used  as evaluation metric. Note that $IC_1$ corresponds to the conditions of the running example; (ii) longer horizons at different times $T\geq 50$. The rollout error is evaluated at time $50,100, 200$ and $300$; (iii) mesh and time convergence.} 
  \label{tab:res-OOD}
  \centering
  \begin{tabular}{|l|r|r|}
   \hline 
        & \multicolumn{1}{|c|}{$PDE_{\cal A}$}  &  \multicolumn{1}{|c|}{$PDE_{\cal B}$} \\
    OOD scenarios & \multicolumn{1}{|c|}{{\scriptsize ${\hat u}_t=-0.99uu_{\mathbf{x}}$}} & \multicolumn{1}{|c|}{{\scriptsize ${\hat u}_t=-0.89uu_{\mathbf{x}}-0.9u_{\mathbf{xx}}$}}\\
    &  \multicolumn{1}{|c|}{{\scriptsize $-0.98u_{\mathbf{xx}}-0.985u_{\mathbf{xxxx}}$}} & \multicolumn{1}{|c|}{{\scriptsize $-0.85u_{\mathbf{xxxx}} + 0.0004u-0.0008u_\mathbf{x}u_\mathbf{xx}$}} \\   
    \hline  {\small  $IC_1=\cos(2\pi x/L) + 0.5\cos(4\pi x/L + 0.3)$ } & $0.128$ \greenpoint &  $0.661$ \redpoint \\
    \hline    \small{$IC_2=\sin(2\pi x/L) + 0.3\sin(6\pi x/L + 0.5)$} & $0.126$ \greenpoint &  $0.572$ \redpoint \\
    \hline    {\small  $IC_3=\cos(6\pi x/L) + 0.2\cos(10\pi x/L + 0.7)$} & $0.052$ \greenpoint &  $0.898$ \redpoint \\
    \hline    {\small $\epsilon^{T'=50}_{rollout}({\hat u},u)$ }& 0.817 \greenpoint &  16.6 \redpoint \\
    \hline  {\small $\epsilon^{T'=100}_{rollout}({\hat u},u)$} & 14.1 \greenpoint  &  38.8 \redpoint \\
    \hline  {\small $\epsilon^{T'=200}_{rollout}({\hat u},u)$} & 36.8 \greenpoint  &  55.4 \redpoint \\
    \hline {\small $\epsilon^{T'=300}_{rollout}({\hat u},u)$} & 48.3 \greenpoint  &  61.1 \redpoint \\
    \hline  {\small $Conv_{\mathbf{x}}(\bm{\hat{\alpha}})$} & 
    $0$ \orangepoint &  
    $0$ \orangepoint \\
    \hline  {\small $Conv_{t}(\bm{\hat{\alpha}})$} & 
    $0$ \orangepoint &  
    $0$ \orangepoint \\
    \hline 
  \end{tabular}
\end{table}

\noindent {\bf \underline{Running Example:}} 
The  generalization capacity of $PDE_{\cal A}$ and $PDE_{\cal B}$ in different OOD settings is reported in Table~\ref{tab:res-OOD}. Through the different changes of initial conditions, we can note that $PDE_{\cal B}$, because of its erroneous form, is not always able to accurately extrapolate. The most striking illustration is given by $IC3$ for which the $MAE$ increases significantly (0.898) while $PDE_{\cal A}$ predicts efficiently the correct solutions under these conditions ($MAE$=0.052). The deteriorated behavior of $PDE_{\cal B}$ is  confirmed by the simulation data presented in Fig.~\ref{fig:different_ic} (last panel). We can see that $IC3$ prevents the  pattern formation from this learned equation. In order to understand the reasons behind such undesirable dynamics, we evaluated the individual terms of $PDE_{\cal B}$ under $IC3$. We noticed that $u_{xxxx}$ becomes too small to play the expected role in the  formation of non linear patterns.

\begin{figure}[t]
    \centering
    \includegraphics[width=0.3\linewidth]{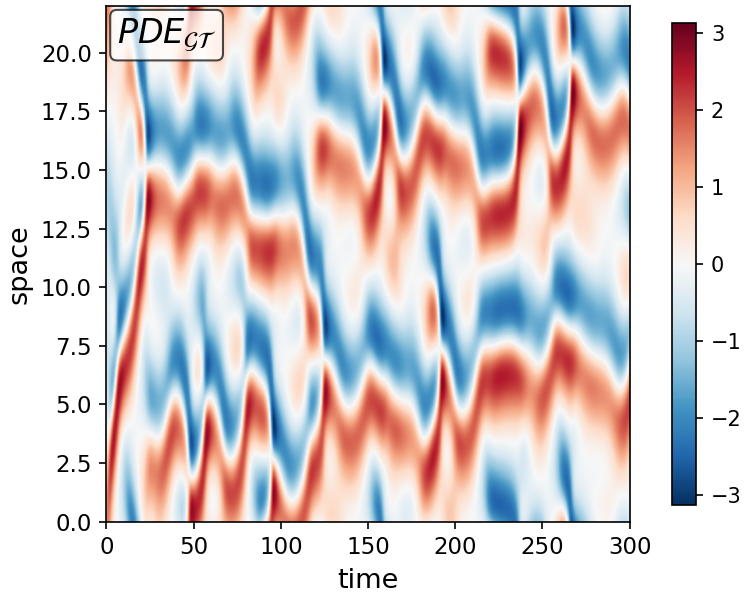}
    \includegraphics[width=0.34\linewidth]{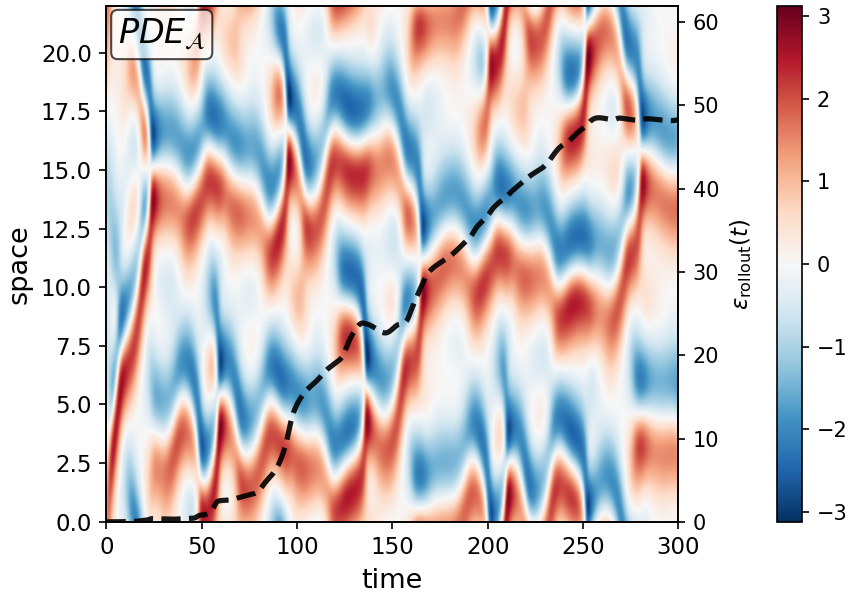}
    \includegraphics[width=0.34\linewidth]{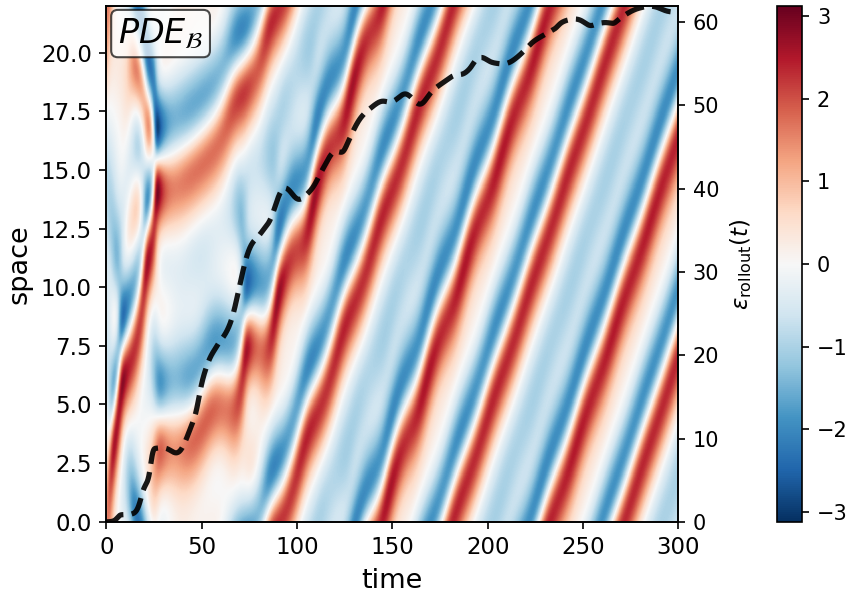} 
    \caption{Long-horizon simulations ($t\in [0,300]$) of the KS $PDE_{\cal GT}$ (left),  $PDE_{\cal A}$ (center) and $PDE_{\cal B}$ (right). The rollout error curve $\epsilon^t_{rollout}({\hat u},u)$ over time (dashed-lines) is also reported for both $PDE_{\cal A}$ and $PDE_{\cal B}$.}
    \label{fig:sim300}
\end{figure}

On the other hand, the analysis of the rollout errors is also very informative. While we already noted that $PDE_{\cal A}$ leads to a much smaller  $\epsilon_{rollout}({\hat u},u)$ in the training time interval $[0,T]$ (see Fig.~\ref{fig:heatmap-KS}), surprisingly the gap between the two PDEs tends to narrow as the time horizon is extended (shifting from a ratio of 20-to-1 to 1.3-to-1). This is confirmed by the simulations of   $PDE_{\cal GT}$, $PDE_{\cal A}$ and $PDE_{\cal B}$ between $[0,300]$ presented in Fig.~\ref{fig:sim300}. Although $PDE_{\cal A}$ appears to be  visually much closer to $PDE_{\cal GT}$ than $PDE_{\cal B}$, with similar patterns still clearly visible after $t=50$, the rollout error (dashed line) reveals that $PDE_{\cal A}$ also deteriorates rapidly, eventually reaching nearly the same error level as $PDE_{\cal B}$ at $T = 300$. This example clearly illustrates how even small discrepancies in the analytical form of a PDE can lead to dramatic errors in extrapolation. More precisely, although $PDE_{\cal A}$ contains the correct terms and its coefficients are very close to those of $PDE_{\cal GT}$, these seemingly minor differences are sufficient to cause a substantial degradation in long-term predictions. We can therefore conclude that the rollout error should play a crucial role in PDE evaluation, which is far from being the case in papers on the subject, regardless of whether or not they use the ground truth $PDE_{\cal GT}$ for this task. Finally, regarding $Conv_{\mathbf{x}}(\bm{\hat{\alpha}})$ and $Conv_{t}(\bm{\hat{\alpha}})$, the numerical convergence of both $PDE_{\cal A}$ and $PDE_{\cal B}$ (in time and space) is satisfied, which is a good point from an evaluation perspective, but is not really helpful for distinguishing between the two equations. \\

\noindent    \fbox{\textbf{PROS} \cmark vs \textbf{CONS} \xmark}
\vspace{0.3cm}

\noindent \cmark All the metrics are informative about the robustness of the equation to OOD situations. They also make it possible to detect pathological situations in which MAE and MSE-based metrics show good behaviors on in-distribution data, while the PDE solutions diverge over longer time horizons due to spurious terms or inaccurate coefficients. Moreover, they are essential when it comes to validating the discovery of a new knowledge/law.
\vspace{0.15cm}

\noindent \xmark  Performing OOD experiments requires either the ground truth $PDE_{\cal GT}$ to be known or access to additional OOD observation data obtained under varying conditions, which is not always possible in real-applications. A fallback solution is to learn neural operators that can simulate, once optimized, new observations under different initial conditions and beyond the time interval used for training.

\begin{center}
\rule{0.1\textwidth}{1pt} {\raisebox{-0.4ex}{\cmark \xmark}}
\rule{0.1\textwidth}{1pt}
\end{center}



\begin{table}[t]
\begin{small}
  \caption{Capacity of the investigated metrics to answer the 5 key questions in PDE evaluation. The last four rows of the table do not, strictly speaking, represent evaluation metrics, but rather physical properties discussed in Sec.~\ref{sec:consistency} that can be used to assess the physical consistency of the PDE.} 
  \label{tab:sota}
  \centering
  \begin{tabular}{l|ccccc}
    \toprule
         & Solution &  Physical     & Interpretation &  Generalization   &  GT PDE  \\
         & Accuracy & Consistency & Sparsity & Capacity & Recovery  \\
    Metrics & {\bf Q1} & {\bf Q2} & {\bf Q3} & {\bf Q4} & {\bf Q5}  \\
    \midrule   $TPR(\bm{\alpha},\bm{\hat{\alpha}})~  \ref{eq:TPR}$  $J(\bm{\alpha},\bm{\hat{\alpha}})~  \ref{eq:Jaccard}$  & \xmark  &  \xmark &   \cmark & \xmark   & \cmark \\
    $P(\bm{\alpha},\bm{\hat{\alpha}})~  \ref{eq:Precision}$  $R(\bm{\alpha},\bm{\hat{\alpha}})~  \ref{eq:Recall}$  & \xmark  &  \xmark &   \cmark & \xmark   & \cmark \\
        $TED~  \ref{eq:TED}$  & \xmark  &  \xmark &   \cmark & \xmark   & \cmark \\
       $LMM$  & \xmark  &  \xmark &   \cmark & \xmark   & \cmark \\
 $SRecov_1(\bm{\alpha},\bm{\hat{\alpha}}),SRecov_2(\bm{\alpha},\bm{\hat{\alpha}})$ \ref{eq:CAS}  & \xmark  &  \xmark &   \xmark & \xmark   & \cmark \\

       $\epsilon^p_{coef}(\bm{\alpha},\bm{\hat{\alpha}})$ \ref{eq:nL2coef}-\ref{eq:maxerror} & \xmark  &  \xmark &   \xmark & \xmark   & \cmark \\
$w\epsilon^2_{coef}(\bm{\alpha},\bm{\hat{\alpha}})$ \ref{eq:wL2coef} & \xmark  &  \cmark &   \xmark & \xmark   & \cmark \\
$NDCG(\bm{\alpha},\bm{\hat{\alpha}})$ \ref{eq:NDCG} & \xmark  &  \cmark &   \xmark & \xmark   & \cmark \\   
$T_G(\bm{\alpha},\bm{\hat{\alpha}})$ \ref{eq:Tanimoto} & \xmark  &  \xmark &   \xmark & \xmark   & \cmark \\      
    $MSE/MAE$ on $u$  \ref{eq:MSE}-\ref{eq:nMSE-bar} & \cmark        & \xmark & \xmark & \cmark  & \xmark \\
    $MSE$ on $u_t$  \ref{eq:n.utMSE} & \xmark        & \cmark & \xmark & \cmark  & \xmark \\
      $\overline{f}_{MSE}({\cal F}(\hat{u}),{\cal F}({u}))$   \ref{eq:mf_{MSE}} & \cmark        & \cmark & \xmark & \cmark  & \xmark \\
      $\epsilon_{rollout}({\hat u},u)$   \ref{eq:rollout} & \cmark        & \cmark & \xmark & \cmark  & \xmark \\   
    $S_{terms}(\bm{\hat{\alpha}})$ \ref{eq:sparsity}    & \xmark        & \xmark & \cmark & \xmark  & \xmark \\
    ${\cal C}(\bm{\hat{\alpha}}^\top.\Theta)$ \ref{eq:ExpTree} ${\cal C}_2(\bm{\hat{\alpha}}^\top.\Theta)$ \ref{eq:ExpTreeV2}  & \xmark        & \xmark & \cmark & \xmark  & \xmark \\
    $Score(\bm{\hat{\alpha}})$ \ref{eq:Score}   & \cmark        & \xmark & \cmark & \xmark  & \xmark \\
    $Reward_{1,2}(\bm{\hat{\alpha}})$ \ref{eq:reward1} \ref{eq:reward2}  & \xmark   & \cmark & \cmark & \xmark & \xmark \\
    $AIC$~\ref{eq:cAIC}, $BIC$~\ref{eq:BIC}   & \xmark      & \cmark & \cmark & \xmark  & \xmark \\
    $PIC$~\ref{eq:PIC}   & \cmark      & \cmark & \cmark & \xmark  & \xmark \\
    $MDL$~\ref{eq:MDL-Feynman} \ref{eq:mdl-SymLang}  & \cmark      & \xmark & \cmark & \xmark  & \xmark \\
    $\epsilon^{T'}_{rollout}({\hat u},u)$~\ref{eq:rollout2}  & \cmark      & \cmark & \xmark & \cmark  & \xmark \\ 
    $Conv_{\mathbf{x}}(\bm{\hat{\alpha}})$~\ref{eq:mesh-refinement}  & \xmark      & \cmark & \xmark & \cmark  & \xmark \\
    $Conv_{t}(\bm{\hat{\alpha}})$~\ref{eq:time-refinement}   & \xmark      & \cmark & \xmark & \cmark  & \xmark \\
  Physical properties~\ref{sec:consistency}: & \xmark        & \cmark & \xmark & \cmark  & \xmark \\
  \textit{Symmetries, Conservation Laws}   &      &  &  &  &  \\
   \textit{Residuals, Dimensional Consistency }  &      &  &  &  &  \\
    \textit{Positivity Preservation, Convection}  &      &  &  &  &  \\
    \textit{Energy Dissipation, etc.} &      &  &  &  &  \\
    \bottomrule
  \end{tabular}
\end{small}
\end{table}

\textcolor{black}{The taxonomy proposed in this section has made it possible to group evaluation metrics that were previously scattered throughout the literature. It has also helped clarify the type of information they capture, as well as the limitations of using them in isolation from one another. Table \ref{tab:sota} summarizes our analysis and shows the metrics that address each of the questions \textbf{Q1}$\rightarrow$\textbf{Q5}. This table confirms that these metrics are designed to address only certain facets of the evaluation problem, and that selecting only one or two of them may lead to erroneous conclusions regarding the relevance of the PDE.}

\section{Experiments}
\label{sec:expePDEFIND-DEEpMoD}

\textcolor{black}{Throughout Sections \ref{sec:GT} and \ref{sec:noGT}, the Kuramoto–Sivashinsky running example not only illustrated how the different metrics in the proposed taxonomy operate, but also confirmed their ability to properly discriminate between $PDE_{\cal A}$ and $PDE_{\cal B}$ and to unambiguously identify the former as providing a closer approximation to the ground-truth $PDE_{\cal GT}$. However, by construction, $PDE_{\cal A}$ was substantially closer to $PDE_{\cal GT}$ than $PDE_{\cal B}$. Therefore, in order to draw sufficiently general conclusions that can be leveraged to develop an operational evaluation protocol, it is important to extend our study to real scenarios. In what follows, we apply a set of metrics on PDEs learned by actual discovery algorithms, and assess their ability to distinguish between equations that might be close to another. 
To this end, we conduct experiments using two PDE discovery algorithms (PDE-FIND \citep{rudy2016datadrivendiscoverypartialdifferential} and DeepMoD \citep{Both_2021}) on four PDEs (Burgers, Advection–diffusion, Korteweg–de Vries and Kuramoto-Sivashinsky). The analysis of the results provides particularly interesting insights into the risks of relying on one or two metrics in isolation, without covering the other aspects that a discovered PDE should satisfy, especially the adaptability to OOD situations. All the experiments have been conducted using our open-source Python \href{https://github.com/bapt-mat/PostHoc-PDE-Evaluation}{toolkit}.}

\subsection{Experimental setup and Toolkit}
\textcolor{black}{In the following, we present the PDEs on which we will work, as well as the experimental protocol used throughout our experiments. We also recall some of the characteristics of the two PDE discovery algorithms that we implemented. We conclude this section by presenting the main features of our Python-based toolkit, which allows users to reproduce the experiments and to make use of the various metrics discussed throughout the article.}
\subsubsection{Experimental setup}
\textcolor{black}{
The experiments are performed on the following PDEs that cover a variety of dynamics (high-order derivatives, instability, nonlinearity, advection, steep gradients):
\begin{itemize}
    \item \textbf{Burgers' equation}: It models the balance between nonlinearity (convection) and diffusion (viscosity). It serves as a simplified model for understanding complex dynamics (as in fluid mechanics) and is defined as follows: $u_t = - 1.0 uu_\mathbf{x} + 0.05 u_{\mathbf{xx}}$.
    \item \textbf{Advection-Diffusion equation}: It describes the transport of a substance by fluid motion alongside molecular spreading. It serves as a foundational model for mass, heat, and pollutant dispersion phenomena and is given by: $u_t = - 1.0u_\mathbf{x} + 0.1 u_{\mathbf{xx}}$.
    \item \textbf{Korteweg-de Vries (KdV) equation}: It characterizes wave phenomena where nonlinear steepening and wave dispersion interact, making it a standard model for the evolution of water waves. It is expressed as follows: $u_t =  - 6.0 uu_\mathbf{x} -1.0 u_{\mathbf{xxx}}$.
    \item \textbf{Kuramoto-Sivashinsky (KS) equation}: As already discussed, it is known for exhibiting spatiotemporal chaos. It is widely used to model complex pattern formation and instabilities in various physical systems. It takes the following fourth-order form: $u_t =  - 1.0 uu_x -1.0 u_{xx} - 1.0u_{xxxx}$.
\end{itemize}
We use the numerical time-integration scheme ETD2-RK and solve the equations using the parameters reported in \Cref{tab:domain-params}.
$N_x$ represents the number of points of the spatial domain, $L$ the length of this domain, $\delta_t$ the time-step used by the solver, $T$ the total time of the simulation, $N_t$ the number of time-steps, and $N$ is the number of training points uniformly sampled on this simulation used by the algorithms. We use $N_x \times \delta_t \times T$ test data for assessing the pointwise metrics. 
The equations are solved on periodic domains and using the initial conditions reported in \Cref{tab:ic-params}. For each algorithm, and each PDE, the dictionary of candidate terms consists of spatial partial derivatives of the field $u$ up to fourth order, \textit{i.e.}, from $u_{\mathbf{x}}$ to $u_{\mathbf{xxxx}}$. 
\begin{table}[t]
    \centering
    \begin{tabular}{|c|c|c|c|c|c|c|}
    \hline
         PDE & $N_x$& $L$ & $\delta t$ & $T$ & $N_t$ &  $N$ (training points) \\
         \hline
         Burgers & $64$ & $2\pi$  & $0.01 $& $2.0$ & $201$ & $2176$\\
         \hline
         Advection-Diffusion & $126$ & $2\pi$ & $0.01$ & $2.5$ & $251$ & $3213$\\
         \hline
         KdV & $600$ & $60$ & $0.01$ & $20.0$ & $2001$ & $6966$\\
         \hline
         KS & $220$ & $22$ & $0.01$ & $50.0$ & $5001$ & $2805$\\
         \hline
    \end{tabular}
    \caption{\textcolor{black}{Parameters for the four PDEs used in the experiments. $N_x$ represents the number of points of the spatial domain, $L$ the length of this domain, $\delta_t$ the time-step used by the solver, $T$ the total time of the simulation, $N_t$ the number of time-steps, and $N$ is the number of training points.}}
    \label{tab:domain-params}
\end{table}
\begin{table}[ht]
    \centering
    \begin{tabular}{|c|l|}
    \hline
         PDE & Initial condition \\
         \hline
         \multirow{2}{*}{Burgers} & Training: $u_0(x) = \cos(x)$ \\
         & OOD: $u_0(x) = -1.5\tanh\!\left(\frac{x - L/3}{0.2}\right) + 1.5\tanh\!\left(\frac{x - 2L/3}{0.2}\right)$ \\
         \hline
         \multirow{2}{*}{Advection-Diffusion} & Training: $u_0(x) = \exp\!\left(-\left(\frac{((x - L/4 + L/2) \bmod L) - L/2}{0.4}\right)^2\right)$ \\
         & OOD: $u_0(x) = 6\exp\!\left(-\left(\frac{((x - L/4 + L/2) \bmod L) - L/2}{0.4}\right)^2\right)$ \\
         \hline
         \multirow{2}{*}{KdV} & Training: two-soliton ($c_0{=}1.0,\, c_1{=}0.5$, (\href{http://lie.math.brocku.ca/~sanco/solitons/kdv_solitons.php}{ref.})) \\
         & OOD: $u_0(x) = 0.45\,\mathrm{sech}^2\!\left(\frac{x - L/2}{3.0}\right)$ \\
         \hline
         \multirow{2}{*}{KS} & Training: $u_0(x) = \cos \left(\frac{2\pi x}{L}\right) + 0.5\cos \left(\frac{4\pi x}{L} + 0.3\right)$ \\
         & OOD: $u_0(x) = \sin \left(\frac{2\pi x}{L}\right) + 0.3\sin \left(\frac{6\pi x}{L} + 0.5\right)$\\
         \hline
    \end{tabular}
    \caption{\textcolor{black}{Training and OOD initial conditions for the four PDEs used in the experiments.}}
    \label{tab:ic-params}
\end{table}
It also includes the independent variables $x$ and $t$, the field $u$, and products of these quantities and spatial derivatives, such as $u u_{\mathbf{x}}$ and $t u_{\mathbf{xx}}$. 
Finally, the dictionary includes monomials as products between spatial derivatives, such as $u_{\mathbf{x}}u_{\mathbf{xx}}$ or $u_{\mathbf{xxx}}^2$.
Using the training data, PDEs are learned from the following  discovery algorithms:
\begin{itemize}
    \item \textbf{PDE-FIND \citep{rudy2016datadrivendiscoverypartialdifferential}} which frames PDE discovery as a sparse regression problem. 
    It builds a library of candidate mathematical terms from the data and then uses the Sequentially Thresholded Least Squares (STLSQ) algorithm to identify a sparse set of terms defining the equation.
    Spatial derivatives are estimated using finite differences for Burgers' and KS equations, and splines for KdV and advection-diffusion. 
    The temporal derivative is estimated using finite differences. 
    STLSQ combines ridge (L2) regularization with iterative thresholding to promote sparsity. 
    The sparsity threshold is set to 0.0015, 0.005, 0.05, 0.1 for Burgers, advection-diffusion, KdV, and KS, respectively. 
    The ridge regularization parameter is set to $10^{-8}$ by default, with values of $10^{-4}$ for KdV, and $10^{-3}$ for the advection-diffusion equation.
    \item \textbf{DeepMoD \citep{Both_2021}} which combines neural network function approximation, automatic differentiation, and sparse regression for PDE discovery.
    A multilayer perceptron maps the spatio-temporal coordinates ($t,x$) to the field $u$, using four hidden layers of $30$ neurons ($40$ for KdV).
    The same candidate library as in PDE-FIND is constructed from the neural network representation of $u$, with all derivatives (spatial and temporal) computed through automatic differentiation.
    Sparse regression is performed through iterative sparsification, where terms with coefficients below a threshold are removed.
    The thresholds are set to 0.02, 0.02, 0.04, and 0.5 for Burgers, advection-diffusion, KdV, and KS, respectively.
    The neural network is optimized using Adam with a learning rate of $10^{-3}$ for a maximum of $100,000$ iterations.
\end{itemize}
 Concerning the implementation, we rely on the corresponding packages for each of the algorithms (\href{https://github.com/dynamicslab/pysindy}{PySINDy}, \href{https://github.com/PhIMaL/DeePyMoD}{DeePyMoD}) with minor modifications to adapt to our settings. 
 For each PDE learned by PDE-FIND and DeepMoD, we compute the behavior of a subset of representative metrics discussed in this article: $nMAE(\hat{u},u)$~\ref{eq:nMAE} is used as a representative of the category of metrics assessing the prediction errors; $TPR(\hat{\alpha},\alpha)$~\ref{eq:TPR} and $\epsilon^1_{coef}(\hat{\alpha},\alpha)$~\ref{eq:nL1coef} are employed for checking the capacity to recover the structural form and the coefficients of the GT PDE, respectively; $\epsilon^{T}_{rollout}({\hat{u}},u)$~\ref{eq:rollout} is used to estimate the cumulative error over the training time interval, but also on OOD data over long-horizons and on different initial conditions. This is used to assess the extrapolation capabilities of the learned PDEs. All these metrics are computed using the toolkit that we specifically developed and present in the following section.
\subsubsection{Toolkit}
In order to facilitate reproducibility, standardization and application of the metrics involved in our proposed taxonomy, we provide an open-source python toolkit. The latter implements most of the surveyed metrics (32 metrics are provided in the toolkit), and is organized into five core evaluation modules: terms recovery (\Cref{sec:terms}), coefficient recovery (\Cref{sec:coef}), prediction errors (\Cref{sec:errors}), sparsity (\Cref{sec:sparsity}), prediction-sparsity trade-offs (\Cref{sec:tradeoff-sparsity-accuracy}), and generalization (\Cref{sec:generalization}).
For practical reasons, such as high implementation complexity (\Cref{eq:Lyapunov-error}) or the requirement for highly specific architectures (\Cref{eq:PIC}), a few metrics discussed are not implemented in this release. In the latter case, the interested reader is referred to the original papers. 
As input, the toolkit expects a plain-text configuration file specifying a candidate equation as a list of mathematical terms and their corresponding coefficients.
The practitioner can also use the simulation file in order to create visualization plots such as solution fields, spatio-temporal error heatmaps or spectral error profiles.
The repository, along with a more detailed documentation and examples, is available from the following \href{https://github.com/bapt-mat/PostHoc-PDE-Evaluation}{GitHub}.
}

\begin{table}[t]
  \caption{Results on learned PDEs from both PDE-FIND and DeepMoD for the Burgers' equation. A subset of representative evaluation metrics is used, covering (i) the prediction errors ($nMAE$), (ii) the ability to recover the symbolic form ($TPR$) and the coefficients ($\epsilon^1_{coef}$) of the GT PDE, (iii) the short and long-term rollout errors at different time horizons ($\epsilon^{T'}_{rollout}$) to evaluate  the cumulative error on both in- and out-of-distribution data, and on changing initial conditions ($IC_2$); and (iv) the respect of physical properties.} 
  \label{tab:res-burgers}
  \centering
  \begin{tabular}{|l|r|r|}
   \hline 
    & \multicolumn{1}{|c|}{PDE-FIND}  &  \multicolumn{1}{|c|}{DeepMoD} \\
    & \multicolumn{1}{|c|}{{\scriptsize ${\hat u}_t=-1.0136uu_{\mathbf{x}} + 0.0337 u_{\mathbf{xx}}$}} 
    & \multicolumn{1}{|c|}{{\scriptsize ${\hat u}_t=-0.9671uu_{\mathbf{x}} + 0.0355 u_{\mathbf{xx}}$}}\\
    & \multicolumn{1}{|c|}{{\scriptsize $-0.0157u_\mathbf{x}u_{\mathbf{xx}} -0.0144 u +  0.0078 tu$}} 
    & \multicolumn{1}{|c|}{{\scriptsize $-0.0123tu - 0.0058xu + 0.0072 tu_{\mathbf{xx}} -0.0029xu_{\mathbf{xx}}$}} \\
    \hline
    $nMAE(\hat{u},u)$~\ref{eq:nMAE} & $0.0095$ \redpoint &  $0.0073$ \greenpoint \\
    \hline
    $\epsilon^{T'=2}_{rollout}({\hat{u}},u)$~\ref{eq:rollout} & $0.0017$ \redpoint &  $0.0004$ \greenpoint \\
    \hline
    $TPR(\hat{\alpha},\alpha)$~\ref{eq:TPR} & $2/5$ \greenpoint &  $1/3$ \redpoint \\
    \hline
    $\epsilon^1_{coef}(\hat{\alpha},\alpha)$~\ref{eq:nL1coef} & $0.0646$ \greenpoint &  $0.0720$ \redpoint \\
    \hline
    $\epsilon^{T'=40}_{rollout}({\hat{u}},u)$~\ref{eq:rollout} & $0.2369$ \redpoint &  $0.0157$ \greenpoint \\
    \hline
   $IC_2:\epsilon^{T'=20}_{rollout}$ & $5.1622$ \redpoint &  $2.4512$ \greenpoint \\
    \hline
    Phys. consistency & \xmark violation of time invariance& \xmark violation of spatial/time invariance \\
    \hline
  \end{tabular}
\end{table}

\begin{figure}[t]
\centering
\begin{minipage}[c]{0.16\linewidth}
    \centering
    \includegraphics[width=\linewidth]{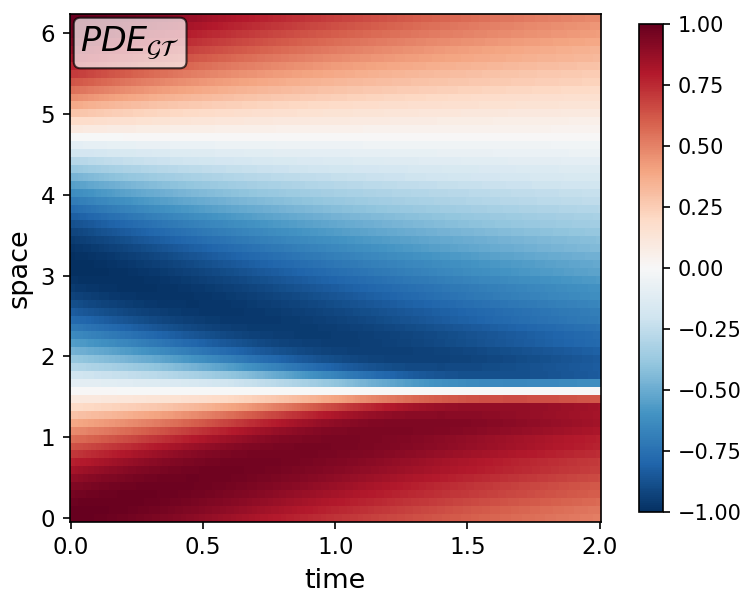}
    \includegraphics[width=\linewidth]{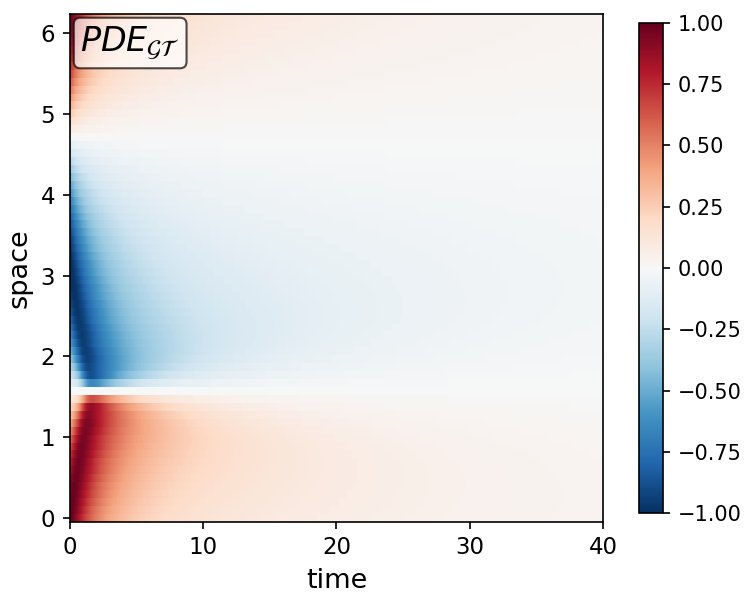}
    \captionof*{subfigure}{GT ($T=2,40$)}
\end{minipage}
\hfill
\begin{minipage}[c]{0.1\linewidth}
    \centering
    \vspace{-1.2cm}
    \textbf{\scriptsize PDE-FIND} \\
    \vspace{1.8cm}
    \textbf{\scriptsize DeepMoD}
\end{minipage}
\hfill
\begin{minipage}[c]{0.16\linewidth}
    \centering
    \includegraphics[width=\linewidth]{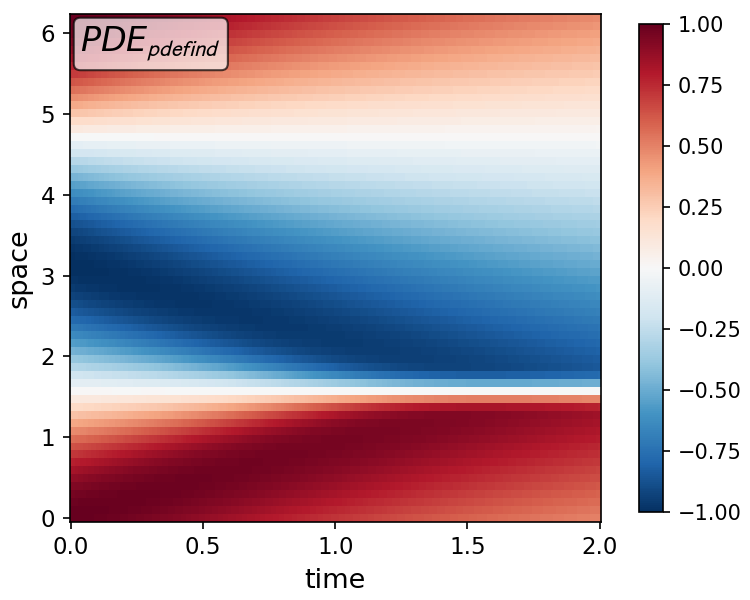}
    \includegraphics[width=\linewidth]{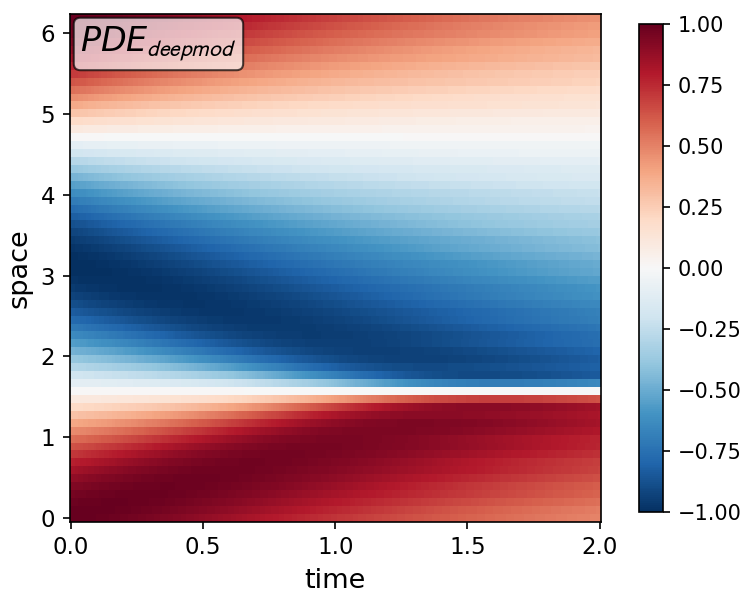}
    \captionof*{subfigure}{$T=2$}
\end{minipage}
\hfill
\begin{minipage}[c]{0.16\linewidth}
    \centering
    \includegraphics[width=\linewidth]{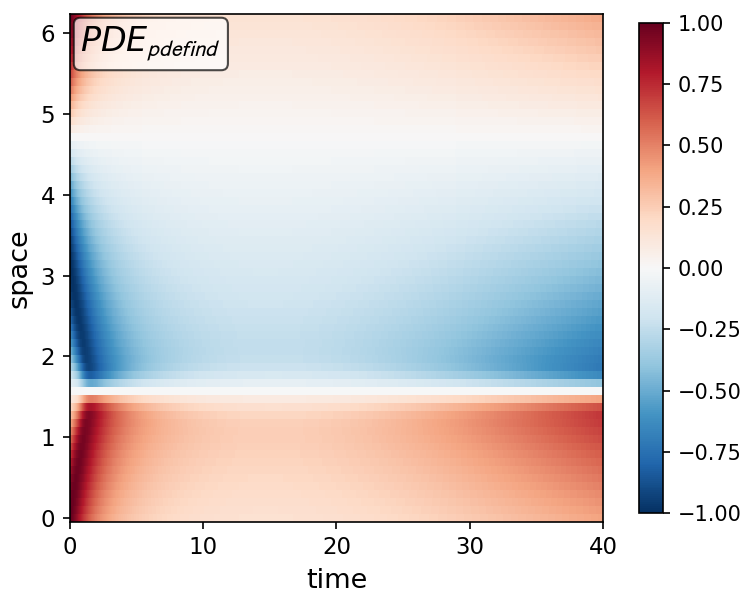}
    \includegraphics[width=\linewidth]{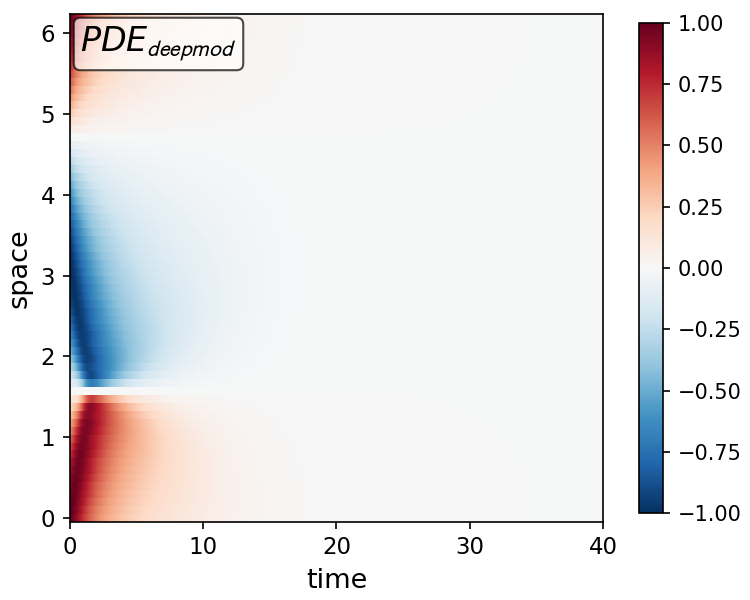}
    \captionof*{subfigure}{$T=40$}
\end{minipage}
\hfill
\begin{minipage}[c]{0.16\linewidth}
    \centering
    \includegraphics[width=1.13\linewidth]{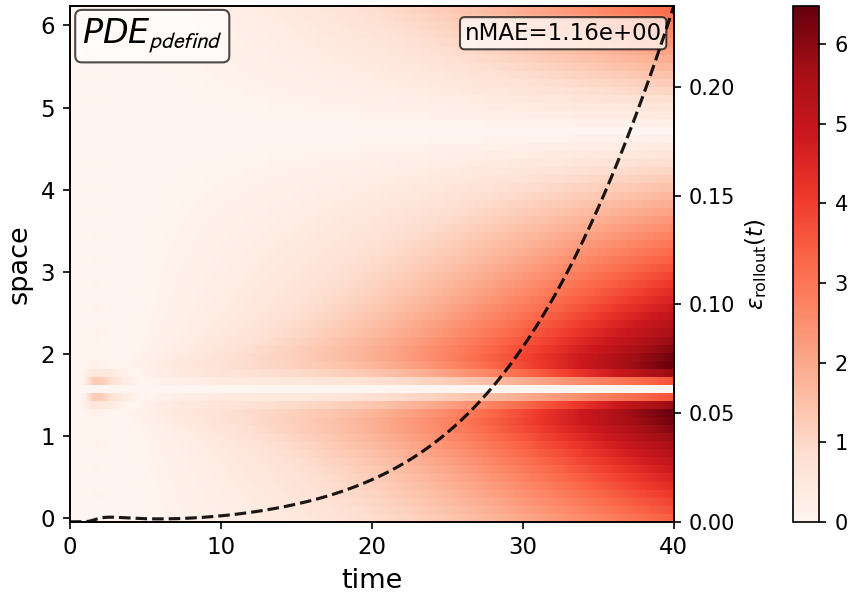}
    \includegraphics[width=1.15\linewidth]{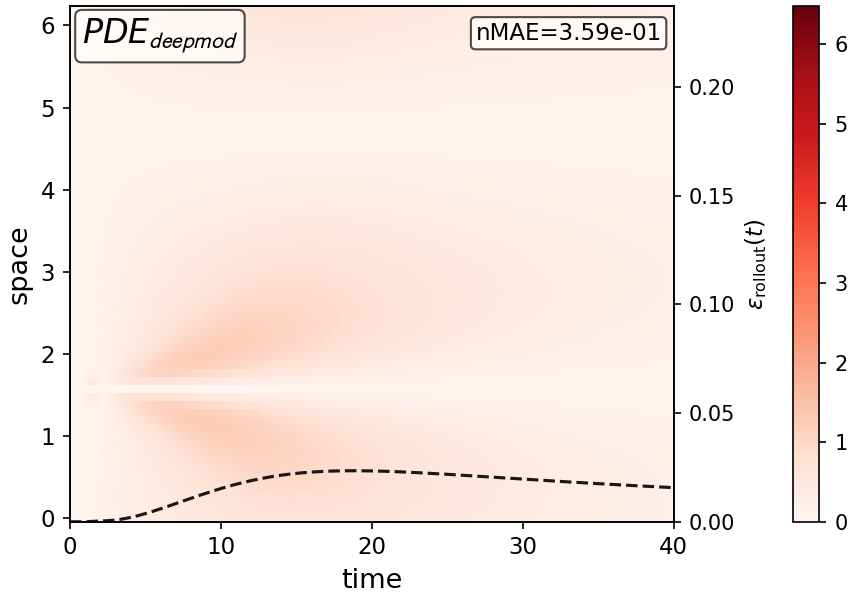}
    \captionof*{subfigure}{Rollout ($T=40$)}
\end{minipage}
\hfill
\begin{minipage}[c]{0.21\linewidth}
    \centering
    \includegraphics[width=0.86\linewidth]{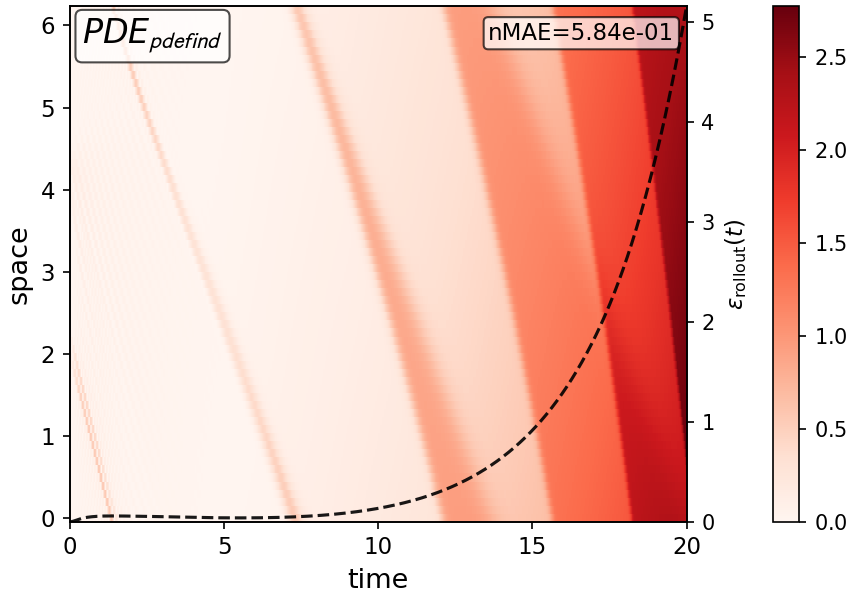}
    \includegraphics[width=0.86\linewidth]{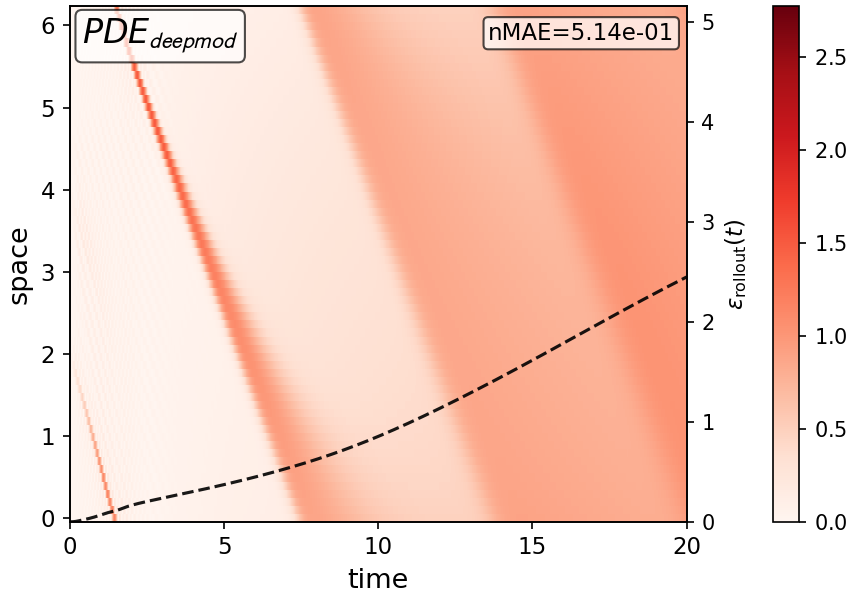}
    \captionof*{subfigure}{$IC_2$: rollout at $T=20$}
\end{minipage}
\caption{Burgers GT equation (left column, simulated up to $T=2$ (top-left) and $T=40$ (bottom-left)); Results with PDE-FIND (top row), and DeepMod (bottom row)  up to $T=2$ (second column) and up to $T=40$ (third); Error heatmaps and rollout curves up to $T=40$ (fourth). Error heatmaps and rollout curves  with different initial conditions ($IC_2$) (fifth).}
\label{fig:burgers_pdefind_deepmod}
\end{figure}

\subsection{Analysis of the results}
\textcolor{black}{We report the results obtained by the two PDE discovery algorithms on the four considered PDEs in the following \Cref{tab:res-burgers,tab:res-adv-diff,tab:res-kdv,tab:res-ks}.
We emphasize here that the goal is not to evaluate the intrinsic ability of each algorithm to recover these benchmark equations in the form of learned PDEs as close to the GT as possible, as their capabilities have already been demonstrated in the respective original papers.
Therefore, we did not push the hyperparameter tuning to its limits, in order to allow for the generation of imperfect PDEs that are nevertheless actually learned by the algorithms. 
This allows us to focus on the behavior of the metrics applied on discovered equations rather than the algorithms themselves. 
%
The results illustrate an interesting recurring scenario in which the ranking  of the learned PDEs (highlighted by the green and red dots) depends on the metric being considered. 
Let us take \Cref{tab:res-burgers}, obtained for the Burger's equation, as an example.  We see that the metrics designed to assess the quality of the predictions of the field $u$ over the training time interval $[0,2]$ ($nMAE(\hat{u},u)$ and $\epsilon^{T'=2}_{rollout}({\hat{u}},u)$, related to \textbf{Q1}) favor the equation produced by DeepMoD.
However, this ranking is reversed when using metrics which assess the algorithm's ability to recover the symbolic form and coefficients of the GT equation ($TPR$ and $\epsilon^1_{coef}(\hat{\alpha},\alpha)$, related to \textbf{Q5}). Ranking inversions (sometimes in the opposite direction) can be observed for the Advection-Diffusion (Table \ref{tab:res-adv-diff}), KdV (Table \ref{tab:res-kdv}) and Kuramoto Sivashinsky (Table \ref{tab:res-ks}) equations, with a special case for the latter, where $TPR$ and $\epsilon^1_{coef}(\hat{\alpha},\alpha)$ even disagree, the former favoring DeepMoD while the latter indicating a preference for PDE-FIND.}

\textcolor{black}{These examples show that the ranking of PDEs can be highly volatile when the metrics only assess the GT recovery capacity and the interpolation ability on in-distribution data, especially when comparing PDEs that appear quite similar at first glance. To draw more robust conclusions, it is therefore \textbf{necessary to also rely on metrics that measure out-of-distribution generalization capabilities} (related to \textbf{Q4}). In this regard, the long-horizon rollout error and the error resulting from a change in initial conditions appear to be complementary measures for assessing the stability and consistency of the equations.          
A representative example is reported in \Cref{tab:res-burgers}, where PDE-FIND applied to the Burgers' equation uses the monomial $tu$.
Although physically inconsistent (Burgers PDE is known to be time invariant), over the training time window (up to $T=2$), this term helps minimize the local prediction error, resulting in misleadingly good $nMAE$ and $\epsilon^{T=2}_{rollout}$ scores (both close to $0$). 
This behavior is confirmed by the heatmaps of \Cref{fig:burgers_pdefind_deepmod} (top row) showing a high similarity between the GT (left column, top figure) and the predictions (second column). 
However, by breaking time-invariance and injecting energy into a dissipative system, this term violates fundamental physical consistency (related to \textbf{Q2}). The long-horizon rollout metric successfully exposes this hidden flaw as it explodes at $T=40$ (dashed-line curve, fourth column). This is confirmed by the predictions (third column) which differ from the true dynamics. Interestingly, the very same rollout error  $\epsilon^{T'=40}_{rollout}({\hat{u}},u)$ is not sufficient to detect alone the physical inconsistency caused by the presence of terms that violate invariance properties in the equation learned by DeepMoD. The reason is that the predictions close to zero produced by this learned  PDE after $T=10$ are consistent with the dissipative nature of the system, allowing it to maintain an overly optimistic low cumulative error. It is through the use of a different initial condition ($IC_2$) that the presence of the spurious terms is detected, as the corresponding rollout eventually diverges. Note that while a decrease in performance is not systematically observed for every new initial condition (it depends on the mathematical nature of the spurious terms), observing such a failure once is sufficient to conclude that the discovered PDE is not a consistent law.}

\begin{figure}[t]
\centering
\begin{minipage}[c]{0.16\linewidth}
    \centering
    \includegraphics[width=\linewidth]{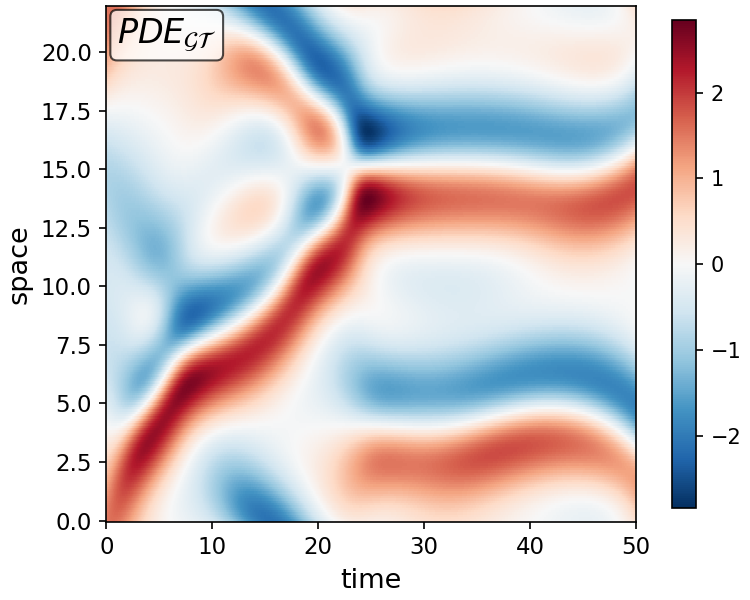}
    \includegraphics[width=\linewidth]{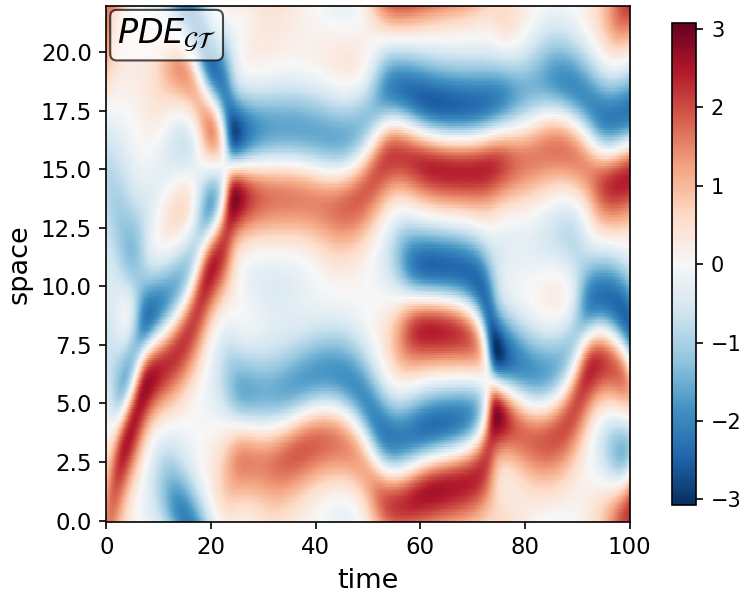}
    \captionof*{subfigure}{GT ($T=50,100$)}
\end{minipage}
\hfill
\begin{minipage}[c]{0.1\linewidth}
    \centering
    \vspace{-1.2cm}
    \textbf{\scriptsize PDE-FIND} \\
    \vspace{1.8cm}
    \textbf{\scriptsize DeepMod}
\end{minipage}
\hfill
\begin{minipage}[c]{0.16\linewidth}
    \centering
    \includegraphics[width=\linewidth]{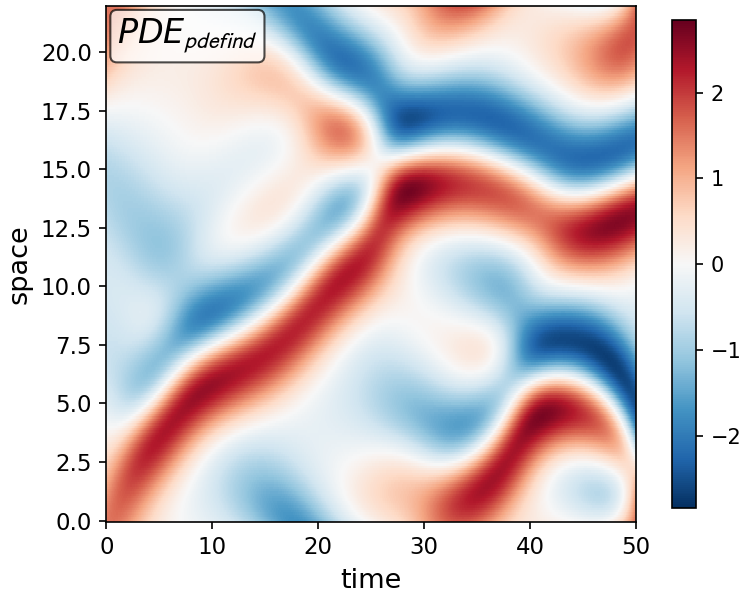}
    \includegraphics[width=\linewidth]{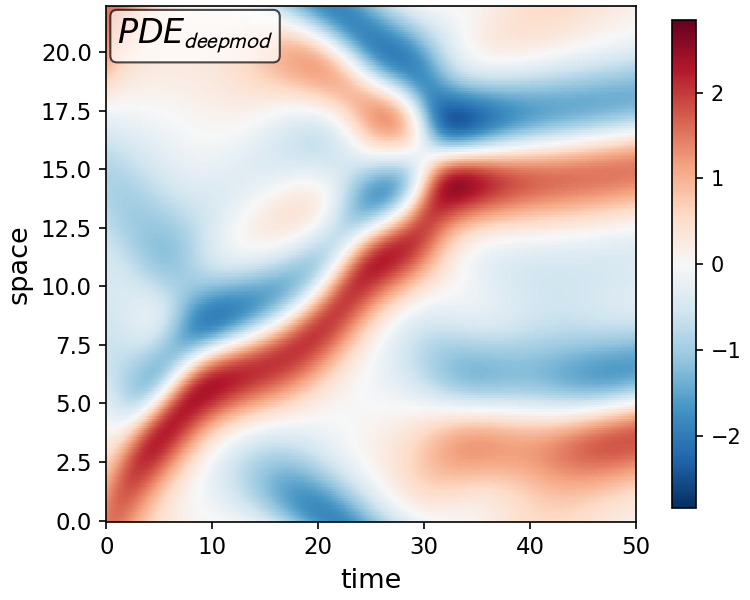}
    \captionof*{subfigure}{$T=50$}
\end{minipage}
\hfill
\begin{minipage}[c]{0.16\linewidth}
    \centering
    \includegraphics[width=\linewidth]{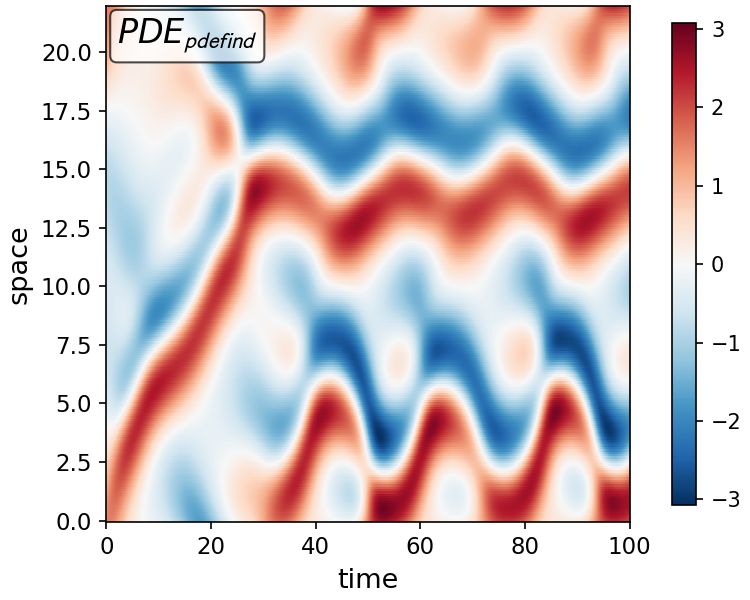}
    \includegraphics[width=\linewidth]{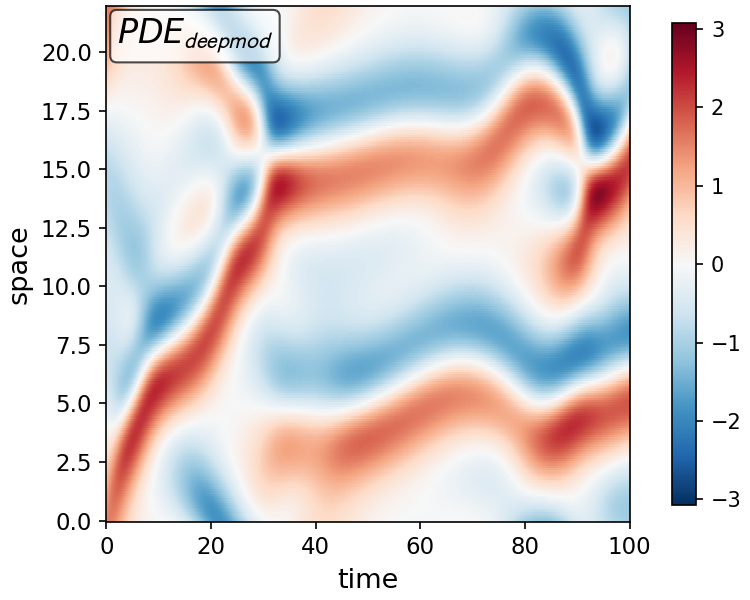}
    \captionof*{subfigure}{$T=100$}
\end{minipage}
\hfill
\begin{minipage}[c]{0.16\linewidth}
    \centering
    \includegraphics[width=1.13\linewidth]{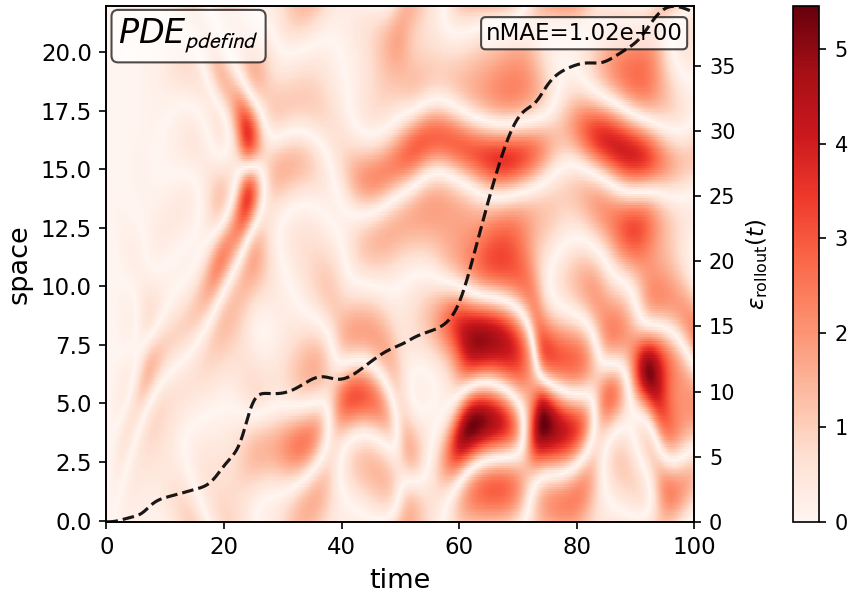}
    \includegraphics[width=1.12\linewidth]{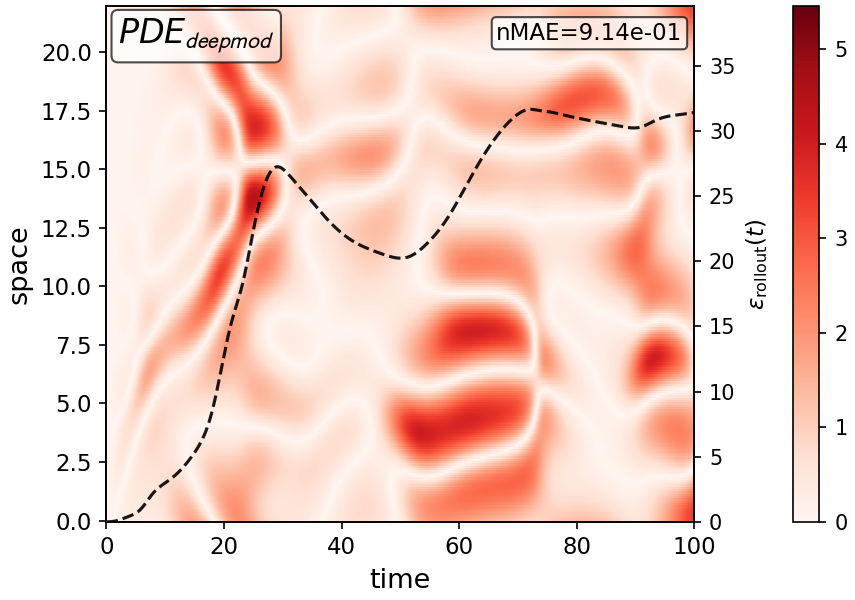}
    \captionof*{subfigure}{\mbox{rollout ($T=100$)}}
\end{minipage}
\hfill
\begin{minipage}[c]{0.21\linewidth}
    \centering
    \includegraphics[width=0.86\linewidth]{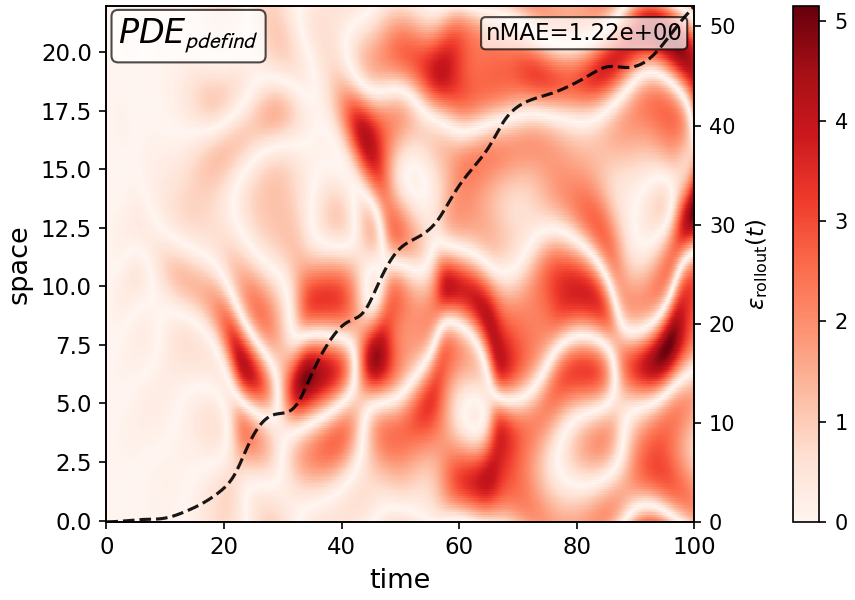}
    \includegraphics[width=0.86\linewidth]{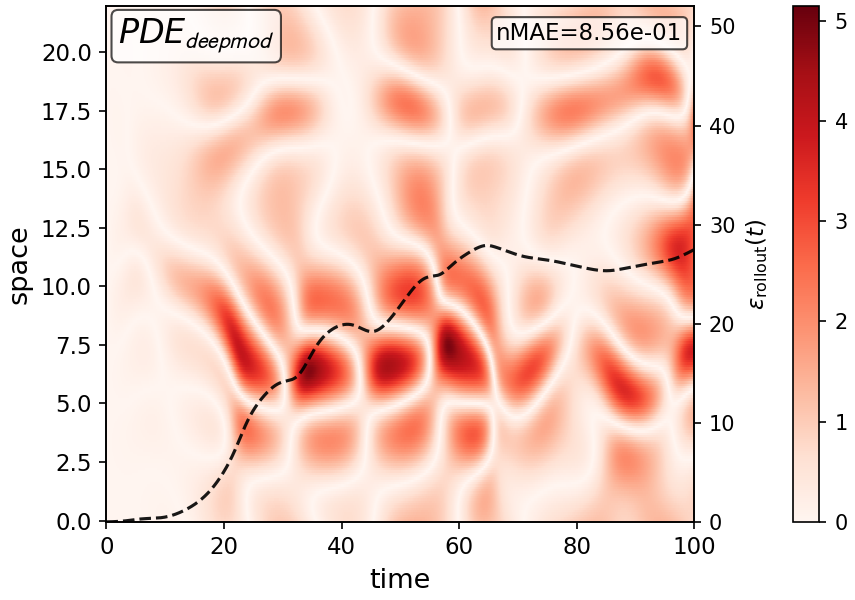}
    \captionof*{subfigure}{\mbox{$IC_2$: rollout ($T=100$)}}
\end{minipage}
\caption{Kuramoto-Sivashinsky GT equation (left column, simulated up to $T=50$ (top-left) and $T=100$ (bottom-left)); Results with PDE-FIND (top row), and DeepMoD (bottom row)  up to $T=50$ (second column) and up to $T=100$ (third); Error heatmaps and rollout curves up to $T=100$ (fourth). Error heatmaps and rollout curves  with different initial conditions ($IC_2$) (fifth).}
\label{fig:ks_pdefind_deepmod}
\end{figure}

\textcolor{black}{The complementarity of the two OOD measures is also observed for the Advection-Diffusion’s equation (Table \ref{tab:res-adv-diff}). Their behaviors for the Kuramoto-Sivashinsky are also graphically reported on Figure \ref{fig:ks_pdefind_deepmod}. Across all tables, it is worth noting that the long-horizon rollout error and the error resulting from a change in initial conditions consistently agree, yielding the same ranking of the PDEs.
These results emphasize that relying 
on metrics considered in isolation from the others may lead to overstated conclusions and that a multi-faceted evaluation is mandatory to validate a discovered equation, even more so when it concerns a potential new  physical law.
}

\begin{table}[t]
  \caption{Results on learned PDEs from both PDE-FIND and DeepMoD for the Advection-Diffusion  equation. A subset of representative evaluation metrics is used, covering (i) the prediction errors ($nMAE$), (ii) the ability to recover the symbolic form ($TPR$) and the coefficients ($\epsilon^1_{coef}$) of the GT PDE, (iii) the short and long-term rollout errors at different time horizon ($\epsilon^{T'}_{rollout}$) to evaluate  the cumulative error on both in- and out-of-distribution data, and on changing initial conditions ($IC_2$); and (iv) the respect of physical properties.} 
  \label{tab:res-adv-diff}
  \centering
  \begin{tabular}{|l|r|r|}
   \hline 
    & \multicolumn{1}{|c|}{PDE-FIND}  &  \multicolumn{1}{|c|}{DeepMoD} \\
    & \multicolumn{1}{|c|}{{\scriptsize ${\hat u}_t=-0.9249u_{\mathbf{x}} + 0.0938u_\mathbf{xx} + 0.3865u$}} 
    & \multicolumn{1}{|c|}{{\scriptsize ${\hat u}_t=-0.9992u_\mathbf{x} +  0.0382u_\mathbf{xx}$}}\\
    & \multicolumn{1}{|c|}{{\scriptsize $ + 0.3198tu -0.3098xu -0.0623xu_\mathbf{x}+ 0.0591u^2-0.0118tu_\mathbf{xx}$}} 
    & \multicolumn{1}{|c|}{{\scriptsize $+0.0557 tu_\mathbf{xx}-0.0403u_\mathbf{x}u_\mathbf{x}$ }} \\
    \hline
    $nMAE(\hat{u},u)$~\ref{eq:nMAE} & $0.0083$ \greenpoint &  $0.1166$ \redpoint \\
    \hline
    $\epsilon^{T'=2.5}_{rollout}({\hat{u}},u)$~\ref{eq:rollout} & $1.4e$-$05$ \greenpoint &  $0.0030$ \redpoint \\
    \hline
    $TPR(\hat{\alpha},\alpha)$~\ref{eq:TPR} & $1/4$ \redpoint &  $1/2$ \greenpoint \\
    \hline
    $\epsilon^1_{coef}(\hat{\alpha},\alpha)$~\ref{eq:nL1coef} & $1.1188$ \redpoint &  $0.1442$ \greenpoint \\
    \hline
    $\epsilon^{T'=5}_{rollout}({\hat{u}},u)$~\ref{eq:rollout} & $0.0706$ \redpoint &  $0.0049$ \greenpoint \\
    \hline
   $IC_2:\epsilon^{T'=5}_{rollout}$ & $19.1720$ \redpoint &  $1.4091$ \greenpoint \\
    \hline
    Phys. consistency & \xmark violation of spatial/time invariance & \xmark violation of time invariance \\
    \hline
  \end{tabular}
\end{table}

\begin{table}[t]
  \caption{Results on learned PDEs from both PDE-FIND and DeepMoD for the KdV's equation. A subset of representative evaluation metrics is used, covering (i) the prediction errors ($nMAE$), (ii) the ability to recover the symbolic form ($TPR$) and the coefficients ($\epsilon^1_{coef}$) of the GT PDE, (iii) the short and long-term rollout errors at different time horizon ($\epsilon^{T'}_{rollout}$) to evaluate  the cumulative error on both in- and out-of-distribution data, and on changing initial conditions ($IC_2$); and (iv) the respect of physical properties.} 
  \label{tab:res-kdv}
  \centering
  \begin{tabular}{|l|r|r|}
   \hline 
    & \multicolumn{1}{|c|}{PDE-FIND}  &  \multicolumn{1}{|c|}{DeepMoD} \\
    & \multicolumn{1}{|c|}{{\scriptsize ${\hat u}_t=-5.8984uu_\mathbf{x} -0.9924u_{\mathbf{xxx}}$}} 
    & \multicolumn{1}{|c|}{{\scriptsize ${\hat u}_t=-5.6701uu_\mathbf{x} -0.9695u_{\mathbf{xxx}}$}}\\
      & \multicolumn{1}{|c|}{{\scriptsize $ -0.6104u_\mathbf{x}u_{\mathbf{xx}}$}} 
    & \multicolumn{1}{|c|}{{\scriptsize $-0.5163 u_{\mathbf{xx}}u_{\mathbf{xxx}} -0.0471 u_\mathbf{x}$ }} \\
    \hline
    $nMAE(\hat{u},u)$~\ref{eq:nMAE} & $0.0489$ \redpoint &  $0.0335$ \greenpoint \\
    \hline
    $\epsilon^{T'=20}_{rollout}({\hat{u}},u)$~\ref{eq:rollout} & $0.0019$ \redpoint &  $0.0009$ \greenpoint \\
    \hline    
    $TPR(\hat{\alpha},\alpha)$~\ref{eq:TPR} & $2/3$ \greenpoint &  $1/2$ \redpoint \\
    \hline
    $\epsilon^1_{coef}(\hat{\alpha},\alpha)$~\ref{eq:nL1coef} & $0.1028$ \greenpoint &  $0.1319$ \redpoint \\
    \hline
    $\epsilon^{T'=100}_{rollout}({\hat{u}},u)$~\ref{eq:rollout} & $0.0611$ \greenpoint &  $NaN$ \redpoint \\
    \hline
   $IC_2:\epsilon^{T'=100}_{rollout}$ & $0.2240$ \greenpoint &  $NaN$ \redpoint \\
    \hline
    Phys. consistency & \cmark & \cmark \\
    \hline
  \end{tabular}
\end{table}

\begin{table}[t]
  \caption{Results on learned PDEs from both PDE-FIND and DeepMoD for the Kuramoto Sivashinsky's equation. A subset of representative evaluation metrics is used, covering (i) the prediction errors ($nMAE$), (ii) the ability to recover the symbolic form ($TPR$) and the coefficients ($\epsilon^1_{coef}$) of the GT PDE, (iii) the short and long-term rollout errors at different time horizon ($\epsilon^{T'}_{rollout}$) to evaluate  the cumulative error on both in- and out-of-distribution data, and on changing initial conditions ($IC_2$); and (iv) the respect of physical properties.} 
  \label{tab:res-ks}
  \centering
  \begin{tabular}{|l|r|r|}
   \hline 
    & \multicolumn{1}{|c|}{PDE-FIND}  &  \multicolumn{1}{|c|}{DeepMoD} \\
    & \multicolumn{1}{|c|}{{\scriptsize ${\hat u}_t=-0.9603uu_\mathbf{x} -0.6570 u_\mathbf{xx} -0.7097 u_\mathbf{xxxx}$}} 
    & \multicolumn{1}{|c|}{{\scriptsize ${\hat u}_t=-0.8627uu_\mathbf{x} -0.7138 u_\mathbf{xx} -0.8417 u_\mathbf{xxxx}$}}\\
    & \multicolumn{1}{|c|}{{\scriptsize $-0.4977u_\mathbf{x}u_\mathbf{xx} -0.1264 u_\mathbf{x}u_\mathbf{xxxx} -0.1084 uu_\mathbf{xxx}$}} 
    & \multicolumn{1}{|c|}{{\scriptsize $-0.3847u_\mathbf{x}u_\mathbf{xx}-0.1423 u_\mathbf{x}u_\mathbf{xxxx}$ }} \\
    \hline
    $nMAE(\hat{u},u)$~\ref{eq:nMAE} & $0.7114$ \greenpoint &  $0.8385$ \redpoint \\
    \hline
    $\epsilon^{T'=50}_{rollout}({\hat{u}},u)$~\ref{eq:rollout} & $13.5861$ \greenpoint &  $20.2438$ \redpoint \\
    \hline
    $TPR(\hat{\alpha},\alpha)$~\ref{eq:TPR} & $1/2$ \redpoint &  $3/5$ \greenpoint \\
    \hline
    $\epsilon^1_{coef}(\hat{\alpha},\alpha)$~\ref{eq:nL1coef} & $0.4056$ \greenpoint &  $0.4326$ \redpoint \\
    \hline
    $\epsilon^{T'=100}_{rollout}({\hat{u}},u)$~\ref{eq:rollout} & $39.1099$ \redpoint &  $31.4324$ \greenpoint \\
    \hline
   $IC_2:\epsilon^{T'=100}_{rollout}$ & $52.0380$ \redpoint &  $27.4807$ \greenpoint \\
    \hline
    Phys. consistency & \cmark & \cmark \\
    \hline
  \end{tabular}
\end{table}

\subsection{Correlation between in-distribution interpolation and out-of-distribution extrapolation} \label{sec:in-out}
\textcolor{black}{To further support the conclusions drawn in the previous section, we conducted an additional study running PDE-FIND multiple times on the Korteweg–de Vries PDE. More specifically, we considered 50 different values of the ridge regularization parameter, ranging from $0$ to $0.015$. With this parameterization, which could be viewed as a hyperparameter search procedure, we obtained 50 different discovered PDEs learned from a grid of $86 \times 81=6966$ training simulation data. For each of them, we computed the following metrics:
\begin{itemize}
    \item The L1 error on the coefficients ($\epsilon^1_{coef}(\hat{\alpha},\alpha)$~\ref{eq:nL1coef}). 
    \item The prediction test error on the solutions ($nMAE(\hat{u},u)$ \ref{eq:nMAE}). 
     We used $N_x \times \delta_t \times T$ test data for $T=20,40,60,80,100$, where $[0,20]$ corresponds to the time interval used to learn the 50 PDEs.
    \item The long-horizon rollout test error ($\epsilon^{T}_{rollout}({\hat{u}},u)$ \ref{eq:rollout2}) at time $T=20,40,60,80,100$. We also used $N_x \times \delta_t \times T$ test data.
\end{itemize}}

\begin{figure}
    \centering
    \begin{minipage}{0.45\linewidth}
        \centering
        \includegraphics[width=\linewidth]{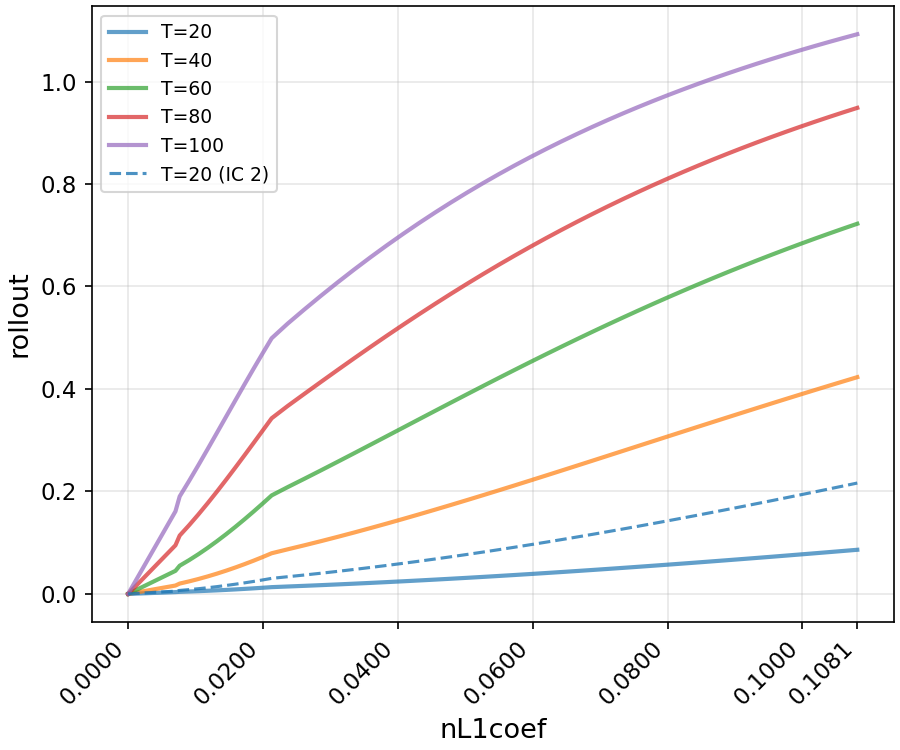}
        \captionof*{figure}{(a)}
    \end{minipage}
    \begin{minipage}{0.45\linewidth}
        \centering
        \includegraphics[width=\linewidth]{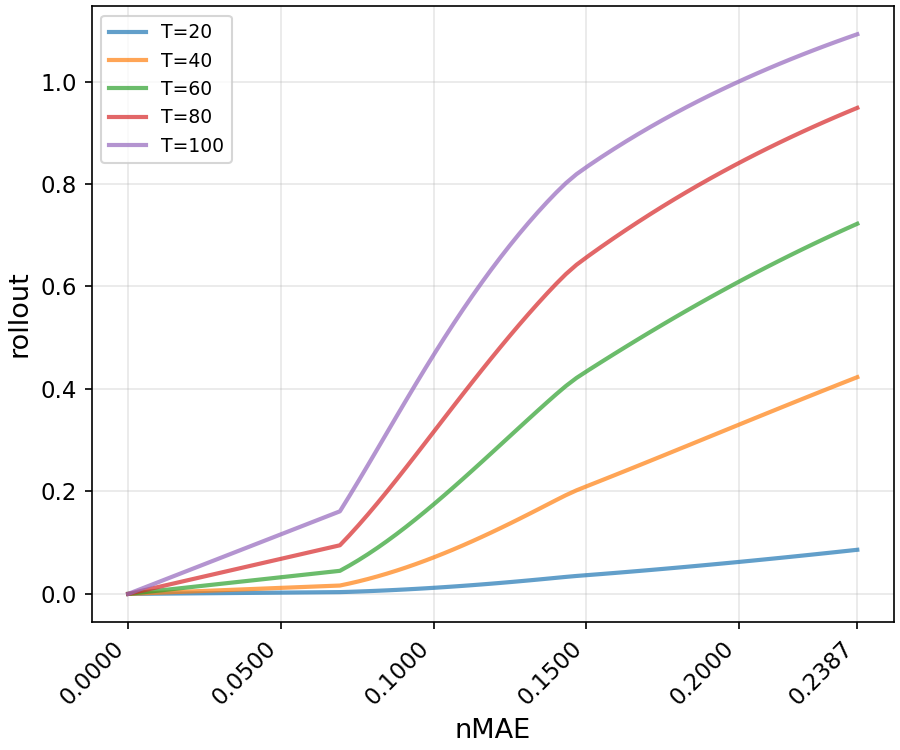}
        \captionof*{figure}{(b)}
    \end{minipage}
    \caption{Results obtained from 50 PDEs learned by PDE-FIND from a grid of $86\times81=6966$ simulation data of the KdV equation with different values of the ridge parameter (from 0 to 0.015). 
    On the left (a) (resp. on the right (b)), the x-axis represents the error on the coefficients $\epsilon^1_{coef}(\hat{\alpha},\alpha)$ (resp. the $nMAE(\hat{u},u)$ on the predictions) computed from the 50 PDEs while the y-axis reports the rollout error $\epsilon^{T}_{rollout}({\hat{u}},u)$ computed at 5 different time-horizons ($T=20, 40, 50, 80, 100$). On each figure, to observe the impact of $T$ on the magnitude of the rollout error, the $50 \times 5$ values are grouped into five solid-line curves, each corresponding to a different value of $T$. For readability, each curve generated from the 50 data points has been rendered as a continuous curve. The last curve (blue dashed-line) represented on Figure (a) corresponds to the rollout error over the training time frame computed from a different initial condition.}
    \label{fig:L1vsRollout}
\end{figure}

\textcolor{black}{Using the training initial conditions (see Table~\ref{tab:ic-params}), Figure \ref{fig:L1vsRollout}.(a) (resp. \ref{fig:L1vsRollout}.(b)) shows the results obtained by cross-comparing the values of $\epsilon^1_{coef}(\hat{\alpha},\alpha)$ (resp. $nMAE(\hat{u},u)$) versus $\epsilon^{T}_{rollout}({\hat{u}},u)$. From these two figures, we can make the following observations:
\begin{itemize}
    \item Within the training time frame (up to $T=20$, see blue solid-line  curve for both figures), a small $\epsilon^1_{coef}(\hat{\alpha},\alpha)$ and a small $nMAE(\hat{u},u)$ are correlated with a small $\epsilon^{T}_{rollout}({\hat{u}},u)$. This is particularly apparent up to 0.02 for the coefficient error and up to 0.07 for the prediction error. One could therefore be misled into concluding that the discovered PDEs are "\textit{almost equivalent}" to the GT based solely on their \textit{interpolation} capacity (over $[0,20]$) evaluated by these three metrics.
    \item However, as soon as we assess their \textit{extrapolation} ability  beyond $T=20$ (see the other four solid-line curves), the long-horizon rollout error rapidly diverges when the learned PDEs differ, \textbf{even slightly} (\textit{i.e.} very small $\epsilon^1_{coef}(\hat{\alpha},\alpha)$ or $nMAE(\hat{u},u)$), from the ground-truth PDE. 
    \item This therefore demonstrates that relying on  prediction or coefficient-based test errors alone, as done in many PDE discovery papers (see Table \ref{tab:sota2}), can be misleading. We therefore argue that whenever long-horizon predictions are possible (\textit{i.e.} when the GT is known or additional data is available beyond the training time interval),  it is essential to evaluate the long-horizon rollout error $\epsilon^{T}_{rollout}({\hat{u}},u)$ as well before drawing conclusions. 
    \item If it is not operationally feasible to obtain OOD data, an alternative way to circumvent this issue is to use varying initial conditions and evaluate the capacity of the learned equations to adapt to these changes. This is what we did in Figure \ref{fig:L1vsRollout}.(a) where the blue dashed-line curve reports the results with a different initial condition (see Table~\ref{tab:ic-params}). We can observe that simulating the 50 PDEs, which don't change and thus give the same coefficient error, results in a significant increase in the rollout error compared with that observed with the training initial condition. 
\end{itemize}
}

\section{Towards standardizing PDE evaluation practices: guidelines and protocol}
\label{sec:guidelines}

\textcolor{black}{The in-depth analysis conducted in this paper which led to the taxonomy presented in Sections \ref{sec:GT} and \ref{sec:noGT}, together with the observations drawn from the experiments above, paves the way for the formulation of a set of guidelines to help practitioners avoid the pitfall of  drawing erroneous conclusions about discovered PDEs. 
The recommendations below are structured around three questions that characterize the context in which the practitioner operates.
They  constitute the building blocks of {\sc Pde-Check}, a rules-based protocol that aims to standardize the evaluation task while ensuring that  all five key questions \textbf{Q1}$\rightarrow$\textbf{Q5} are addressed.}


\subsection{Discover or Recover? That's the question}
 Somewhat surprisingly, the distinction between discovery and recovery of PDEs, as related but distinct tasks, is not often made explicit in scientific papers. Yet, we argue that this differentiation is crucial because it determines the form of the evaluation framework, and therefore the categories of metrics that should be used or not. 
The finding that emerges from the PDE discovery literature is that the aim of most papers is to introduce a new equation identification algorithm, and  compare the latter with other competing methods. 
In that case, benchmark PDEs are employed for which the complete ground truth is known, including the analytical form $PDE_{\cal GT}$, the parameter values as well as the associated initial and boundary conditions. 
 In this setting, the task under consideration is not to genuinely discover a new theory, but rather to evaluate, as a proof of concept, the algorithm’s ability to \textbf{accurately recover the governing law}  $PDE_{\cal GT}$.
 The evaluation can then be carried out using metrics designed to recover the symbolic form of the PDE ({\it e.g.,} $TPR$ \ref{eq:TPR}, $P$~\ref{eq:Precision}, $R$~\ref{eq:Recall}, $J$~\ref{eq:Jaccard}, $TED$~\ref{eq:TED}, $SRecov$~\ref{eq:CAS}, as presented in Section~\ref{sec:terms}) as well as its coefficients ({\it e.g.,} $\epsilon^2_{coef}$~\ref{eq:nL2coef}, $\epsilon^1_{coef}$~\ref{eq:nL1coef}, $\epsilon^\infty_{coef}$~\ref{eq:maxerror},
    $w\epsilon^2_{coef}$~\ref{eq:wL2coef}, $NDCG$~\ref{eq:NDCG}, $T_G~\ref{eq:Tanimoto}$, as discussed in Section~\ref{sec:coef}). 
 It is worth noting that metrics aimed at measuring the sparsity of the learned equation ({\it e.g.,} $S_{terms}$~\ref{eq:sparsity},
    $\mathcal{C}(\hat{\alpha}^\top\Theta)$~\ref{eq:ExpTree},    $\mathcal{C}_2(\hat{\alpha}^\top\Theta)$~\ref{eq:ExpTreeV2}, see Section~\ref{sec:sparsity}) are of little interest here, since the GT is known and the sparsity is therefore already accounted for by those incorporating, \textit{e.g.}, the notions of \textit{recall} and \textit{precision}. Thus, if the GT is correctly recovered, the procedure can stop there.
    
On the other hand, when $PDE_{\cal GT}$ is not perfectly recovered, unless the discrepancy in the coefficients is numerically negligible,  it becomes crucial to assess the impact of spurious terms or  coefficient deviations. As we have seen from the experiments conducted so far, small discrepancies can have a significant impact on the validity of the PDE. 
Claiming that $PDE_{\cal GT}$ is \enquote{\textit{well recovered}} on the basis of a certain similarity to its structural form, without measuring the consequences of  even small  differences, cannot be considered a scientifically valid approach. However,  a number of papers conclude their experiments by typically stating: \enquote{{\it Our novel algorithm A outperforms method B because the symbolic shape and coefficients of the  PDE identified with A are more similar to $PDE_{\cal GT}$ than that of competitor B.}} Yet, nothing guarantees the stability and physical consistency of the equation learned by \textit{A} based solely on the metrics presented in Sections~\ref{sec:terms} and \ref{sec:coef}. 
To avoid such premature and potentially misleading conclusions, a common yet necessary strategy is to employ in-distribution validation data
simulated from $PDE_{\cal GT}$, and compare them with solutions estimated from the identified PDE. This can be done with pointwise errors-based metrics that aim to evaluate the interpolation capacity of the equation, such as those discussed in Section \ref{sec:errors} ({\it e.g.,} $MSE$~\ref{eq:nMSE}, $MAE$~\ref{eq:nMAE}, $f_{MSE}$~\ref{eq:f_{MSE}}, heatmaps, etc.). 
 Leveraging these metrics often makes it possible to capture certain types of pathological behaviors by checking the local consistency of the learned equation with respect to $PDE_{\cal GT}$.  While this \textit{interpolation} evaluation may be sometimes sufficient, this is not generally the case, and even less so when it comes to \textbf{knowledge discovery}, \textit{i.e.} when $PDE_{\cal GT}$ is assumed to be unknown. In this case, metrics assessing the form and coefficients become unusable, while those such as \textit{MAE} and \textit{MSE}, computed solely on in-distribution data, may be insufficient and even result in conflicting conclusions, as demonstrated by the experiments in the previous section. 

 Our expertise gained from the
 in-depth study and extensive experimentation with
  the different categories of metrics presented in Sections~\ref{sec:GT}, \ref{sec:errors}, \ref{sec:sparsity} and \ref{sec:tradeoff-sparsity-accuracy} shows that, depending on the application under consideration, different evaluation criteria may disagree, potentially calling into question the resulting rankings of identified equations or discovery algorithms. What should be done in such cases or when a tie needs to be broken? 
We argue that, whenever the application allows it, the most relevant tools for making meaningful comparisons between competing equations or algorithms are undoubtedly criteria that (i) assess the \textit{extrapolation} capabilities of the models, \textit{i.e.}, their ability to adapt to out-of-distribution scenarios (Section~\ref{sec:generalization}), and (ii) verify their physical consistency, namely, whether they do not violate fundamental physical properties (Section~\ref{sec:consistency}). These two crucial aspects are discussed in the following two subsections.


\subsection{Interpolation or Extrapolation?  Get the best of both worlds...if possible} 
To avoid overstating the validity of a discovered PDE, 
it is crucial, beyond interpolation capacity, to evaluate its extrapolation behavior on out-of-distribution data. Extrapolation can be seen as a more stringent criterion as it tests  whether the learned equation captures the underlying dynamics beyond the regime seen during its identification. Thus,  \textbf{interpolation and extrapolation play complementary roles in post-hoc evaluation}, each yielding different insights into the quality of the discovered PDE. {\bf Interpolation ensures local consistency}, verifying that the inferred dynamics is coherent in observed regimes. On the other hand, \textbf{extrapolation} provides a stronger indicator of whether the PDE {\bf captures true physical structure rather than statistical regularities}. 
In this context, we discussed in Sec.~\ref{sec:generalization} different scenarios for evaluating the OOD generalization capacity of an equation.  Based on our analysis and the experiments conducted, we strongly advocate the use of the long-horizon rollout error $\epsilon^{T}_{rollout}({\hat{u}},u)$ (Eq.~\ref{eq:rollout2}) and varying initial conditions to evaluate the stability of the equation. While this strategy is essential for validating scientific discoveries, we strongly recommend also using it when $PDE_{\cal GT}$ is known. 
 The experiments conducted in Section \ref{sec:expePDEFIND-DEEpMoD} confirm this guideline. Despite having symbolic equations close to $PDE_{\cal GT}$, with metrics of Sec.~\ref{sec:GT} and \ref{sec:noGT} showing satisfactory performance, errors can accumulate over the long term and lead to substantial divergence in the predictions. \textbf{Only OOD generalization could capture such shortcomings}.
 
Paradoxically, while real-world applications have a pressing need for extrapolation to validate scientific discovery, they often suffer from a  lack of observation data (compared to the unlimited number of simulations available when $PDE_{\cal GT}$ is known), preventing the straightforward implementation of the experimental setup described in  Sec.~\ref{sec:generalization}. To overcome this issue, different solutions exist, one of them consisting in learning a surrogate model allowing to generate data under new conditions. For instance, neural operators like FNOs~\citep{LiKALBSA21}, which model mappings between function spaces, can be used as a surrogate numerical solver. Once trained on in-distribution data, it can generalize by outputting solution trajectories for new conditions, thereby providing a practical way to construct controlled OOD scenarios. However, it should be noted that, in addition to numerical errors associated with trajectories simulated from the learned PDE, extrapolation may also be impacted by approximation and generalization errors of the FNO, which are respectively related to the model architecture and the amount of data available for training.

\subsection{To know or not to know a priori?}
 In Section~\ref{sec:consistency}, we showed how physical properties can be leveraged to complement evaluation metrics. 
However, such strategy raises the key question of the {\bf access to  physical priors} about the characteristics that the discovered equation is expected to satisfy, such as conserved quantities or symmetries. 
 The review of the PDE discovery literature  reveals that a number of  papers that employ  inductive bias derived from physical laws often rely on the  assumption that these properties are known in advance.  
This is the case, {\it e.g.}, in  \cite{yang2026discoveringsymbolicdifferentialequations} where symmetry is enforced by using differential invariants of the symmetry group as the variable set, hypothesizing this group is already given. In \cite{pmlr-v235-dalton24a}, the authors incorporate  a priori known Lie symmetries to enhance Gaussian Processes-based models both in the context of forward and inverse problems (for learning unknown PDE parameters). 
In the same vein, in \cite{zhang2022enforcingcontinuoussymmetriesphysicsinformed}, Lie symmetries are supposed to be known and are embedded into the loss function of PINNs to solve inverse problems.

Assuming that such physical characteristics are given remains reasonable provided that one has some background knowledge on the class of physical dynamical systems for which the PDE discovery is being performed. In \cite{yang2026discoveringsymbolicdifferentialequations}, the authors argue that physicists themselves often operate in this manner, hypothesizing the symmetries of a system before seeking governing equations allowed by those invariances. We illustrated this way of proceeding using the case of laser–matter interaction in the \enquote{Running Example} paragraph of Section~\ref{sec:consistency}.  
 However, for a number of real-world situations, this postulate becomes less tenable because of a lack of reliable prior knowledge about the underlying physical structure, making it difficult to impose physical constraints on a learned PDE. This can happen, {\it e.g.}, in socio-economic systems, where interactions are heterogeneous, in biology, where mechanisms are only partially understood and vary across scales, or in surface engineering involving the complex intertwining of several phenomena. To circumvent this limitation,  recent works aimed to automatically discover physical properties from data and leverage them for PDE discovery/evaluation. For example, in \cite{yang2024latentspacesymmetrydiscovery}, the authors propose {\sc LaLiGAN}, a novel
generative model which can discover symmetries of nonlinear
group actions. They show that their method   can be applied to downstream tasks such as equation discovery.
On the other hand, in \cite{gabel2023learningliegroupsymmetry} and \cite{moskalev2023lieggstudyinglearnedlie}, to avoid 
hard-coding symmetries in neural networks, the authors aim to  discover Lie group symmetry transformations present
in the training dataset. The reader interested in learning more about symmetry discovery may also refer to \cite{udrescu2020aifeynmanphysicsinspiredmethod,otto2025unifiedframeworkenforcediscover,gui2025discovering,yang2024symmetryinformedgoverningequationdiscovery,dehmamy2021automaticsymmetrydiscoverylie}. Similarly, note that several recent papers have investigated the discovery of conservation laws from data (see, \textit{e.g.}, \cite{Doshi2025,chen2024datadrivendiscoveryconservationlaws}).

\subsection{PDE-CHECK: An operational evaluation protocol}
\textcolor{black}{The previous sections highlighted three key elements for characterizing the evaluation scenario faced by the practitioner: (i) the actual objective pursued, \textit{i.e.} discovery or recovery of equations; (ii) the access to out-of-distribution data; and (iii) the availability of physical knowledge. These elements form the basis of the rules underlying {\sc Pde-Check}, the evaluation protocol proposed in this section, as a synthesis of our analytical and review work. Although the evaluation of an identified PDE, and consequently, the comparison of competing equations/algorithms, remains a challenging problem, {\sc Pde-Check} has the advantage of being readily usable by leveraging clear rules.  The goal of {\sc Pde-Check} is three-fold: (i) it guides practitioners through the evaluation process by suggesting families of metrics adapted to the considered scenario; (ii) by collecting multiple information about the identified PDE, it reduces the risk of overstatements regarding the validity of the latter; (iii) although it remains open to further refinement, {\sc Pde-Check} provides an operational tool for standardizing the evaluation of PDEs. Its pseudo-code is presented 
in Algorithm~\ref{alg:pde-check}}.\\

\begin{algorithm}[ht]
\caption{{\sc Pde-Check}}
\label{alg:pde-check}
\textcolor{black}{
\begin{algorithmic}[1]
\If{the GT PDE is available (\textbf{{\it i.e.} RECOVERY} task)}
    \State \textbf{Q5:} Check \textit{Symbolic Form} using, {\it e.g.,} $TPR$ \ref{eq:TPR}, $P$~\ref{eq:Precision}, $R$~\ref{eq:Recall}, $J$~\ref{eq:Jaccard}, $TED$~\ref{eq:TED}, $SRecov$~\ref{eq:CAS}  and 
    \Statex \hspace{\algorithmicindent} \hspace{1.6cm} \textit{Coefficients} using, {\it e.g.,} $\epsilon^2_{coef}$~\ref{eq:nL2coef}, $\epsilon^1_{coef}$~\ref{eq:nL1coef}, $\epsilon^\infty_{coef}$~\ref{eq:maxerror},
    $w\epsilon^2_{coef}$~\ref{eq:wL2coef}, $NDCG$~\ref{eq:NDCG}, $T_G~\ref{eq:Tanimoto}$
    \If{GT is accurately recovered}
        \State \textbf{EXIT} $\rightarrow$ Analysis of the results - Comparison between PDEs/Algorithms
    \Else 
        \If{Physical priors are available}
            \State \textbf{Q2:} Check \textit{Physical Consistency} (invariance, conserv. laws, symmetries, dim. consistency)
            \If{Violation of physical properties}
                \State \textbf{EXIT} $\rightarrow$ The PDE is not physically consistent
            \EndIf
        \EndIf
        \State \textbf{Q1:} Check \textit{In-distribution prediction accuracy}
        using, {\it e.g.,} $MSE$~\ref{eq:nMSE}, $MAE$~\ref{eq:nMAE}, $f_{MSE}$~\ref{eq:f_{MSE}}, heatmaps
        \State \textbf{Q4:} Check \textit{OOD Generalization}
        using, {\it e.g.,} $\epsilon^{T'}_{rollout}$~\ref{eq:rollout2}, $Conv_x$~\ref{eq:mesh-refinement}, $Conv_t$~\ref{eq:time-refinement}
        and varying IC/BC
        \State \textbf{EXIT} $\rightarrow$ Analysis of the results - Comparison between PDEs/Algorithms
    \EndIf
\Else{ ({\it i.e.} \textbf{DISCOVERY} task)}
    \If{Physical priors are available}
            \State \textbf{Q2:} Check \textit{Physical Consistency} (invariance, conserv. laws, symmetries, dim. consistency)
            \If{Violation of physical properties}
                \State \textbf{EXIT} $\rightarrow$ The PDE is not physically consistent
            \EndIf
        \EndIf
    \State \textbf{Q1:} Check \textit{In-distribution prediction accuracy}
        using, {\it e.g.,} $MSE$~\ref{eq:nMSE}, $MAE$~\ref{eq:nMAE}, $f_{MSE}$~\ref{eq:f_{MSE}}, heatmaps
    \State \textbf{Q3:} Check \textit{Sparsity}
    using, {\it e.g.,} $S_{terms}$~\ref{eq:sparsity},
    $\mathcal{C}(\hat{\alpha}^\top\Theta)$~\ref{eq:ExpTree}, $\mathcal{C}_2(\hat{\alpha}^\top\Theta)$~\ref{eq:ExpTreeV2}
    \State \textbf{Q1 \& Q3:} Check \textit{Accuracy--Sparsity trade-off}
    using, {\it e.g.,} $Score$~\ref{eq:Score}, $Reward_1$~\ref{eq:reward1}, $Reward_2$~\ref{eq:reward2},
    \Statex \hspace{\algorithmicindent} \hspace{1.6cm} $AIC$~\ref{eq:cAIC}, $BIC$~\ref{eq:BIC}, $PIC$~\ref{eq:PIC}, $MDL$~\ref{eq:MDL-Feynman} \ref{eq:mdl-SymLang}
    \If{OOD data is available}
        \State \textbf{Q4:} Check \textit{OOD Generalization}
        using, {\it e.g.,} $\epsilon^{T'}_{rollout}$~\ref{eq:rollout2}, $Conv_x$~\ref{eq:mesh-refinement}, $Conv_t$~\ref{eq:time-refinement}
        and varying IC/BC
    \EndIf
    \State \textbf{EXIT} $\rightarrow$ Analysis of the results - Comparison between PDEs/Algorithms
\EndIf
\end{algorithmic}}
\end{algorithm}

 \textcolor{black}{Through its successive rules, {\sc Pde-Check} aims to cover all the properties that need to be evaluated in order to decide on the status of an equation, \textit{i.e.} to determine if the latter can reasonably be considered valid, or at least better than another. At one point or another, {\sc Pde-Check} addresses each of the five questions \textbf{Q1}$\rightarrow$\textbf{Q5} investigated throughout this article. Although we do not exclude any metrics by default from the different options available as the evaluation process unfolds, our analysis from Sections \ref{sec:GT} and \ref{sec:noGT} has shown that some of them have certain shortcomings  that may prompt us to set them aside. This is the case, \textit{e.g.}, for $TED$, $D\!\left(\Theta^{\top}\bm{\hat{\alpha}}\right)$, $S_{terms}$, $AIC$ or $BIC$ (for more details, we refer the reader to the pros and cons boxes provided at the end of each category). \\
 To conclude this section, it is worth noting that {\sc Pde-Check} fully allows for the development and subsequent use of new metrics, which may be introduced in future work, provided we indicate which category of the taxonomy they belong to. They can then simply be added to the list of evaluations to be performed at the relevant stage of the {\sc Pde-Check} evaluation process.}

\section{And what next? Future directions and open problems} \label{sec:perspectives}

While some of the metrics discussed in this paper possess more desirable characteristics than others, notably by providing better coverage of the key requirements that a discovered PDE is expected to satisfy (see Table~\ref{tab:sota}), it is worthwhile mentioning that the most of them are strongly grounded in a pointwise perspective. This applies to all metrics based on prediction errors, residuals, and trajectory-based evaluations of Sec.~\ref{sec:noGT}. A common consequence of these measures is that they provide only a partial assessment of the PDE quality, as they do not capture higher-level functional properties beyond local discrepancies and trajectory matching. Although more challenging to implement, we argue that a more holistic evaluation is needed by accounting not only for local approximation errors but also for the structural, geometric, and regularity properties of the solution. We sketch below three perspectives, among other potential ones, which are likely to address this need. They rely on the use of (i) the Sobolev semi-norm, (ii) the Lyapunov exponents and (iii) the optimal transport and gradient flows. On the other hand, leveraging \citep{lou2026datadrivendiscoverygoverningdifferential}, we also analyze the necessity to develop new metrics related to the solvability of the identified PDE. Indeed, an equation, even if correct, has little utility if no solver is available, or if the simulation time is not reasonable. Finally, we present a last promising line of research on the evaluation of fractional PDEs, which, although richer than classic equations, raise numerical challenges and require a specific assessment to be of practical interest.\\
   
    $\blacktriangleright$ \textbf{Functional analysis of PDEs:} 
    The Sobolev semi-norm of a function takes into account not only the values of the function itself but also its derivatives up to a given order $k$, providing insights into its regularity and smoothness. Fundamentally rooted in functional analysis, it is defined as follows:
\begin{eqnarray}
|u|_{H^k(\Omega)}
=
\left(
\sum_{|\beta|\leq k}
\|D^\beta u\|_{L^2(\Omega)}^2
\right)^{1/2}
\label{eq:Sobolev-norm}
\end{eqnarray}
where $D^\beta u
=
\frac{\partial^{|\beta|} u}
{\partial \mathbf{x}_1^{\beta_1} \cdots \partial \mathbf{x}_d^{\beta_d}}
$, with $\beta=(\beta_1, \beta_2, \dots, \beta_d) \in \mathbb{N}^d$ is a multi-index and $|\beta| = \beta_1 + \cdots + \beta_d$.
    
    This  norm has been used for a long time in numerical analysis, notably to analyze the convergence, stability and approximation quality of numerical solutions to PDEs \citep{citeulike:8936200}, and more recently in PiML models for regularizing PINNs \citep{10.3150/24-BEJ1799} allowing to theoretically maintain consistency with the underlying physics.
Surprisingly, while it enables better capture of local structures and helps avoid learning overly smooth solutions, it is worth noting that the Sobolev norm has  scarcely been leveraged in the literature concerned with PDE discovery. One  possible reason is that as $k$ increases, obtaining and differentiating through these higher-order terms becomes progressively more costly and can make gradient-based training unstable. Moreover, using the Sobolev norm in its most general form as defined in Eq.~\ref{eq:Sobolev-norm}, {\it e.g.} as a regularization term, makes it difficult to derive generalization guarantees. Therefore, in \cite{10.3150/24-BEJ1799}, the authors restrict their convergence analysis of PINNs to affine differential operators, {\it i.e.} limiting the approach to linear PDEs.
However, in post-hoc PDE evaluation, we argue that these constraints can be more readily relaxed, paving the way for the design of new metrics based on this norm. Note that in MDBENCH~\citep{bideh2026mdbenchbenchmarkingdatadrivenmethods}, the authors share the observation that there is no established metric for quantifying equation fidelity, and they claim that there is an urgent need for a\textit{ \enquote{robust metric that captures functional equivalence, e.g., using a discretized version of the Sobolev semi-norm for PDE systems}}.\\


$\blacktriangleright$ \textbf{Lyapunov exponents and sensitivity analysis:} 
A complementary strategy to that of \citep{norman2026learninggradientflowusing} to design weighted errors according to the impact of each term of the PDE may account for the exponential amplification of perturbations along unstable dynamical directions. Let $\lambda > 0$ denote a dominant Lyapunov exponent \citep{wilkinson2016lyapunovexponentsinteresting} associated with the underlying physical system. $\lambda$ quantifies the exponential rate of separation of infinitesimally close trajectories. Built upon this exponent, one may define a time-dependent weighted sensitivity of the solution with respect to each parameter $\bm{{\alpha}}_i$, as follows:
\[
w_i(t) \propto \exp(\lambda t)\left\| \frac{\partial u}{\partial \bm{{\alpha}}_i} \right\|_{2}^2.
\]
This would lead to a Lyapunov-informed metric of the form:
\begin{eqnarray}
w_{Lyap}\epsilon^2_{coef}(\bm{\alpha},\bm{\hat{\alpha}})    = \sum_{i=1}^{|\theta|} \left( \int_0^T \exp(\lambda t)\left\| \frac{\partial u(t)}{\partial \bm{{\alpha}}_i} \right\|_{2}^2 \, dt \right)
(\bm{{\alpha}}_i - \bm{{\hat{\alpha}}}_i)^2 \in \mathbb{R}_{\geq 0}.
\label{eq:Lyapunov-error}
\end{eqnarray}


Compared to Eq.~\ref{eq:wL2coef}, this formulation explicitly accounts for the fact that perturbations in chaotic or weakly stable regimes may remain negligible in the short term but grow exponentially over time.
However, although intrinsically interesting, this metric is difficult to compute because calculating the Lyapunov exponents is known to be NP-hard in general \citep{Altschuler_2020}. 
Even though they can be estimated efficiently for certain PDEs, it remains very expensive for PDEs such as Kuramoto–Sivashinsky (the interested reader may refer to \cite{edson2019} to understand how to evaluate the Lyapunov exponents in this case). 
More generally, estimating the largest Lyapunov exponent in $w_{Lyap}\epsilon^2_{coef}(\bm{\alpha},\bm{\hat{\alpha}})$ typically requires long trajectory integrations, reorthonormalizations, making it computationally expensive.
\\
   
  $\blacktriangleright$ \textbf{PDEs viewed as gradient flows}: Another promising direction in post-hoc evaluation is to consider PDEs as evolving probability measures, and compare the ground truth data with the solutions of the discovered equation through the lens of Optimal Transport (OT)~\citep{inbook} and Gradient Flows (GF) \citep{gagneux2025a}. 
Although OT is a longstanding theory~\citep{Monge1781}, it has only attracted renewed interest from the machine learning community over the past decade following  (i) the design of efficient algorithms for Wasserstein distance computation ({\it e.g.} Sinkhorn algorithm~\citep{NIPS2013_af21d0c9}), (ii) the development of Wasserstein GANNs~\citep{arjovsky2017wassersteingan}, and the use of GF in generative AI~\citep{xu2025generativediscoverypartialdifferential}.

On the other hand, OT and GF offer significant advantages for capturing the functional properties of dynamical systems. 
Indeed, many PDEs, such as Fokker-Planck, Keller-Segel or diﬀusion, advection and aggregation equations in general, with many related real-world applications, have been shown to be  interpreted as GF \citep{santambrogio2016euclideanmetricwasserstein,Arnrich_2011,ferreira2015gradientflowstimedependentfunctionals}. This provides an original way to understand the evolution of an equation as the steepest descent of an energy functional in a (typically) Wasserstein metric space, offering the possibility to leverage its underlying variational structure. 
Under this viewpoint, OT and GF provide new mathematical tools for studying stability and long time behavior of PDEs. 
 They notably enabled the design of a new class of numerical methods (see {\it e.g.} \cite{1504,zbMATH06499444,carrillo2021primaldualmethodswasserstein})  endowed  with better stability and convergence guarantees. 

In this context, 
 a few papers have exploited the Wasserstein distance to evaluate the quality of a learned PDE. For instance, in  \cite{lang2020learninginteractionkernelsmeanfield}, the authors deal with learning systems of interacting particles and compare  GT and discovered solutions by leveraging the Wasserstein distance. We recall that the latter measures the discrepancy between two probability distributions by computing the minimum transportation cost required to transform one distribution into the other. 
Under this definition, it enables the characterization of properties that are not accessible through metrics relying on pointwise errors or trajectory-based evaluation. For instance, two otherwise identical PDE solutions with a slight spatial or temporal shift will be considered significantly different under pointwise metrics, despite exhibiting nearly identical dynamics. By contrast, the Wasserstein distance will capture this geometric displacement and therefore identifies the solutions as highly similar. Moreover,  unlike most of other alternative metrics, the Wasserstein distance is characterized by remaining informative over longer horizons, even with non-overlapping support. 

Some recent works have leveraged the Wasserstein distance from a theoretical standpoint to establish new stability estimates in equation discovery. This is the case, for example, in \cite{carrillo2025sparseidentificationnonlocalinteraction} where the authors deal with sparse identification of interaction kernels. Focusing on PDEs with interaction potentials (which arise in many applications) that can be formulated as GF, they control the Wasserstein distance between solutions generated using the true and learned interaction potentials. In \cite{lu_2022}, discovery is envisioned under the assumption that  the  dataset is not governed by the same dynamics. In this parametric PDE learning setting, they combine the
Wasserstein distance with diffusion maps, to discover conserved quantities that vary across trajectories. 

Despite these few references, it is important to note that works exploiting OT and GF for PDE evaluation (during learning or as downstream task) are still at an early stage. Beyond the straightforward use of the Wasserstein distance between solution distributions (as seen above), no well-established methodology has yet emerged.\\ 


$\blacktriangleright$ \textbf{Discovered  PDEs must also be numerically solvable to be useful:} As discussed throughout this survey,   PDE evaluation metrics aim to deal with the requirements associated with Questions {\bf Q1} to {\bf Q5}. Nevertheless, as recently highlighted in \cite{lou2026datadrivendiscoverygoverningdifferential}, even an equation that leads to accurate solutions, which is sparse, physically consistent, and generalizes well, may still be of limited practical value. For instance, a supposedly good PDE according to these criteria may also (i) exhibit non-standard terms which significantly complexify its numerical solution, (ii) describe an ill-posed problem (as discussed earlier in Sec.~\ref{sec:consistency}), (iii) lead to an unstable numerical discretization due, {\it e.g.}, to excessively stiff dynamics, or more generally (iv) entail a prohibitive computational cost, making large-scale simulations or uncertainty quantification impractical.
Numerical tractability is therefore not motivated by mathematical aesthetics, but because in most real-world applications, a PDE is only  useful if it can actually be practically deployable, that is, simulated reliably, repeatedly, and at scale.
As stated by \cite{lou2026datadrivendiscoverygoverningdifferential}, this issue has largely been overlooked in the literature, thereby justifying the need for a new class of dedicated metrics. 
 Recent works have started to consider simulation time as a PDE quality criterion. This is the case in  
PhysPDE~\citep{feng2025physpde} where the efficiency is measured by the total {\it simulation wall-time}. The wall-clock timing is also employed in AutoSINDy \citep{basiri2026discoverynonlineardynamicsautomated} allowing, {\it e.g.} to detect stiff PDEs that require finer time-stepping, leading to substantially longer integration times. 
Despite these few papers (see also SciGym \citep{duan2026measuring}), further investigation is still required to develop metrics that can  detect and assess solvability issues.\\

$\blacktriangleright$ \textbf{Towards evaluation metrics for fractional PDEs}: Fractional PDEs generalize PDEs by allowing  derivatives of arbitrary, thus possibly non-integer, order. They
arise in various physical phenomena where non-locality plays a crucial role \citep{SUN2018213}, such as in anomalous diffusion \citep{MAGIN2008255}, transport in porous media \citep{PLOCINICZAK2015169}, viscoelastic materials \citep{ZHANG2024104699}, geophysical modelling \citep{geophysics}, or in biological systems \citep{Ali2024}.  Although the discovery of such PDEs has been investigated in a few papers \citep{Xiangnan,Singh2022,Vats,Singh2021,gulian2019machinelearningspacefractionaldifferential}, these equations pose several challenges lying in their nonlocal operators, which introduce long-range spatial or temporal interactions and significantly complicate their numerical processing \citep{Chaudhary_2026}.
Due to this nonlocal nature of the fractional operator, conventional evaluation metrics discussed in our study may become misleading. For instance, the sparsity/complexity of a fractional law is  more difficult to capture since a single fractional derivative, seen as an atomic element, can encode substantially richer dynamics than a integer order-based term. 
By way of example, in \cite{gulian2019machinelearningspacefractionaldifferential}, the authors show in their experiments that one fractional derivative with the same approximation quality can be used  as an interpolation of two integer-order operators. This example shows that, although the resulting model is objectively sparser, it would be questionable to claim that the use of a fractional derivative, which is inherently more expressive and therefore more complex in a certain sense, leads to a simpler  governing law. This illustrates that, in the context of non-integer-order PDEs, it becomes necessary to \textbf{distinguish more carefully between the notions of sparsity and simplicity/interpretability} through more refined evaluation metrics.
On the other hand, the numerical evaluation of residuals is more sensitive to the choice of fractional derivative definition ({\it e.g.}, Caputo, Riemann–Liouville, or Riesz) depending on the property one primarily aims to preserve, which are not taken into account by classic metrics. 
Moreover, the fractional orders themselves constitute continuous model parameters whose correctness should be evaluated alongside the PDE coefficients and the presence/absence of terms, adding an extra dimension to the evaluation procedure.  
These distinctive characteristics introduced by non-integer-order derivatives call for the development of a new generation of post-hoc evaluation metrics. We argue that this is essential to fairly assess future fractional PDE discovery methods, whose emergence is expected to accelerate in light of the significant progress made in the numerical computation or neural approximation of these derivatives, and the growing interest in fractional models due to their  ability to describe complex nonlocal phenomena.


\subsubsection*{Acknowledgments}
This research was funded by the French National Research Agency (ANR) under the project MELISSA~"\textit{Methodological contributions in statistical Learning InSpired by SurfAce engineering}" - ANR-24-CE23-7140-01.

\noindent The authors gratefully acknowledge Eduardo Brandao for his careful reading of the manuscript and for his valuable and insightful comments, which contributed to its improvement.

\bibliographystyle{tmlr}
\bibliography{biblioEvalPDE}

\end{document}